%% file: bare_jrnl_compsoc.tex
\documentclass[10pt,journal,compsoc]{IEEEtran}
\ifCLASSOPTIONcompsoc
  \usepackage[nocompress]{cite}
\else
  \usepackage{cite}
\fi
\usepackage{subcaption}
\ifCLASSINFOpdf
  \usepackage[pdftex]{graphicx}
\else
\fi
\usepackage{amsmath}
\usepackage{amssymb}
\usepackage{tabularx}
\usepackage{tikz}
\usetikzlibrary{positioning,shapes.geometric}
\usepackage{booktabs, multirow, makecell} 
\newcommand\MYhyperrefoptions{bookmarks=true,bookmarksnumbered=true,
pdfpagemode={UseOutlines},plainpages=false,pdfpagelabels=true,
colorlinks=true,linkcolor={black},citecolor={black},urlcolor={black},
pdftitle={Model Compression and Efficient Inference for Large Language Models: A Survey},
pdfsubject={Typesetting},
pdfauthor={Wenxiao Wang},
pdfkeywords={Large language models, model compression, efficient inference}}

\usepackage[\MYhyperrefoptions,pdftex]{hyperref}
\begin{document}
%
\title{Model Compression and Efficient Inference for Large Language Models: A Survey}
%
%
%
%

\author{Wenxiao Wang$^\dagger$,
        Wei Chen$^\dagger$,
        Yicong Luo$^\dagger$, 
        Yongliu Long$^\dagger$, 
        Zhengkai Lin$^\dagger$, 
        Liye Zhang$^\dagger$, 
        Binbin Lin,
        Deng Cai,
        Xiaofei He~\IEEEmembership{Senior~Member,~IEEE}
\IEEEcompsocitemizethanks{
\IEEEcompsocthanksitem $^\dagger$ indicates equal contributions with specific .
\IEEEcompsocthanksitem Wenxiao Wang, Yicong Luo, Yongliu Long, and Binbin Lin are with the School of Software Technology, Zhejiang University.
\IEEEcompsocthanksitem Liye Zhang is with School of Mathematical Sciences, Zhejiang University.
\IEEEcompsocthanksitem Wei Chen, Zhengkai Lin, Deng Cai, and Xiaofei He are with State Key Lab of CAD$\&$CG, Zhejiang University.}
\thanks{Manuscript released in February 10, 2024.}}

%
%

\newcommand\ie{\textit{i.e.}}
\newcommand\eg{\textit{e.g.}}
\newcommand\etc{\textit{etc}}

\markboth{Journal of \LaTeX\ Class Files,~Vol.~14, No.~8, August~2015}%
{Shell \MakeLowercase{\textit{et al.}}: Bare Demo of IEEEtran.cls for Computer Society Journals}
%



\IEEEtitleabstractindextext{%
\begin{abstract}
Transformer based large language models have achieved tremendous success. However, the significant memory and computational costs incurred during the inference process make it challenging to deploy large models on resource-constrained devices. In this paper, we investigate compression and efficient inference methods for large language models from an algorithmic perspective. Regarding taxonomy, similar to smaller models, compression and acceleration algorithms for large language models can still be categorized into quantization, pruning, distillation, compact architecture design, dynamic networks. However, Large language models have two prominent characteristics compared to smaller models: (1) Most of compression algorithms require finetuning or even retraining the model after compression. The most notable aspect of large models is the very high cost associated with model finetuning or training. Therefore, many algorithms for large models, such as quantization and pruning, start to explore tuning-free algorithms. (2) Large models emphasize versatility and generalization rather than performance on a single task. Hence, many algorithms, such as knowledge distillation, focus on how to preserving their versatility and generalization after compression. Since these two characteristics were not very pronounced in early large models, we further distinguish large language models into medium models and ``real'' large models. Additionally, we also provide an introduction to some mature frameworks for efficient inference of large models, which can support basic compression or acceleration algorithms, greatly facilitating model deployment for users.
\end{abstract}

}

\maketitle

\IEEEdisplaynontitleabstractindextext

%
\IEEEpeerreviewmaketitle




\section{Introduction}
\label{sec:intro}
\IEEEPARstart{L}{arge} language models (LLMs) has become an important and popular topic in the artificial intelligence field. Compared with previous language models, LLMs (\eg, ChatGPT, LLaMA, Claude) show much greater generalization capability for their unseen data. Furthermore, they even present many abilities that smaller models do not present (\ie, emergent abilities), such as multi-step reasoning and instruction following abilities. These progresses demonstrate great potentials of LLMs.

However, the forbidding memory and computational budgets in the inference process also prevent the deployment of LLMs. For example, a 10B models with \textit{float32} weights consumes 37GB memory, needless to say that the inference memory cost will further increase in a speed square to the sequence length. To deploy the models on resource constrained devices, or even mobile devices, many LLMs resort to model compression methods such as quantization to reduce the inference memory and computational cost.








Model compression for deep learning models is a field that appear much earlier than LLMs. It assumes that we have already a pre-defined (or even pretrained) model. Model compression devotes to reducing the memory and computational cost of the model in the inference process, so that the model can run on various resource-constrained devices. Algorithmically, common model compression methods include:
\begin{itemize}
    \item \textbf{Quantization} transforms \textit{float32} weights or activations into lower-bit float or integer numbers. Less bits means less memory requirement. Further, less bits may indicate higher parallelism and faster inference speed.
    \item \textbf{Pruning} devotes to removing unimportant components (\eg, neurons, layers, \etc) in a pre-designed model, thus reducing the memory and computational cost in the inference cost.
    \item \textbf{Knowledge distillation} introduces a pretrained large model as a teacher and transfers its knowledge to a new smaller model which is called a student model. Then, the smaller model will share nearly the same ability as the teacher and enjoy less memory and computational cost.
    \item \textbf{Compact architecture design} designs new operators with less cost to replace (often approximate) the cumbersome operators in the original model. For the Transformer models, self-attentions are the main targets and are often replaced with other operators.
    \item \textbf{Dynamic networks} treat each inference sample differently. The original model is a super-net, and each sample only selects a sub-structure of the super-net for inference. Mixture of experts (MoE) is a kind of dynamic inference.
\end{itemize}


Besides, the above methods can also be combined for further compression and speedup. Existing compression methods have provided us with important cornerstones and insights to compress LLMs. However, LLMs also bring many new challenges for model compression:

\begin{enumerate}
    \item Many previous model compression methods often require to finetuning models after compression. However, since the great budget to finetuning LLMs, researchers have to explore finetuning free or, at least, more efficient finetuning methods.
    \item Instead of handling one single task such as neural machine translation, large language models emphsize versatility and generalization across various tasks and unseen data, or even emergent abilities. Thus, large language models after compression require more careful validation of their versatility and generalization.
\end{enumerate}

To face these challenges, many compression methods specialized for LLMs are proposed. In this paper, we will give a comprehensive survey about these methods. To better present these methods, we further isolate language models around one billion or less parameters, such as BERT, GPT-2, and call them medium models, though they are usually taken as large language models. And Models with over one billion parameters, such as LLaMA, Claude, ChatGPT, and \etc, keep the name of large language models. The reasons is that medium models are less affected by the above two challenges, \ie, medium models are relatively easy to finetune, demonstrate fewer emergent abilities. As a result, many compression methods for medium models are still similar to those for smaller models.


The following sections are organized as: Some preliminaries will be introduced in Section~\ref{sec:preliminary}. Then, we will discuss pruning, knowledge distillation, quantization, compact architecture design and dynamic networks in Section~\ref{quant}, \ref{pruning}, \ref{knowledge_distillation}, \ref{compact_architecture_design}, \ref{dynamic_inference}, \ref{acceleration_framework}, respectively.

\section{Preliminaries} \label{sec:preliminary}
In this section, we will introduce some essential preliminaries about Transformer, large language models, parameter-efficient training, and \etc.

\subsection{Transformer}
\begin{figure}
    \centering
    \includegraphics[width=0.7\linewidth]{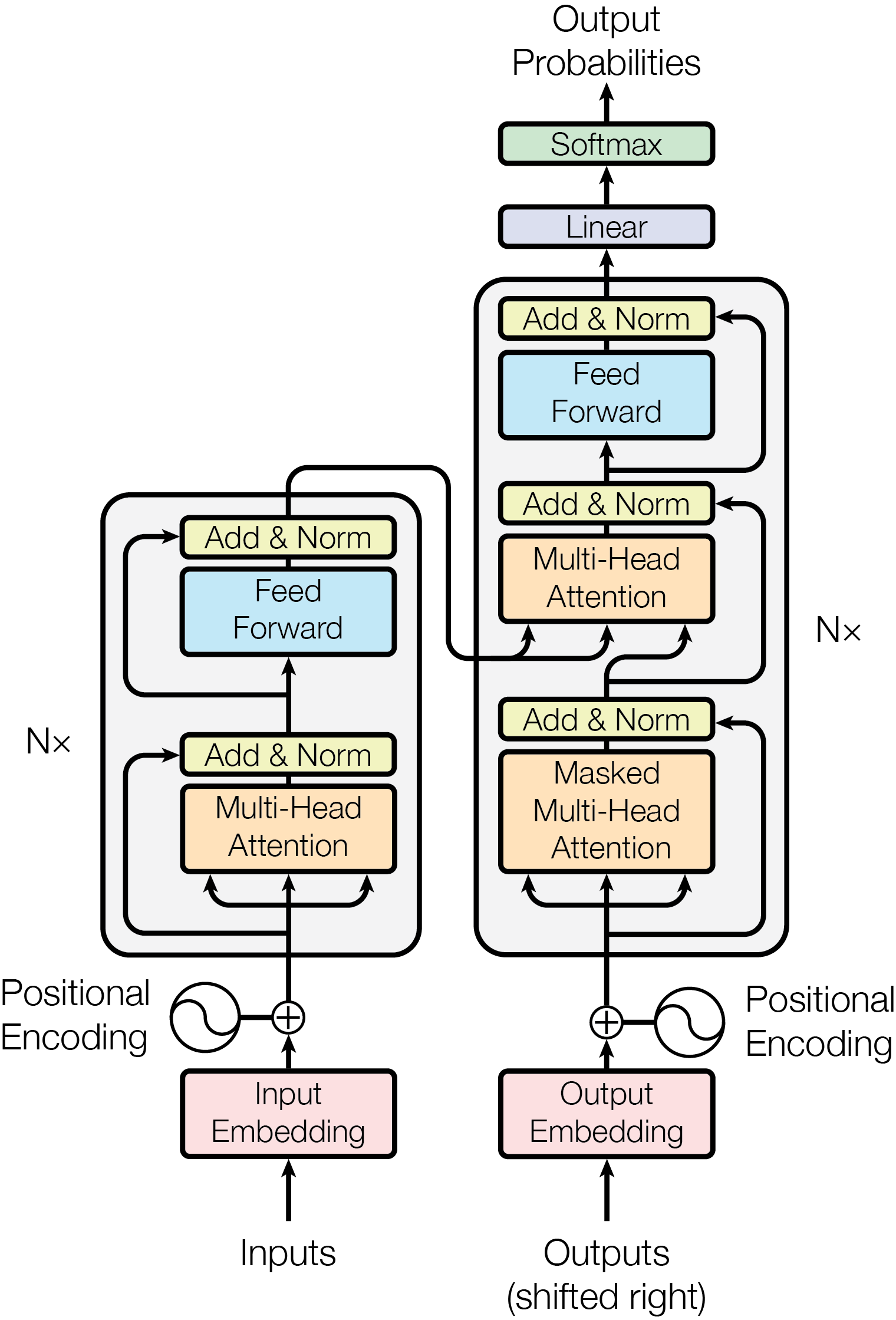}
    \caption{The Transformer architecture drawn from ~\cite{MoE-Transformer}.}
    \label{fig:intro}
\end{figure}

Transformer is first introduced in~\cite{MoE-Transformer}, which is employed for machine translation initially. Its fundamental structure is depicted in Fig.~\ref{fig:intro}. The input (a sentence) is often projected through an embedding layer to a sequence of vectors (called tokens) for a Transformer's input. Each Transformer block consists of an attention module and a multi-layer preceptron (MLP) module. 

\textbf{Attention.} For each token in the input sequence, it is first mapped (often with a linear function) into vectors of query ($Q$) and/or key-value pairs ($K$ and $V$). Then, the attention module can be described as mapping a query and a set of key-value pairs to an output. The output is computed as a weighted sum of the values, where the weight assigned to each value is computed by a compatibility function of the query with the corresponding key. The most common attention module is scaled dot-product function:
\begin{equation}
\label{equ:attention}
{\rm Attention} (Q, K, V) = {\rm softmax}(\frac{QK^T}{\sqrt{d_k}})V,   
\end{equation}
where the weight is computed through the dot-product of $Q$ and $K$, and $\sqrt{d_k}$ is a constant scaling factor.

\textbf{Multi-head Attention.} Further, instead of performing a single attention function with keys, values and queries, Transformer employs a multi-head attention~\cite{MoE-Transformer}, as shown in Fig.~\ref{fig:intro}. It maps input tokens into $h$ distinct queries, keys and values ($\{Q_i, K_i, V_i | i \in [1, h]\}$) with different linear layers. Then, the final output becomes:
\begin{equation}
\label{equ:multi-head}
\begin{aligned}
\operatorname{Muti-Head\ Attention} &= {\rm Concat}({\rm head_1, \cdots, head_h})W_o \\
{\rm head_i} &= {\rm Attention}(Q_i, K_i, V_i),
\end{aligned}
\end{equation}
where $W_o$ is a linear projection matrix.

\textbf{Encoder and docoder.} The first intial Transformer is for neural machine translation, which employs an encoder-decoder structure. The encoder first handles input sequnce (\eg, wirtten in the source language) independently, and the decoder takes the encoder's output as input and predicts the final output (\eg, the target language). There are two core differences between an encoder's and a decoder's attention module: (1) The encoder employs a full attention, where any two tokens in the input sequence are visible to each other. On the other hand, the decoder employs a single-direction attention. The reason is the decoder generates output tokens one by one, and each token can only seen output tokens before itself. (2) The encoder employs a self-attention module, that is, $Q, K, V$ are all from input tokens of the source language. In contrast, the decoder employs a cross-attention, where $K, V$ are from the encoder's output, while $Q$ is last output token of the decoder. As the development, in addition to the encoder-decoder models (\eg, T5~\cite{MoE-T5}), plenty of following language models also employ pure encoder structure (\eg, BERT~\cite{MoE-Bert}) and pure decoder structure such as GPT-series~\cite{radford2018improving,radford2019language,DBLP:conf/nips/Ouyang0JAWMZASR22} models.

While we brifly introduce some important concepts in Transformer, more subtle introduction can be seen in many previous surveys~\cite{DBLP:journals/corr/abs-2106-04554,DBLP:journals/corr/abs-2306-07303}.

\subsection{Medium/Large Language Models}
As the success of Transformer, more and more pure Transformer based language models emerge, and the parameters of models also increase. Though there is no specific threshold for large language models' scale of parameter, it is commonly accepted that ``large'' language models can be dated back from BERT~\cite{MoE-Bert} and GPT-1~\cite{radford2018improving}, which are both proposed in 2018. Their scales of parameter both reach several hundred million.

After that, more language models such as GPT-3~\cite{MoE-GPT3}, PanGu~\cite{DBLP:journals/corr/abs-2312-17276}, T5~\cite{MoE-T5}, CPM-2~\cite{DBLP:journals/aiopen/ZhangHZKGYQSJGQ21}, BLOOM~\cite{DBLP:journals/corr/abs-2211-05100}, OPT~\cite{DBLP:journals/corr/abs-2205-01068}, GLM~\cite{zeng2022glm}, PaLM~\cite{DBLP:journals/jmlr/ChowdheryNDBMRBCSGSSTMRBTSPRDHPBAI23}, QWen~\cite{DBLP:journals/corr/abs-2309-16609}, ERNIE~\cite{DBLP:conf/acl/ZhangHLJSL19}, LLaMA~\cite{MoE-Llama2}, and \etc are proposed. Besides scale of parameter, the most significant property that distinguish these models from the previous are their emergence. As proposed in~\cite{DBLP:journals/tmlr/WeiTBRZBYBZMCHVLDF22}, large language models utilize large-scale self-supervised pretraining to enable the models with many abilities (\ie, emergent abilities) that do not appear in smaller models, including multi-step reasoning, instruction following, program execution abilities, and \etc. For example, GPT-3, LLaMA and many other LLMs can solve few-shot tasks through in-context learning, and even zero-shot tasks. The breakout of large language models present the surprising abilities (called emergence) in solving a series of complex tasks over smaller models. 

To further emphasize this difference, we catorgorize language models over hundred millions of parameters into medium models and large models. Specifically, models with around one billion or less parameters are called medium models, while those with over one billion parameters are called large models.

\subsection{Parameter-efficient Finetuning (PEFT)}
As we discussed in the introduction, many model compression algorithms such as knowledge distillation and pruning require finetuning or even training for accuracy recovery after compression. Nevertheless, full-parameter finetuning or training is very costly for medium or large models. To this end, many parameter efficient finetuning (PEFT) algorithms are proposed. They devote to finetune as few parameters or epochs as possible to lower the finetuning cost. In this paper, model compression and acceleration algorithms in the inference stage (rather than training/finetuning stage) are mainly discussed, but we still supplement some PEFT algorithms in the Appendix~\ref{appendix}. 




\section{Quantization}
\label{quant}

\input{quantization}

\section{Pruning}
\label{pruning}
\input{pruning}

\section{Knowledge Distillation}
\label{knowledge_distillation}
\input{knowledge_distillation}

\section{Compact Architecture Design}
\label{compact_architecture_design}
\input{Operators}

\section{Dynamic Networks}
\label{dynamic_inference}
\input{dynamic_inference}

\section{Acceleration Framework}
\label{acceleration_framework}
\input{AccelerationFramework}



\section{Conclusions}

In this paper, we conducted a comprehensive investigation of compression and efficient inference for large language models from an algorithmic perspective, including quantization, pruning, distillation, compact architecture design, dynamic networks. Additionally, we introduced some popular compression and acceleration frameworks tailored for large language models. However, as we mentioned in the introduction, compression and acceleration of large models face more challenges compared to smaller models. While existing algorithms have made significant efforts to address these challenges, many algorithms still rely on frameworks designed for compressing small models, and challenges in compressing large models persist. In the future, further exploration is needed to develop more efficient and effective compression algorithms while ensuring the versatility and generalization of large models.

\section*{Contributors}
Wenxiao Wang is responsible for this paper's overall structure, content arrangement, the writing of Section~\ref{sec:intro} and Section~\ref{sec:preliminary}, and refinement of each section in this paper. Wei Chen, Yongliu Long, Zhengkai Lin and Liye Zhang are responsible for the surveys and writing of quantization (Section~\ref{quant}), pruning (Section~\ref{pruning}), dynamic networks (Section~\ref{dynamic_inference}), and knowledge distillation (Section~\ref{knowledge_distillation}), respectively. Yicong Luo is responsible for surveys and writing compact architecture design and acceleration framework (Section~\ref{compact_architecture_design} and Section~\ref{acceleration_framework}). All co-first authors are listed in alphabetical order of their surnames. Binbin Lin, Deng Cai, and Xiaofei He participate in the comprehensive discussion and provide many great insights.

\ifCLASSOPTIONcaptionsoff
  \newpage
\fi



\bibliographystyle{IEEEtran}
\bibliography{reference}

\appendices
\section{Parameter-efficient Finetuning (PEFT)}
\label{appendix}
\input{PEFT}

\end{document}

%% file: quantization.tex
\begin{figure} [t!]
    \centering
    \subfloat[\label{fig:quan-eg-01} Illustration of uniform quantization process]{
        \includegraphics[width=\linewidth]{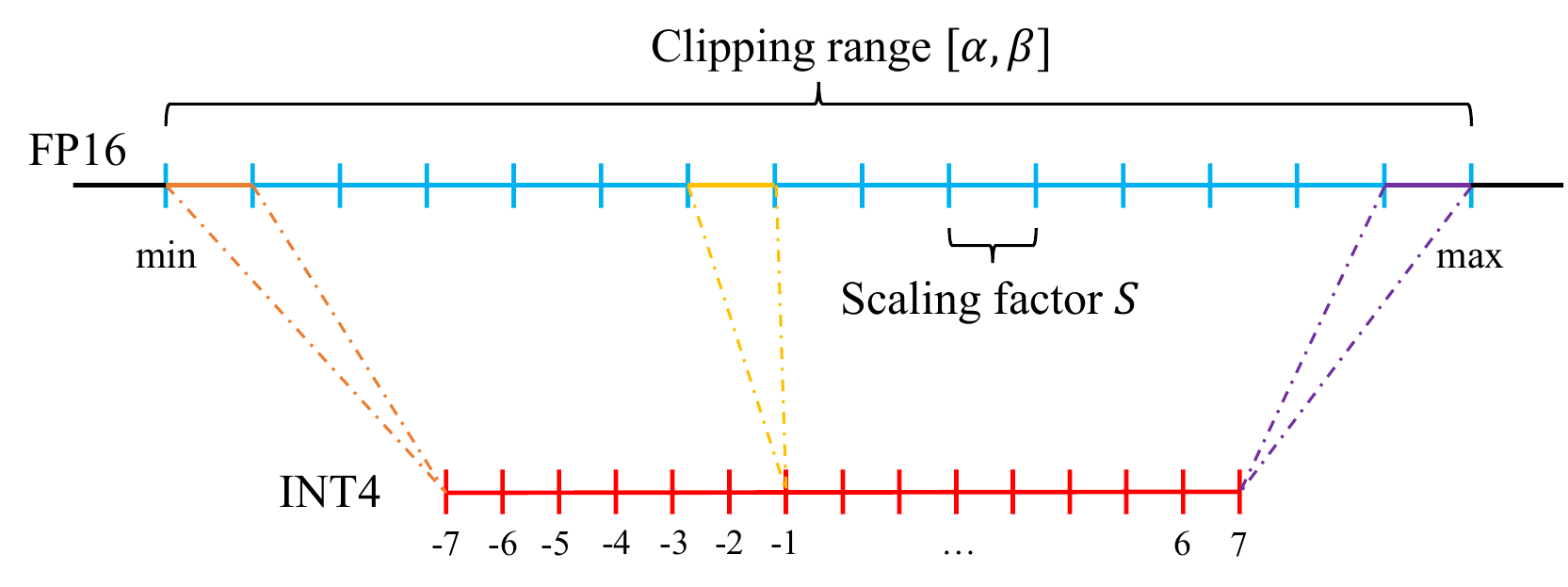}}
    \\
    \subfloat[\label{fig:quan-eg-02} Quantization and De-quantization of a FP16 tensor]{
        \includegraphics[width=.9\linewidth]{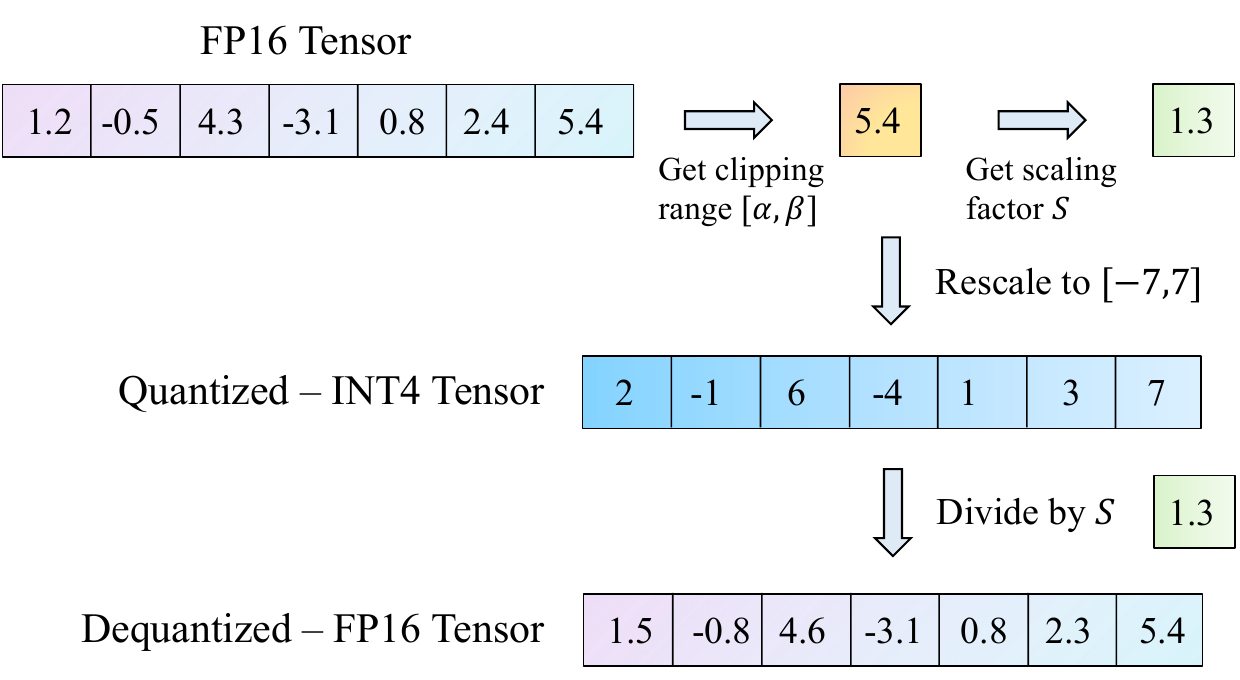}}
    \caption{(a) Uniform quantization separates a real-valued range into \textit{uniform, finite} intervals and then maps real values within the same interval to the same integer. (b) An FP16 tensor is quantized into INT4 format and then dequantized back into FP16.}
    \label{fig:quan-example}
\end{figure}

Quantization refers to the process of mapping input values in a \textit{large (often continuous)} set to output values in a \textit{small (often finite)} set (see Fig.~\ref{fig:quan-example} for an example). It is the most straightforward method to cut down memory cost and improve inference speed for LLMs, especially on hardware with support for fast operation of low-bit datatypes (\eg, INT4). It should be noted that quantization has achieved impressive success in both neural network training \cite{yang2023dynamic} and inference, while the focus of this survey is only the inference part.

Quantization methods have several advantages over other compression methods, such as pruning and distillation. 
1) \textit{High compression ratio}: quantizing the weights in LLMs from 32-bit float to 4-bit integer could drastically compress the model size to approximately $1/8$, essential for memory-bound\footnote{"memory-bound" means that the transfer between the device and global memory nearly reaches the limitation or fetching data from the memory is the bottleneck of the whole process. On the contrary, the "compute-bound" process spends most of its time calculating and not accessing memory.\label{foot:bound}} processes like LLM inference. 
2) \textit{Low cost}: a bunch of quantization methods doesn't require re-training the entire LLMs, making it more affordable to researchers with limited computing resources. 
3) \textit{High flexibility}: quantization is compatible with most other compression methods, which introduces exceptional opportunities for further improving the performance.

To help readers better understand quantization methods, we will first introduce the standard quantization method and some basic concepts in Subsection \ref{quan: basic-concepts}. Then, in Section \ref{quan: before-llm}, we will briefly summarize some of the most important works for medium-size Language models (\eg, BERT, GPT-2, \etc.) before the emergence of LLMs. Section \ref{quan: ptq-llm} and Section \ref{quan: qat-llm} covers recent advances in quantization methods that focus on LLMs inference. 
Considering the pains in re-training a model with tens of billions of parameters, we generally divide LLM quantization methods into two parts based on whether the technique needs re-training. Methods without re-training (\ie, \textit{post-training quantization, PTQ}) are discussed in Section \ref{quan: ptq-llm} while methods that demand re-training (\ie, \textit{quantization-aware training, QAT}) is discussed in Section \ref{quan: qat-llm}. Finally, in Section \ref{quan: advance-topic}, we discuss some advanced topics showing potential for future research but not covered in previous sections.

\subsection{Basic Concepts}
\label{quan: basic-concepts}

Quantization has a history that is much longer than neural networks, and specific quantization methods vary a lot. To give readers a clear grasp of the diverse quantization concepts, we will first introduce the standard \textit{uniform} quantization and the corresponding dequantization process. After that, we explain several fundamental concepts frequently mentioned in different quantization methods.

1) \textit{Uniform quantization.} The most basic form of quantization is to separate a real-valued range into \textit{uniform, finite} intervals (\eg, $2^b$ intervals for $b$-bit integer) and then map real values within the same interval to the same integer. The formula of such a process is as follows:
\begin{equation}
  \label{quan: eq-1}
  Q(r) = \mathrm{ROUND}(\frac{r}{S}) + Z
\end{equation}
where $Q(\cdot)$ is the quantization operator, $r$ is the real values to be quantized, $S$ is a real valued scaling factor, $Z$ is an integer zero point and $\mathrm{ROUND}(\cdot)$ is a rounding operation (\eg , round to nearest). This method is known as \textit{uniform quantization} since the length of each interval is the same (equal to scale factor $S$) and the quantized values are uniformly spaced (\eg, integers $0, 1, 2, \dots$).

The operation to recover real values from quantized values is called \textit{dequantization}:
\begin{equation}
  \label{quan: eq-2}
  \tilde{r} = S \cdot (Q(r) - Z)
\end{equation}
It should be noted that the recovered values $\tilde{r}$ may be different from the original real values $r$ due to the information loss introduced by $\mathrm{ROUND}(\cdot)$ function.

2) \textit{Non-uniform quantization.} The counterpart of uniform quantization is \textit{non-uniform quantization}, where quantized values are not necessarily uniformly spaced, and the length of intervals can be different. The general formula for non-uniform quantization is:
\begin{equation}
  \label{quan: eq-3}
  Q(r) = Q_i, \ \mathrm{if} \ r \in [\Delta_i, \Delta_{i+1})
\end{equation}
where $Q_i$ is the candidate quantized values called \textit{quantization levels}, $\Delta_i$ and $\Delta_{i+1}$ defines an interval in which real values would be mapped to corresponding $Q_i$.

Given a fixed bit-width, non-uniform quantization can often achieve higher accuracy and lower quantization error than its uniform counterpart, as the weights of neural networks are generally not uniformly distributed. However, non-uniform methods may suffer low computation efficiency because these methods usually involve a time-consuming lookup operation that is not directly compatible with parallel computation hardware like GPU.

3) \textit{Clipping range and calibration.} An important factor associated with uniform quantization is \textit{clipping range} $[\alpha, \beta]$ so that real values lower than $\alpha$ or higher than $\beta$ would be clipped to $\alpha$ or $\beta$ respectively. The clipping range also directly influences the scaling factor $S$ in uniform quantization:
\begin{equation}
  \label{quan: eq-4} S = \frac{\beta - \alpha}{2^b - 1}
\end{equation}
In general, a wider clipping range results in fewer outliers requiring clipping in the input data. However, this comes at the cost of a larger scale factor, as shown in Equation \eqref{quan: eq-4}. Consequently, the quantization error for real values within the clipping range would also be larger.

Choosing the clipping range is referred to as \textit{calibration}.
Common choices of calibration involve using min/max values (\ie,
$-\alpha = r_{min}, \beta = r_{max}$), using absolute max values (\ie, $-\alpha=\beta=max(|r|)$) or minimizing the information loss (\ie, KL divergence) between the real values and quantized values.

4) \textit{Symmetric/Asymmetric quantization.} When the clipping range $[\alpha, \beta]$ is symmetric with respect to $0$ ($\alpha + \beta = 0$ and $Z=0$), then corresponding method is often referred to as \textit{symmetric quantization}; otherwise \textit{asymmetric quantization}.

The clipping range for asymmetric quantization is more compact, which is especially important for activations in neural networks whose range may be significantly imbalanced (\eg, after ReLU, all activations become non-negative). Symmetric quantization, however, is more compute-efficient at the dequantization step.

5) \textit{Quantization Granularity.} A categorization criterion specified for quantization of neural network is the \textit{granularity}, 
which corresponds to which weights/activations are quantized together and share quantization parameters. We list typical granularity from coarse to fine as follows:
\begin{itemize}
  \item \textit{Layer-wise}: Weights of all filters in a layer for convolution layers or the full weight matrix for linear layers are quantized together.
  \item \textit{Channel-wise}: Weights of a single filter for convolution layers are quantized 
  together.
  \item \textit{Row/Column-wise}: Weights of a single row/column of weight matrix for linear layers are quantized together.
  \item \textit{Token-wise}: Activations for each token are quantized together.
  \item \textit{Group-wise}: Several consecutive real values in weights or activations are viewed as a group and quantized together. The group size is typically small (\eg, $64$ consecutive real values).
\end{itemize}
In general, finer granularity divides model weights or activations into smaller groups, which can reduce quantization errors. However, finer granularity requires storing more quantization parameters and introduces higher quantization overhead during computation.

Note that different works may use different notations. For clarity, in this survey, we set input $\mathbf{X} \in \mathbb{R}^{N \times D_{\mathrm{in}}}$, weight matrix $\mathbf{W} \in \mathbb{R}^{D_{\mathrm{in}} \times D_{\mathrm{out}}}$, where $N$ is the batch size, $D_{\mathrm{in}}$ and $D_{\mathrm{out}}$ are input and output dimensions respectively. A linear layer can be written as $\mathbf{Y} = \mathbf{X}\mathbf{W} \in \mathbb{R}^{N \times D_{\mathrm{out}}}$.

6) \textit{Post-Training Quantization/Quantization-Aware Training.} An effective way to reduce quantization error is through training. Quantization methods can either require re-training after quantization or not:
\begin{itemize}
  \item \textit{Post-Training Quantization, PTQ}: methods without re-training, quantized models can be directly used in inference. 
  \item \textit{Quantization-Aware Training, QAT}: methods with re-training, which helps recover the error introduced by quantization. 
\end{itemize}
Perturbation to parameters of a trained model caused by quantization may push the model away from the point where it had converged in floating-point-precision training. QAT addresses this issue through either simulating the quantization process in re-training (\eg, inject nodes in the computation graph that simulate weight quantization) or using additional parameters to finetune the quantized model (\eg, combined with Adapters or LoRA) so that the model can learn to converge to a point which will have a better performance after quantization. However, re-training of full LLM with billions of parameters has an unbearable cost to most researchers. Hence, PTQ methods are also extensively studied to achieve better performance without introducing additional computation budgets through specific non-training methods (\eg, Optimal Brain Surgery).

Generally speaking, the rounding operation in quantization is a non-differentiable process, so QAT may seem impossible at first sight. However, researchers found that a simple method called Straight Through Estimator (STE) works well under most circumstances, ignoring the rounding operation and approximating it with an identity function.

7) \textit{Static/Dynamic quantization.} A key difference between quantizing the weights and activations of a neural network lies in the fact that weights are mostly fixed during inference so that the statistics used in clipping range calibration can be computed statically. It's not the case for activations since its range and statistics are unknown until runtime. Thus, activation quantization can be divided into two categories.
\begin{itemize}
  \item \textit{Static quantization} refers to the methods whose quantization parameters are pre-calculated before inference using some calibration inputs to find typical activation statistics or learned jointly during neural network training. 
  \item \textit{Dynamic quantization} calculates quantization parameters dynamically at runtime, which is generally more accurate than its static counterpart but can have high overhead for calculating required statistics (\eg, min, max, \etc.).
\end{itemize}

8) \textit{Simulated/Integer-Only quantization.} Still, another category criterion is whether quantized values are used for actual operations (\eg, matrix multiplication). For \textit{simulated quantization} (also called \textit{fake quantization}), the weights and activations are only stored in low-bit datatypes (\eg, INT4) and have to be \textit{dequantized} to high-bit datatypes (\eg, float16) to carry out actual operations. For \textit{Integer-only quantization}, the operations can be carried out using low-bit datatypes. Simulated quantization can reduce the memory cost and data-moving time of neural networks, which is helpful since several works have shown that LLM inference is \textit{memory-bound} rather than \textit{compute-bound}\footref{foot:bound}. In contrast, integer-only quantization can further enjoy the acceleration of efficient low-bit operations supported by specific hardware.

9) \textit{Weight-only/Weight + Activation quantization.} Whether the quantization objective is only weights or both weights and activations. Previous work \cite{yao2023zeroquantv2} found that activation quantization is generally more susceptible to weight quantization, so weight-only quantization can reach lower bit-width. However, quantized weights must be dequantized before multiplying with activations, so weight-only quantization can not be integer-only and will introduce additional computation overhead during inference.

We've briefly covered some of the most essential concepts in quantization. These concepts are universally applied to all neural network quantization, and each specific method may suit several different concepts (\eg, a \textit{uniform symmetric dynamic layer-wise simulated} quantization method for LLMs). We categorize the main quantization methods for LLMs according to these basic concepts in TABLE~\ref{quan: summarization}.

\subsection{Quantization Methods for Medium-Size Language Models}
\label{quan: before-llm}

For ease of expression, we refer to models with sizes smaller than or close to 1B as medium-size language models, represented by BERT, GPT-2, and BART.





Quantization methods for medium-size language models \cite{bondarenko2021understanding} mainly adopt the QAT 
framework instead of PTQ, as the cost of re-training such a model is relatively acceptable. The improvement in evaluation metric (\eg, accuracy) brought by re-training is significant, especially under extreme low-bit settings (\eg, $1$-bit or $2$-bit quantization). As a result, we will first introduce mainstream methods, 
\ie, QAT methods, for medium-size language models and then cover the PTQ methods.


\subsubsection{QAT for Medium-Size Language Models}

Early works aim at quantizing weights of BERT-like models into INT8. Q8BERT \cite{zafrir2019q8bert} applies the basic QAT framework from \cite{jacob2018quantization} to quantize both weights and activations of BERT into 8-bits without significant reduction in model performance.  

Some works enable quantization into bit-width lower than 8-bit using more complicated methods \cite{shen2020q, piao2022sensimix, zhang2020ternarybert, qin2021bibert, zhao2021automatic, zhao2020investigation, wang2021exploring}. 
For example, Q-BERT \cite{shen2020q} maintains 8-bit activations and mixed-precision weights down to 2/3-bits. It uses the Hessian matrix to determine the bit-width for weights of each layer so that more aggressive quantization is performed for layers with smaller top eigenvalues of the corresponding Hessian. 
Further, TernaryBERT \cite{zhang2020ternarybert} restricts its weights to {-1, 0, +1}, using only 2 bits, and employs 8-bit activations. Knowledge distillation is adopted to overcome performance degradation by minimizing the sum of mean-square error (MSE) of the activations and attention scores between the original and quantized model.
Following TernaryBERT, BinaryBERT \cite{bai2021binarybert} pushes the limit of BERT quantization to weight binarization, \ie, restricts weights in $\{-\alpha, +\alpha\}$. The authors propose to initialize BinaryBERT by equivalently splitting from a half-sized TernaryBERT to inherit the good performance of the ternary one.
In addition, BiBERT \cite{qin2021bibert} is a full binarization of BERT (\ie, 1-bit weight, embedding, and activation). The authors identify the severe performance degradation of the full binary model comes from information degradation and optimization direction mismatch. A Bi-Attention structure and a DirectionMatching Distillation (DMD) scheme are proposed accordingly to retain most of the ability of the original BERT.

Some works enable an automatic balance between the model performance degradation and quantization bit-width.
Zhao et al. \cite{zhao2021automatic} leverages a \textit{Differentiable Neural Architecture Search} approach to assign precision for parameters automatically. In detail, the weights and the bit assignment of weights are optimized alternatively under an objective function that combines the cross entropy loss with the penalty of model size. The optimization process aims to obtain a set of bit assignments for each group of parameters close to optimal.

\subsubsection{PTQ for Medium-Size Language Models}

PTQ methods are carefully designed so they generally do not require extra finetuning or re-training to compensate for quantization errors. 
GOBO \cite{zadeh2020gobo} quantizes the vast majority of weights that comply with Gaussian distribution into 3 bits using non-uniform quantization (\ie, clustering) and saves a few outlier weights separately in FP32. 
I-BERT \cite{kim2021bert} designs integer-only approximation methods for specific non-linear functions (\eg, GeLU, Softmax, LayerNorm) to enable end-to-end integer-only BERT inference without any ﬂoating point calculation. 
Dai et al. \cite{dai2021vs} use finer granularity to reduce quantization error. In detail, the authors quantize weights and activations into 4 bits using group-wise quantization (\eg, $16 \sim 64$ consecutive weights as a group). A calibration set is used to determine the scaling factor for each group. 

Furthermore, it should be noted that the quantization parameters obtained by elaborately tailored PTQ methods can, in general, be a good initialization point for re-training in QAT methods.

\subsubsection{Quantize Generative Medium-Size Language Models}

Despite the success of quantization approaches for BERT-like models mentioned above, attempts to quantize generative language models (\eg, GPT, BART) was scarce before the emergence of generative LLMs \cite{li2021short}. The critical difference is that quantization error accumulates in the token-by-token generation process, so quantizing generative language models is generally a more complex problem.

According to Tao et al. \cite{tao2022compression}, applying quantization methods that are designed for BERT-like models directly to generative language models is hindered by homogeneous word embedding and varied distribution of weights. Homogeneous word embedding refers to the problem where the word embeddings of generative language models become less distinguishable from each other after quantization. On the other hand, varied distribution of weights means that the weights of the model are highly skewed with outliers. To tackle these challenges, the authors propose two solutions: token-level contrastive distillation and module-dependent dynamic scaling.
DQ-BART \cite{li2022dq} uses the QAT framework and a distillation training objective to distill and quantize a sequence-to-sequence model, \ie, BART, jointly. DQ-BART adopts the standard symmetric uniform quantization as shown in Equation \eqref{quan: eq-1} and sets the training objective as minimizing the differences of the output logits, attentions, and hidden states between the quantized and distilled low-precision student model and the full precision teacher model.

In this section, we only briefly cover the most important works done on medium-sized language models. For a more detailed summarization of quantization methods for medium-sized language models, we refer interested readers to \cite{gholami2022survey, xu2023survey}.

\begin{table*}[!t]
  \renewcommand{\arraystretch}{1.3}
  \caption{Detailed category of several strong baseline quantization methods for LLMs. \checkmark means that a quantization method belongs to a specific category, $\times$ vice versus, and $\circ$ means a quantization method can be used in both circumstances. For methods that work in different bit widths, we report the lowest effective bit width. For a detailed explanation of each category, please refer to Subsection \ref{quan: basic-concepts}}
  \label{quan: summarization}
  \centering
  \resizebox{\linewidth}{!}{
  \begin{tabular}{@{}lcllcccccc@{}}
    \toprule
    \bfseries Method       & 
    \bfseries \#Bits       &
    \bfseries Weight       &
    \bfseries Activation   &
    \bfseries Uniform      & 
    \bfseries Symmetric    & 
    \bfseries Static       &
    \bfseries Re-Training  &
    \bfseries Zero-shot    &
    \bfseries Integer-Only \\ \midrule

    AWQ\cite{lin2023awq} & 
    $3$               &
    group-size        &
    $-$               &
    \checkmark        &
    $\times$          &
    \checkmark        &
    $\times$          &
    $\times$          &
    $\times$          \\

    OPTQ/GPTQ\cite{frantar2022optq} & 
    $3$               &
    column-wise       &
    $-$               &
    \checkmark        &
    $\times$          &
    \checkmark        &
    $\times$          &
    $\times$          &
    $\times$          \\ \midrule
    
    \texttt{LLM.int8()}\cite{dettmers2022llm} & 
    $8$               &
    column-wise       &
    row-wise          &
    \checkmark        &
    $\circ$           &
    $\times$          &
    $\times$          &
    \checkmark        &
    \checkmark        \\

    ZeroQuant\cite{yao2022zeroquant} & 
    $4$               &
    group-wise        &
    tokenwise-wise    &
    \checkmark        &
    \checkmark        &
    $\times$          &
    $\circ$           &
    $\circ$           &
    \checkmark        \\

    SmoothQuant\cite{xiao2023smoothquant} & 
    $8$               &
    layer-wise        &
    layer-wise        &
    \checkmark        &
    \checkmark        &
    $\circ$           &
    $\times$          &
    $\times$          &
    \checkmark        \\ \midrule

    LLM-QAT\cite{liu2023llm} & 
    $2$               &
    column-wise       &
    tokenwise-wise    &
    \checkmark        &
    \checkmark        &
    $\times$          &
    \checkmark        &
    $\times$          &
    $\times$          \\
    
    INT2.1\cite{chai2023int2} & 
    $2$               &
    column-wise       &
    $-$               &
    \checkmark        &
    $\times$          &
    \checkmark        &
    \checkmark        &
    $\times$          &
    $\times$          \\

    QLoRA\cite{dettmers2023qlora} & 
    $4$               &
    column-wise       &
    $-$               &
    $\times$          &
    \checkmark        &
    \checkmark        &
    \checkmark        &
    $\times$          &
    $\times$          \\ \bottomrule
  \end{tabular}
  }
\end{table*}

\subsection{Post-Training Quantization for LLMs}
\label{quan: ptq-llm}

The past few years have witnessed a remarkable surge in post-training quantization methods (PTQ) for LLMs. This is partly because PTQ doesn't involve LLMs' prohibitively expensive re-training process, so it's a more feasible direction for most researchers.

Further, we roughly divide PTQ works for LLMs into two categories: \textit{weight-only quantization} and \textit{weight + activation quantization}. We'll discuss works related to these categories respectively in the following parts.

\subsubsection{Weight-Only Quantization.}

In this part, we focus on the problem of only quantizing the weights (but \textit{not} the activations) of LLMs. Generally, weight-only quantization belongs to simulated quantization methods; the weights are only stored in low-bit datatype and need to be \textit{dequantized} before real computation. This means such methods can decrease the overall size of LLMs and decrease the time to move weights between memories but cannot enjoy the accelerated low-bit operation supported by specific hardware.




While the previous subsections have discussed various methods that can be used to quantize medium-size language models, LLMs present additional challenges due to their unique characteristics. These challenges include: 
\begin{enumerate}
  \item  LMs rely heavily on memory during the inference process, especially when the inference batch size is small \cite{kim2023squeezellm}. This makes it crucial to minimize memory usage and optimize data transfer between different storage devices.
  \item The activation patterns of LLMs are distinct, which poses a challenge when applying quantization methods that work well for medium-sized language models. Systematic outliers \cite{dettmers2022llm} are one such unique property of LLM activations that hinder the direct application of such methods for weight-only quantization of LLMs.
\end{enumerate}

Some works directly apply uniform, round-to-nearest quantization to LLMs with minor modifications \cite{yao2023zeroquantv2, zeng2022glm, kim2023finequant}.
ZeroQuant-V2 \cite{yao2023zeroquantv2} quantize OPT and BLOOM. It shows that using 16-bit activations and directly quantizing weights of these models to 8-bit integers using row-wise symmetric quantization results in negligible perplexity degradation, while 4-bit weight-only quantization witnesses a significant performance drop. To further push the limit of low-bit quantization, ZeroQuant-V2 \cite{yao2023zeroquantv2} propose the Low-Rank Compensation (LoRC) method, which approximates the error $\mathbf{E}$ between the original weight matrix $\mathbf{W}$ and the quantized weight matrix $\mathbf{\hat{W}}$ using a storage-efficient low-rank matrix $\mathbf{\hat{E}}$ so that $\mathbf{\hat{W} + \hat{E}}$ would be a better approximation of the original weight $\mathbf{W}$. 
However, Zeng et al. \cite{zeng2022glm} found that GLM-130B can be directly quantized into 4-bit weights using row-wise symmetric quantization with negligible performance degradation, which is evaluated by zero-shot accuracy on the LAMBADA dataset. The authors ascribe the appealing 4-bit quantization property of GLM-130B to its weight distributions being well-shaped and not skewed compared to GPT-style models like OPT and BLOOM.

Another line of research considers non-uniform methods in weight-only quantization of LLMs. The critical insight lies in the fact that the weight distribution of LLMs after training is non-uniform, so it makes sense to let interval $[\Delta_i, \Delta_{i+1})$ in Equation \eqref{quan: eq-2} also be non-uniform to push the quantization even to lower bit-width. 
LUT-GEMM \cite{park2022nuqmm} (also known as nuQmm) extends a non-uniform quantization method, binary-coding quantization (BCQ) \cite{rastegari2016xnor}, which factorizes the full-precision parameters into binary parameters and a separate set of scaling factors. The authors add a bias term to conventional BCQ methods to increase the representational capacity and use group-wise quantization to enable a tradeoff between the compression ratio and model performance. 
SqueezeLLM \cite{kim2023squeezellm} verifies that the LLM inference is memory-bound with extremely low arithmetic intensity relative to other neural networks. Besides, SqueezeLLM adopts sensitivity-based k-means centroids as the quantized weight values for non-uniform quantization (see $X_i$ in Equation \eqref{quan: eq-3}). The sensitivity-based k-means method approximates the Hessian matrix of weights as the sensitivity, highlighting the importance of minimizing perturbations for weights with large Hessian values. SqueezeLLM has better perplexity than standard uniform quantization methods while achieving around $2 \times$ speedup compared to the FP16 baseline. 
Dettmers et al. \cite{dettmers2023qlora} propose a new NormalFormat (NF) datatype, which can also be viewed as non-uniform quantization. The NF datatype builds on Quantile Quantization \cite{dettmers20218}, an information-theoretically optimal data type that ensures each quantization interval has an equal number of values assigned from the input tensor. The authors utilize that pre-trained neural network weights usually have a zero-centered normal distribution with standard deviation $\sigma$, thus can be normalized to the standard normal distribution $N(0, 1)$ by scaling $\sigma$. $k$-bit NormalFormat use $k$-bit Quantile Quantization on standard normal distribution $N(0, 1)$ to find its quantized values $Q_i$ (See Equation \eqref{quan: eq-3} for definition of $Q_i$). In practice, weights to be quantized are rescaled to range $[-1, 1]$ and then round to nearest quantized values $X_i$ \ie, round-to-nearest (RTN) methods.

Above are \textit{zero-shot methods} that only consider the minimize the difference between the original weight matrix $W$ and the quantized weight matrix 
$Q(\textbf{W})$, \ie , to minimize the quantization error of weight matrix 
$\mathrm{argmin}_{\hat{\textbf{W}}} ||\textbf{W} - Q(\textbf{W})||$. However, 
considering the high non-linearity of neural networks, a small distance in weight space doesn't necessarily mean a small difference between the output of the original and quantized models. Thus, if given a small set of typical examples $C$, called \textit{calibration set}, there are some \textit{one-shot methods} \cite{lee2023flexround, frantar2022optq, chee2023quip, lin2023awq} consider to optimize the difference between the output activations of the original and quantized layers:
\begin{equation}
  \label{quan: eq-5}
  \mathrm{argmin}_{\hat{\textbf{W}}} ||\textbf{XW} - \textbf{X}Q(\textbf{W})|| 
  \quad \mathrm{for} \ \textbf{X} \in C
\end{equation}

A typical work of one-shot methods for weight-only LLM quantization is GPTQ (also known as OPTQ) \cite{frantar2022optq}, which is built on an adaptive rounding method called Optimal Brain Quantization (OBQ) \cite{frantar2022optimal}. OBQ handles each row of the weight matrix independently, quantizing one weight at a time while updating all not-yet-quantized weights to compensate for the error incurred by quantizing a single weight. However, OBQ is not explicitly designed for LLMs and can be slow and inaccurate in practice. To fix these problems, GPTQ quantizes weights of all rows in parallel to improve efficiency, uses lazy batch updates to achieve a higher compute-to-memory ratio in the quantization process, and uses Cholesky reformulation to help numerical stability. GPTQ can quantize OPT-175B or BLOOM-176B in around 4 hours on a single NVIDIA A100 GPU with these modifications. Further, GPTQ can provide reasonable accuracy under extreme quantization where weights are quantized to 2-bit or lower. 
QuIP \cite{chee2023quip} defines a family of adaptive rounding methods for optimizing the Equation \eqref{quan: eq-5} and defines the optimal method within the pre-defined family of methods, called LDLQ. LDLQ uses the LDL decomposition of the second-moment matrix of vectors in the calibration set to find the optimal way to update the not-yet-quantized weights. The authors show that GPTQ is a particular case of LDLQ. Further, QuIP proposes incoherence processing that can transform the weight matrix into a more suitable form for quantization. Combining LDLQ and incoherence processing, QuIP is the first LLM quantization method that has viable results on 2-bit weight-only quantization. 
AWQ \cite{lin2023awq} shows that preserving only $0.1\%$ channels corresponding to significant activation in FP16 and quantizing the rest of the weight matrix can contribute to much better model performance, meaning weights are not equally important. Further, AWQ intends to reduce the quantization error for the essential weights without using mixed-precision, which is hard to implement efficiently. This is achieved by activation-aware scaling, which automatically finds a per-(input) channel scaling ratio $\mathbf{s}$ with an objective similar to Equation \eqref{quan: eq-5} \ie, $\mathrm{argmin}_{\mathbf{s}} ||(\mathbf{s}^{-1}\cdot \mathbf{X}) Q(\mathbf{W} \cdot \mathbf{s}) - \mathbf{XW}||$ such that the salient weights with high scaling factor can be better represented. In contrast, non-salient weights will not be neglected.
OWQ \cite{lee2023owq} is an advancement over OPTQ that significantly enhances the quality of the quantized model. It uses a mixed-precision quantization scheme, which applies higher precision to the weights susceptible to quantization caused by activation outliers. The sensitive weights are identified using a sensitivity-based method similar to \cite{kim2023squeezellm}.

There are also some studies focusing on rounding criteria in quantization.
SignRound \cite{cheng2023optimize} suggests that as the bit-width of quantization decreases, the quantization grid broadens, thus emphasizing the importance of up and down rounding. It extends previous work \cite{nagel2020up} to learn weight rounding with signed gradient descent and can achieve good results within 400 optimization steps.

\begin{table*}[]
    \renewcommand{\arraystretch}{1.3}
    \caption{The following table shows the perplexity of various strong baseline quantization methods for LLMs on Wikitext-2 \cite{merity2016pointer} and C4 \cite{raffel2020exploring}. Please note that the intention is not to compare the perplexity after quantization directly, as different quantization methods may perform best on different models with different scales. This table only serves as a rough comparison of the effects of different quantization methods. We strongly recommend readers refer to the original papers for detailed results in different settings. The reported results are derived from the original papers, except for QLoRA, whose result is derived from LoftQ \cite{li2023loftq}.}
     \label{quan: ppl-comparison}
    \centering
    \begin{tabular}{@{}clcclccc@{}}
    \toprule
    
    \multirow{2}{*}{\textbf{Dataset}}    & 
    \multirow{2}{*}{\textbf{Method}} & 
    \multirow{2}{*}{\makecell[c]{\#\textbf{Bits}\\\textbf{Weights}}} & 
    \multirow{2}{*}{\makecell[c]{\#\textbf{Bits}\\\textbf{Activations}}} & 
    \multirow{2}{*}{\textbf{Model}} & 
    \multicolumn{2}{c}{\textbf{Perplexity} ($\downarrow$)} & 
    \multicolumn{1}{l}{\multirow{2}{*}{\textbf{Speedup}}} \\ \cmidrule(lr){6-7}
    
    &                         
    &                                 
    &                                     
    &                        
    & \textbf{FP16 Model}    
    & \textbf{Quantized Model}  & 
    \multicolumn{1}{l}{} \\ \midrule
    
    \multirow{7}{*}{Wikitext-2} & 
    AWQ\cite{lin2023awq}  & 
    3                     & 
    16                    & 
    OPT-66B               & 
    10.09                 & 
    10.46                 & 
    1.85$\times$          \\
                                
    & OPTQ/GPTQ\cite{frantar2022optq} & 
    3                     & 
    16                    & 
    OPT-175B              & 
    8.34                  & 
    8.68                  & 
    3.24$\times$          \\
    
    & ZeroQuant\cite{yao2023zeroquantv2} & 
    4                     & 
    8                     & 
    BLOOM-176B            & 
    8.11                  & 
    8.33                  & 
    -                     \\
    
    & SmoothQuant\cite{xiao2023smoothquant} & 
    8                     & 
    8                     & 
    LLaMA-65B             & 
    6.17                  & 
    6.20                  & 
    1.56$\times$          \\
    
    & LLM-QAT\cite{liu2023llm} & 
    4                     & 
    8                     & 
    LLaMA-30B             & 
    7.00                  & 
    7.50                  & 
    -                     \\
    
    & INT2.1\cite{chai2023int2} & 
    2                     & 
    16                    & 
    LLaMA-7B              & 
    5.08                  & 
    8.74                  & 
    -                     \\
    
    & QLoRA\cite{dettmers2023qlora} & 
    3                     & 
    16                    & 
    LLaMA-13B             & 
    5.12                  & 
    5.22                  & 
    -                     \\ \midrule
    
    \multirow{5}{*}{C4}         
    & OPTQ/GPTQ\cite{frantar2022optq} & 
    3                     & 
    16                    & 
    OPT-175B              & 
    10.13                 & 
    10.67                 & 
    3.24$\times$          \\
    
    & \texttt{LLM.int8()}\cite{dettmers2022llm} & 
    8                     & 
    8                     & 
    OPT-13B               & 
    12.45                 & 
    12.45                 & 
    1.22$\times$          \\
    
    & ZeroQuant\cite{yao2023zeroquantv2} & 
    4                     & 
    8                     & 
    BLOOM-176B            & 
    10.97                 & 
    11.22                 & 
    -                     \\
    
    & LLM-QAT\cite{liu2023llm} & 
    4                     & 
    8                     & 
    LLaMA-30B             & 
    6.00                  & 
    6.90                  & 
    -                     \\
    
    & INT2.1\cite{chai2023int2} & 
    2                     & 
    16                    & 
    LLaMA-7B              & 
    7.52                  &
    12.52                 & 
    -                     \\ \bottomrule
    \end{tabular}
\end{table*}

\subsubsection{Weight + Activation Quantization.}

Quantizing both weights and activations of LLMs is a non-trivial problem for several reasons. First, the range and statistics of activation are unknown until actual runtime. Second, quantizing weights and activations enables efficient low-bit datatype operations on specific hardware. Third, systematic outliers appearing in the LLM activations are vital to the model performance, and shouldn't be clipped in quantization. While the first two reasons apply to all models, the third reason is unique to LLMs and differentiates methods for LLMs from methods for previous models.

Similar to its weight-only counterpart, weight + activation quantization can also use basic uniform quantization methods \cite{dettmers2022llm, yao2022zeroquant, yao2023zeroquantv2, yuan2023rptq} but with a special notification of outliers in activations. 
\texttt{LLM.int8()} \cite{dettmers2022llm} emphasizes the emergence of extreme outliers in LLM's activations as the model size scales up. The authors show that these outliers are highly systematic. Given input activation $X_{f16} \in \mathbb{R}^{N \times D_{\mathrm{in}}}$ to a linear layer, outliers occur systematically for almost all $N$ tokens in a sequence. Still, they are limited to speciﬁc feature/hidden dimensions $\hat{d} \in \{1, 2, \dots, D_{\mathrm{in}}\}$. \texttt{LLM.int8()} thus propose to separate the outlier feature dimensions $O = \{\hat{d}|\hat{d} \in \mathbb{Z}, 1 \le \hat{d} \le D_{\mathrm{in}}\}$ which contains all feature dimensions $\hat{d}$ that have at least one activation outlier with a magnitude more significant than the threshold $\alpha$. The outlier dimensions are preserved in high-precision datatypes (\eg, FP16) while average values are quantized using symmetric uniform quantization into low-precision datatypes (\eg, INT8). With Einstein's notation, the matrix multiplication thus becomes:
\begin{equation}
  \textbf{X}_{f16}\textbf{W}_{f16} \approx 
    \sum_{\hat{d}\in O} \textbf{X}_{f16}^{\hat{d}}\textbf{W}_{f16}^{\hat{d}} +
    \textbf{S}_{f16} \cdot \sum_{d \notin O} \textbf{X}_{i8}^d\textbf{W}_{i8}^d
\end{equation}
where $\textbf{S}_{f16}$ is the dequantization factor. The number of outlier dimensions $|O|$ is quite small, so this decomposition would only consume more than $0.1\%$ additional memory typically. 
Instead of separating outliers into an additional matrix, RPTQ \cite{yuan2023rptq} proposes to cluster and reorder the dimensions of activation $\textbf{X} \in \mathbb{R}^{N \times D_{\mathrm{in}}}$ based on the minimum and maximum of the dimension $i$, denoted as $(\textbf{X}_{min, i}, \textbf{X}_{max, i})$. The idea is to group dimensions with outliers into the same cluster and perform cluster-wise quantization.
It should be noted that the statistics of each activation dimension are measured on a calibration set so that the clustering can be done before inference to find the new order of dimensions. To further reduce latency, RPTQ fuses the reorder operation into other operations: 
1) Combine with the LayerNorm operation to avoid additional data movement and adjust. 
2) Reorder columns of weight $\textbf{W}$ to reorder the dimensions of the output 
$\textbf{Y} = \textbf{XW}$.

Recently, \textit{low-bit floating-point} formats (\eg, FP4, FP8) have emerged as promising alternatives for LLM quantization \cite{zhang2023integer, wu2023zeroquant, wu2023leap}. FP8 format has garnered support from leading hardware vendors like NVIDIA despite its potentially higher hardware costs. Intuitively, low-bit FP formats can be viewed as a particular case of non-uniform quantization, offering a typically more extensive data range and higher precision for small values but lower precision for large ones. Such characteristics of the FP format help solve outlier problems in activations. 
MoFQ (Mixture-of-Formats Quantization) \cite{zhang2023integer}, and ZeroQuant-FP \cite{wu2023zeroquant} both show that FP8 quantization is consistently better than INT8 when it comes to activation quantization.
MoFQ further provides an algorithm to determine the optimal data format from some candidates (INT4, INT8, FP4, FP8) for each layer based on tensor error, layer output error, or model output error. Also, MoFQ reallocates special NaN (Not a Number) and Inf (Infinity) values in standard FP formats to normalized numbers to enhance and let the FP format represent more values, which is especially important for low-bit formats like FP4. ZeroQuant-FP quantizes both weight and activation into FP format. For cases using FP4 weights and FP8 activations, ZeroQuant-FP proposes a bit-shifting method to cast FP4 to FP8 to improve inference efficiency efficiently.

Another promising way is to suppress outliers that appear in the activation dimensions \cite{xiao2023smoothquant, wei2022outlier, wei2023outlier, li2023fptq, shao2023omniquant}. The general idea is that we can scale down the outlier dimensions $i$ in activations by factor $\mathbf{s}_i$ and scale up the corresponding dimension in the weight matrix by factor $\mathbf{s}_i$ without changing the output of the layer:
\begin{equation}
  \textbf{Y} = \textbf{XW} = 
  (\hat{\textbf{X}} \mathrm{diag}(\textbf{s}))\cdot 
  (\mathrm{diag}(\textbf{s})^{-1}\hat{\textbf{W}}) = 
  \hat{\textbf{X}} \hat{\textbf{W}}  
\end{equation}
so that the activation $\hat{\textbf{X}}$ after scaling would be quantization-friendly. 
SmoothQuant \cite{xiao2023smoothquant} computes the per-dimension scaling factor using:
\begin{equation}
  \mathbf{s}_i = \mathrm{max}(|\textbf{X}_i|)^\alpha / \mathrm{max}(|\mathbf{W}_j|)^{1-\alpha}
\end{equation}
where $\alpha$ is a hyperparameter to control how much difficulty will be migrated from activation to weights. Outlier Suppression \cite{wei2022outlier} discovers that $\gamma$ in LayerNorm acts as a sinful amplifier for the outliers. Hence, it proposes the Gamma Migration method, which uses $\gamma^{-1}$ in the previous layer as the scaling factor $\textbf{s}$.
Outlier Suppression+ \cite{wei2023outlier} extends the method through introducing additional shifting factor $\textbf{z}$:
\begin{equation}
    \begin{split}
        \textbf{Y} &= \textbf{XW} 
        = (\hat{\textbf{X}} \mathrm{diag}(\textbf{s}) + \textbf{z})\cdot 
        (\mathrm{diag}(\textbf{s})^{-1}\hat{\textbf{W}}) \\ 
        &= \hat{\textbf{X}} \hat{\textbf{W}} + \textbf{zW} 
        = \hat{\textbf{X}} \hat{\textbf{W}} + \hat{\textbf{b}}.
    \end{split}
\end{equation}
The per-dimension shifting factor is computed as $\mathbf{z}_i = (\mathrm{max}(\textbf{X}_i) + \mathrm{min}(\textbf{X}_i)) / 2$, which helps remove the asymmetry in the activation $\hat{\textbf{X}} = (\textbf{X} - \textbf{z})\mathrm{diag}(\textbf{s})$. 
FPTQ \cite{li2023fptq} proposes a new offline logarithmic activation equalization (LAE) method that moderates activation distributions in a non-linear fashion, each channel of the scaling factor $\mathbf{s}$ is computed as:
\begin{equation}
    \mathbf{s}_i = \mathrm{max}(|\mathbf{X}_i|) / \mathrm{log}_2 (2 + \mathrm{max}(|\mathbf{X}_i|))
\end{equation}

While the above methods use hand-craft quantization parameters such as scaling factor $\mathbf{s}$, Outlier Suppression+ in contrast proposes to find optimal scaling factor $\textbf{s}$ by optimizing the following objective using a calibration set:
\begin{equation}
  \min_{\textbf{s}} \mathbb{E} [||\textbf{XW} - 
  (Q(\hat{\textbf{X}})Q(\hat{\textbf{W}}) + \hat{\textbf{b}})||_2^2].
\end{equation}
Further, OmniQuant \cite{shao2023omniquant} proposes learning the clipping range $[\alpha, \beta]$ to modulate the extreme values of weights.
QLLM \cite{liu2023qllm} proposes a unique approach to handling outliers in activations. The technique involves an adaptive channel reassembly process that splits the outlier channels into multiple sub-channels. This ensures a more even distribution of activation magnitudes. The process also merges similar channels, maintaining the original channel number for efficiency purposes.

There are also some recent studies employing methods that do not fall into any of the previous paragraphs, so we briefly cover them here for completeness.
REx \cite{yvinec2022rex} quantizes the quantization error, \ie, $\textbf{W} - Q(\textbf{W})$, so that there is a smaller quantization error between the original value and the dequantized value, which trades efficiency for higher model performance.
OilVe \cite{guo2023olive} employs the outlier-victim pair (OVP) mechanism, which prunes some quantized low-precision normal values to make extra space for the high-precision outliers.

As a summary of PTQ methods for LLMs quantization, we briefly compare and contrast the weight-only and weight+activation quantization methods here. On the one hand, weight-only quantization methods can push the quantization limit to lower bit-widths (even to 3 bits or 2 bits), which significantly reduces the memory size of devices required by LLMs. This is because model weights use most of the memory. On the other hand, weight+activation quantization can take advantage of the additional speedup that comes with efficient low-bit arithmetic supported by specific hardware and doesn't introduce additional dequantization overhead during inference. However, these methods often require more bit-width ($\sim$ 8 bits) to store the weights and activations. Both weight-only and weight+activation quantization methods have their strengths and weaknesses, and they are both active research directions with great potential and demand.

\subsection{Quantization-Aware Training for LLMs}
\label{quan: qat-llm}

Quantization-Aware Training (QAT) is the method of re-training a quantized model to recover from the performance degradation caused by the quantization. As is illustrated in the previous sections, QAT for models before LLMs (\eg, CNN, medium-size LM) has achieved impressive success. However, such methods often involve full-parameter re-training of the entire model, which is too expensive for LLMs, so there are also some attempts to combine quantization with parameter-efficient training methods to significantly lower the cost of QAT on LLMs. 

As a result, we divide current QAT methods on LLMs into two categories: 
\textit{full-paramter re-training} and \textit{parameter-efficient re-training}. We'll discuss works in these two categories respectively in the following parts.

\subsubsection{Full-Parameter Re-Training.}

The primary concern of using the QAT framework in LLMs is re-training them on a smaller dataset without hurting their emergent abilities, such as in-context learning. Current methods often combine QAT and distillation to preserve these abilities of the original (teacher) model \cite{kim2022understanding, neill2023self}.

LLM-QAT \cite{liu2023llm} directly applies the basic QAT framework \cite{jacob2018quantization} to LLMs. To cope with the problem, LLM-QAT proposes to use a data-free distillation method, where the data is generated using the original LLM, and the quantized LLM is trained to match the output distribution of the original LLM on the generated data. Also, LLM-QAT enables quantization and QAT of key-value caches, which takes up large memory in the long sentence generation process. 

To alleviate the unbearable cost of re-training full LLM, 
ZeroQuant \cite{yao2022zeroquant, wu2023zeroquant, yao2023zeroquantv2} proposes a layer-to-layer knowledge distillation method. The method uses the original LLM as a teacher and processes the weights of LLM in a layer-by-layer order. Assume the LLM has $N$ layers $L_1, L_2, \dots, L_N$ and has some input dataset ${\textbf{X}}$ and the quantized version of $L_k$ is $Q(L_K)$. After QAT and distillation of layers $L_1, L_2, \dots, L_{k-1}$, we use the following loss to update
$L_k$:
\begin{equation}
    \mathcal{L}_{k} = \mathrm{MSE}(L_k \cdot L_{k-1} \dots L_1(\textbf{X}) -
    \hat{L_k} \cdot L_{k-1} \dots L_1(\textbf{X}))
\end{equation}

\subsubsection{Parameter-Efficient Re-Training.}

There are a bunch of works using parameter-efficient methods (\eg, LoRA, Adapter, Prompt Tuning) to finetune the LLMs, which will be discussed in Appendix \ref{subsec:peft}. In this section, we discuss some methods that use parameter-efficient methods in the re-training process of QAT.

Typical works \cite{dettmers2023qlora, hua2023lacos, chai2023int2, li2023loftq, liu2023qllm, kaushal2023lord, xu2023qa} adopt Low-Rank Adaption (LoRA) to re-train quantized LLMs in a relatively acceptable compute budget. 
QLoRA \cite{dettmers2023qlora} quantize the weight of LLMs into 4-bit NormalFormat and subsequently adopt LoRA with 16-bit BrainFloat to finetune the quantized model on downstream tasks with cross entropy loss. It further introduces a technique named \textit{double quantization}, which quantizes the quantization parameters to compress further the model's size in the trade of computation speed. Combining all these techniques, QLoRA enables finetuning a 65B LLM on a GPU with 30G memory efficiently.
Following QLoRA, QA-LoRA \cite{xu2023qa} proposes to integrate group-wise quantization into QLoRA. The authors suggest that quantization parameters in QLoRA are much less than LoRA parameters, resulting in an imbalance between quantization and low-rank adaptation, while group-wise operations can alleviate the problem by increasing the number of parameters of quantization and decreasing that of adaptation.
Besides, LoftQ \cite{li2023loftq} finds the zero initialization of LoRA matrices in QLoRA inefficient for downstream tasks. Instead, LoftQ proposes to initialize the LoRA matrices with the singular value decomposition (SVD) of the difference between the original and quantized weights, \ie, $\mathbf{W} - Q(\mathbf{W})$. LoftQ alternates between quantization and SVD to obtain a better approximation of the original weights.
LACos-BLOOM \cite{hua2023lacos} quantizes the model weights using 8-bit block-wise uniform quantization. The quantized model is then finetuned using a scalable LoRA and 8-bit Adam optimizer. 
INT2.1 \cite{chai2023int2} utilized GPTQ to quantize LLMs into INT2 and found that the behavior of the quantized model deviates significantly from the original full-precision counterpart. INT2.1 integrates additional trainable parameters (LoRA matrices) into the model and solely updates the LoRA matrices with takes up of only $5\%$ of total parameters. The training objective combines a scaled Kullback-Leibler divergence from the full precision model to the quantized one and the cross entropy loss to encourage accurate next token prediction. Their experiment indicates that an INT2 Large Language Model (LLM) finetuned with LoRA can generate linguistically coherent English text and exhibit adherence to prescribed instructions.

Other works \cite{kwon2022alphatuning, kim2023memory} freeze the quantization indices and solely finetune quantization parameters (\eg, scaling factor $S$ in uniform quantization and quantization level $\Delta_i$ in non-uniform quantization). 
AlphaTuning \cite{kwon2022alphatuning} works by employing binary-coding quantization \cite{rastegari2016xnor}. During the adaptation phase, the binary values are frozen for all tasks, while the scaling factors are ﬁne-tuned for the downstream task.
PEQA \cite{kim2023memory} quantizes each fully connected layer of LMs into a matrix of low-bit integers and a scalar vector using uniform quantization. Subsequently, finetuning occurs on the scalar vector for each downstream task.

Works also combine quantization with adapters \cite{park2022quadapter} and prompt tuning \cite{xu2023compress}.

\subsection{Other Topics for LLM Quantization}
\label{quan: advance-topic}

Some quantization-related works can not be categorized into PTQ or QAT; we discuss such works in this section.

One important topic is co-designing efficient kernels along with quantization algorithms \cite{shen2023efficient, pegolotti2023generating}, designing hardware-friendly quantization methods \cite{wang2019haq, yu2023boost} and integrating quantization methods in real-world applications \cite{lin2023pushing, rahman2023quantized, kurtic2023sparse, isik2023gpt, wei2023greener}. 
LUT-GEMM \cite{park2022nuqmm} is an efficient kernel designed for an extended version of BCQ methods \cite{rastegari2016xnor}, which can represent both uniform and non-uniform quantization. Since weights are characterized by a binary vector and scale factors in BCQ, LUT-GEMM can pre-compute and store all possible combinations of full-precision activations and binary patterns in a lookup table (LUT) to avoid repeated computation and remove dequantization of weights, which accelerates the latency of OPT-175B model with 3-bit quantization by $2.1 \times$ compared to conv entional simulated quantization. 
Many uniform \cite{yao2022zeroquant, yao2023zeroquantv2, dettmers2022llm, guo2023olive} and non-uniform quantization methods \cite{kim2023squeezellm} discussed in the above sections also design special kernels to reduce the overall latency.

Other meaningful works study the intrinsic characteristics of LLM quantizations \cite{hu2022empirical, dettmers2023case, liu2023emergent}. 
For example, Dettmers and Zettlemoyer \cite{dettmers2023case} run extensive experiments with 16-bit activations and $k$-bit weights ($3 \le k \le 8$) at scales of 19M to 176B parameters across LLM families BLOOM, OPT, NeoX/Pythia and GPT-2. The authors focus on the tradeoff between zero-shot ability and total model bits and show that 4-bit precision is almost universally optimal for the tradeoff across different LLM classes and quantization methods. 
Liu et al. \cite{liu2023emergent} aim to investigate the impact of quantization on \textit{emergent abilities}, which are essential characteristics that distinguish LLMs from small language models. Their empirical experiments show that emergent abilities still exist in 4-bit quantization models, while 2-bit models encounter severe performance degradation on the test of these abilities. The authors conduct further detailed experiments on enhancing the performance of extremely low-bit models.

Some works \cite{bondarenko2023quantizable, wei2022outlier, ahmadian2023intriguing} also focus on studying the reasons behind the emergence of systematic outliers in LLMs and looking for ways to suppress the outliers from the source. 
Quantizable Transformer \cite{bondarenko2023quantizable} ascribes the outliers in activations to the behavior of attention heads that try not to update residual. The authors designed clipped softmax and gated attention accordingly to grant the model the ability to produce minimal magnitude (or even exact zeros) output of attention function without having outliers. 
Outlier suppression \cite{wei2022outlier}, however, treats $\gamma$ in LayerNorm as the sinful amplifier of outliers. There is still no consensus on the source of activation outliers.
However, Ahmadian et al. \cite{ahmadian2023intriguing} find that outlier dimensions may not be an inherent product of scale as is thought in previous works \cite{dettmers2022llm}, but rather sensitive to the optimization conditions (\eg, dropout rate, weight decay, datatype) present during pre-training.

%% file: pruning.tex
As a conventional technique employed for the compression and acceleration of neural networks, pruning eliminates non-essential weights or structures from models, while preserving the performance of the networks at a level nearly equivalent to their original state. 
Although pruning has shown remarkable results in CNNs\cite{wen2016learning}, its effectiveness is less robust for LLMs when compared to other compression techniques such as quantization and distillation.
The reason why pruning becomes less effective comes from the fine-tuning process. 
The high cost of fine-tuning due to the large number of model parameters makes it more difficult to achieve the full effect of pruning.
Nevertheless, pruning is a crucial technique for compressing models, necessitating further exploration to enhance and refine its effectiveness in yielding improved results in LLMs.

In the following section, we will provide an overview of pruning methods and basic concepts in Section \ref{pru: basic-concepts}. Subsequently, in Section \ref{pru: before-llm}, we will expound upon pruning techniques tailored for medium-size language models (\ie, models with parameters in billions), given their structural similarities with LLMs. Section \ref{pru: llm} will delve into a detailed exploration of pruning methodologies specifically designed for LLMs. Finally, in Section \ref{pru: topic}, we will introduce some auxiliary techniques that are not pruning methods but associated with pruning to improve LLM pruning results, and then discuss the challenges for future advancements in the field of LLM pruning.

\subsection{Basic Concepts}
\label{pru: basic-concepts}
Numerous classification criteria exist for pruning. 
Nevertheless, the most significant things among them are two fundamental problems: what to prune and how to prune. 
The answers to these two problems correspond respectively to the pruning unit and metric.
We will introduce these two fundamental concepts and some other basic ones.

1) \textbf{Pruning Unit.} 
The first fundamental problem with pruning is what kind of elements should be pruned. The pruning units refer to the minimal pruning elements in the pruning process, encompassing elements such as weights, neurons, attention heads, layers, and \etc.  
Based on pruning units, pruning methods can be broadly categorized into unstructured pruning and structured pruning. 
In unstructured pruning, the pruning units focus on individual weights. The weights to be pruned are zeroed out. Whereas in structured pruning, the pruning units encompass broader network structures, such as neurons, attention heads, and layers. The structures to be pruned are removed from the networks.

Unstructured pruning tends to get a higher sparsity ratio and maintain better performance as it is not limited to the network structure and can prune individual weights. 
However, the irregularly sparse patterns of weight matrices, stemming from the non-systematically occurring zero values, exhibit computational efficiency nearly equivalent to dense matrices. Consequently, achieving significant gains in inference speedup is infrequent in unstructured pruning.

Structured pruning makes it easy to achieve inference speedup as it prunes network structures  (\eg, attention heads, feed-forward network (FFN) neurons, and hidden dimensions). Yet inevitably integrated structure deletions may cause the performance descent of the model. To avoid model collapse, the sparsity ratios of structured pruned models are lower than unstructured ones.  

Formally, a binary mask $\textbf{z}$ usually covers the pruning unit during the pruning process and is multiplied into the model after pruned. For unstructured pruning, the pruning process can be defined as a constrained optimization problem:
\begin{equation}
\begin{aligned}
  \label{pru: eq-1}
  &\mathop{\min}_{\textbf{w},\textbf{z}}\mathcal{L} (\textbf{w}\odot\textbf{z};\mathcal{D})=
  \mathop{\min}_{\textbf{w},\textbf{z}}\frac{1}{N} \sum\limits_{i=1}^{N}\ell (\textbf{w}\odot\textbf{z}; (x_i,y_i)),\\
  &s.t.\quad \Vert \textbf{z} \Vert_0 \leq t,
\end{aligned}
\end{equation}
where $\odot$ corresponds to the element-wise product, $\textbf{w}=\{w_1,w_2,...,w_M\}$ is the network weights, $\mathcal{D}$ is a dataset composed of $N$ input $x_i$ and output $y_i$ pairs, and $t$ is the target non-sparsity ratio  (\ie, one minus sparsity ratio). Similarly, the pruning process for structured pruning is as follows:
\begin{equation}
\begin{aligned}
  \label{pru: eq-2}
  &\mathop{\min}_{\textbf{w},\textbf{z}}\mathcal{L} (\textbf{s}\odot\textbf{z};\mathcal{D})=
  \mathop{\min}_{\textbf{w},\textbf{z}}\frac{1}{N} \sum\limits_{i=1}^{N}\ell (\textbf{s}\odot\textbf{z}; (x_i,y_i)),\\
  &s.t.\quad f (\textbf{z};\textbf{s}) \leq t,
\end{aligned}
\end{equation}
where $\textbf{s}=\{s_1,s_2,...,s_K\}$ is the pruning structures composed of $\textbf{w}$, and $f (\cdot)$ is the function to compute non-sparsity ratio according to the binary masks and structures.

2) \textbf{Pruning Metric.} 
The second fundamental problem with pruning is how to determine whether an element is essential and should be pruned or not.
Pruning metric is the answer to this problem. 
The pruning metric is the criterion to identify the importance of the pruning units.
It can be roughly divided into three parts: magnitude-based, loss-based  (\ie, considering the first-order and second-order derivative information of the weighs belonging to the pruning units), and regularization.

The magnitude-based pruning methods use the magnitudes (\ie, absolute values) of weights and activation values as a part of the pruning metrics. 
The fundamental principle underlying this class of methods is that the magnitude of weight or the activation value from the pruning unit intuitively reflects its importance. 
The magnitude of the weight alone can serve as a pruning metric, constituting a well-known foundational pruning method known as Magnitude Pruning\cite{han2015learning}. 
Magnitude Pruning is the vanilla magnitude-based pruning method.
In this method, a threshold is set to zero out weights with smaller magnitude and the threshold typically is derived from sparsity ratio.
Despite the definition of importance score being quite heuristic, Magnitude Pruning demonstrates efficacy across various models.

In addition to the intuitive magnitude-based metric, another more sophisticated kind of metric is the loss-based metric. 
The loss-based metric is designed to attribute the importance of a pruning unit to its impact on loss. 
If the loss increases significantly after pruning an element, it indicates that the element should not be pruned.
More precisely, following the pruning of an element, the greater the increase in loss, the more crucial the importance of that element becomes. 
However, examining the loss after pruning individual elements one by one is resource- and time-intensive.
In contrast, employing the Taylor expansion provides a more convenient expeditious method for elucidating the loss alteration.
The alteration in loss after the pruning can be quantified using a Taylor expansion, incorporating the first-order or second-order derivatives of the pruning units with respect to the loss and higher-order ones, which are usually ignored. 
In comparison to the resource- and time-intensive approach of evaluating loss after pruning each element individually, the computation of the first-order and second-order derivatives emerges as a more efficient and time-saving alternative.


Besides, regularization methods encompass $L_0$, $L_1$, and $L_2$ regularization. While $L_1$ regularization is known for inducing sparsity in weights, $L_0$ regularization is a more commonly employed regularization approach in the context of pruning.


  
\begin{figure}
    \centering
    \includegraphics[width=0.7\linewidth]{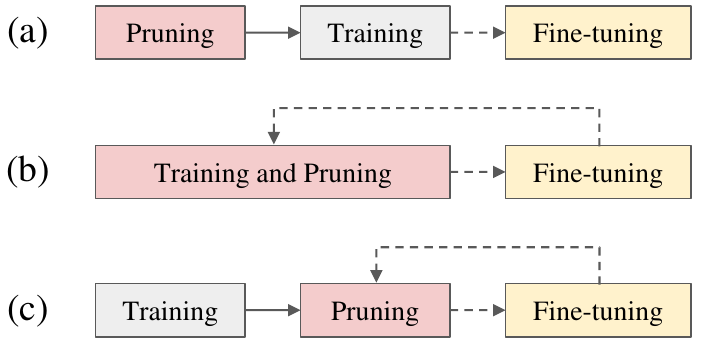}
    \caption{Three classes of static pruning methods. (a) Pre-training pruning; (b) During-training pruning; (c) Post-training pruning.}
    \label{pru:static}
\end{figure}
3) \textbf{Dynamic/Static Pruning.} 
To enhance adaptability to diverse inputs, a kind of pruning method, referred to as \textit{dynamic pruning}, constructs the network in a manner contingent upon the specific input characteristics.
We will these later in Section~\ref{dynamic_inference}.
In contrast,  \textit{static pruning} methods prune the model at training time and fix the architecture after pruning, thus different inputs share the same pruned network. 
Static pruning methods can be classified as \textit{pre-training pruning}, \textit{during-training pruning} and \textit{post-training pruning} according to the pruning period, as shown in Fig.~\ref{pru:static}.
\begin{itemize}
  \item \textit{Pre-training pruning}: prunes the initialized network first and then trains the sparse network.
  \item \textit{During-training pruning}: trains and prunes the dense network at the same time, where regularization methods are representative.
  \item \textit{Post-training pruning}: is the most popular type of pruning pipeline, prunes the trained dense network to get a sparse network, where they usually follow a training, pruning, and fine-tuning paradigm as we mentioned before.
\end{itemize}

4) \textbf{Iterative/One-shot Pruning.} 
As pruning damages model performance inevitably, a popular paradigm of pruning pipeline consists of three steps:  training, pruning, and fine-tuning, as shown in Fig.~\ref{pru:static} (b), (c).
The initial step involves training the network to ascertain the importance of individual pruning units. Subsequently, the second step entails the removal of non-essential pruning units through pruning, and the third step focuses on fine-tuning to recover the performance of the model post-pruning.

Given the potential for the fine-tuning process to render initially zero-valued weights as non-zero, the final two steps are subject to iterative repetition until the targeted sparsity ratio is achieved. This iterative design underscores that each pruning step is succeeded by a fine-tuning step, thereby facilitating the preservation of the model's performance. 
These methods containing iterative pruning and fine-tuning rounds are classified as \textit{iterative pruning}.

However, as the model parameters get huger, the iterative pruning and fine-tuning process is expensive and time-consuming. Thus more pruning methods tend to prune the network only once to the target sparsity ratio, 
discarding the iterative pruning and fine-tuning rounds. These methods are classified as \textit{one-shot pruning}.

5) \textbf{Global/Local Pruning.}
The early pruning approaches compare all the pruning units to identify and eliminate those less essential.
Given that the comparison scope in these methods encompasses the entire network, they are categorized as \textit{global pruning} approaches.
However, global pruning permits distinct sparsity ratios for individual local regions. It might result in excessive pruning of a specific region (\eg, a layer, a column), and exert a notable influence on the overall performance of the model.
The resolution of this issue lies in the application of \textit{local pruning} methodologies.
Local pruning imposes constraints on the sparsity of each region, thereby ensuring that the sparsity ratios within each region do not reach excessively low thresholds, consequently mitigating the risk of model collapse.

6) \textbf{Data-driven/Data-free Pruning.} 
The categorization of pruning methods into \textit{data-driven} and \textit{data-free} modalities distinguishes the reliance on data for pruning decisions. Specifically, \textit{data-driven pruning} methods, exemplified by the majority of pruning techniques, derive pruning decisions from available data. Conversely, \textit{data-free pruning} methods, such as Magnitude Pruning\cite{han2015learning}, execute network pruning independent of data input. In general, data-driven pruning methods tend to exhibit superior performance, given their dependence on data-driven insights, while data-free pruning methods are less effective but data-independent.

7) \textbf{Upstream/Downstream Pruning.} 
The training of language models involves two main stages—pre-training and fine-tuning. Pruning methods can be classified based on when they are applied. Techniques identified as \textit{upstream pruning} involve the pruning of the model before the fine-tuning stage. In contrast, \textit{downstream pruning} methods are characterized by the simultaneous execution of pruning alongside the fine-tuning process.
Accordingly, upstream pruning retains the adaptability of the pruned model for multiple tasks, ensuring its versatility. Conversely, downstream pruning directs the pruned model to concentrate on a specific, well-defined task.

\subsection{Pruning Methods for Medium-Size Language Models}
\label{pru: before-llm}
Language models, such as GPT-2 and BERT, are initially trained on extensive corpora and exhibit applicability across various downstream tasks after fine-tuning. Specifically, the pruning of language models distinguishes itself from the pruning methodologies employed in Convolutional Neural Networks  (CNNs) or Recurrent Neural Networks  (RNNs) in three key aspects. 
First and foremost, the sheer magnitude of parameters in language models surpasses that of CNNs or RNNs. For instance, the BERT-large model encompasses 335 million parameters, whereas 
the parameters of a typical RNN are in the range of tens of millions\cite{narang2017block}.
The increased number of parameters amplifies the temporal and computational demands of the fine-tuning phase.
Consequently, language model pruning necessitates addressing the challenges posed by this substantial parameter abundance. 
Secondly, language models have the potential to undergo fine-tuning for a multitude of downstream tasks.
Certain upstream pruning methodologies necessitate the retention of the language model's capacity to function as a multi-task solver.
Thirdly, transformer-based language models exhibit a distinctly different structural composition. Hence, in light of the model's architecture, certain structured pruning methods may require reconfiguration to align with the structure of the model.
In conclusion, there exist specialized designs of pruning methodologies for language models that are tailored to their unique characteristics, deviating from conventional pruning approaches.

We will introduce these pruning techniques for medium-size language models in the following, including approaches that are specially designed for Transformer-based models and generic to plenty of models with different architectures. 
In consideration of the fundamental features of pruning methods  (\ie, the determination of what to prune and how to prune), we shall introduce these pruning methods in the order of pruning unit and metric.
Initially, we classify pruning methods into two primary components: unstructured and structured ones.  Subsequently, based on the sequence of pruning criteria, we will expound upon each of the three pruning methods: magnitude-based pruning, loss-based pruning, and regularization.

\subsubsection{Unstructured Pruning for Medium-Size Language Models}
Unstructured pruning methods zero out non-essential weights without any specific constraints. We will introduce unstructured pruning methods for medium-size language models in a systematic order based on specific metrics, including magnitude-based pruning, loss-based pruning, and regularization.

\textbf{1) Magnitude-based Pruning}

Magnitude-based pruning, characterized by its simplicity and efficacy, incorporates the magnitudes of weights and activation values into its pruning metrics. In this section on magnitude-based pruning for medium-size language models, we find that all of the related methods exclusively focus on the magnitudes of weights. Consequently, we will introduce these magnitude-based pruning methods with weights.

Magnitude Pruning\cite{han2015learning}, recognized as the most commonly utilized pruning method, has been examined in the context of medium-size language models\cite{gordon2020compressing,chen2020lottery,prasanna2020bert,jaiswal2023instant}. 
Gordon et al.\cite{gordon2020compressing}  conducted a study focusing on the compression of BERT through Magnitude Pruning. The findings reveal that approximately $30-40\%$ of the weights 
are non-essential
and can be discarded without affecting BERT’s performance. 
Furthermore, 
fine-tuning BERT for a specific task does not contribute to an enhancement in the ultimately achievable sparsity ratio. 
This implies that BERT can undergo pruning once during the pre-training phase, obviating the need for separate pruning for each task, all while maintaining performance integrity. 
Based on this, Prune Once for All\cite{zafrir2021prune} is to prune models once for all tasks before fine-tuning. 

Magnitude pruning, characterized by the direct pruning of the model sparsity ratio to the target ratio, may result in a substantial deterioration of model performance. 
Compared to Magnitude Pruning, Gradual Magnitude Pruning  (GMP)\cite{zhu2017prune} introduces a sparsity ratio schedule, gradually reducing the sparsity ratio throughout the pruning process.
Prune Once for All\cite{zafrir2021prune} and GMP$\star$ \cite{kurtic2022gmp} are both implementations of GMP specifically applied to language model pruning. 
Besides, GMP$\star$ introduces an initial substantial pruning step to better adapt to a high target sparsity  (\eg, 97\%).
This approach allows for more recovery time in subsequent pruning steps, ultimately leading to improved performance, outperforming most pruning methods including Prune Once for All\cite{zafrir2021prune}.

\textbf{2) Loss-based Pruning}

While magnitude-based pruning is easy to implement, the magnitude alone may not accurately reflect the importance of weights in some instances.
The magnitude of certain weights may be diminutive, yet their contribution remains essential\cite{yin2023junk}.
Therefore, a more scientifically grounded approach involves assessing these weights within the context of a specific task. The methods in this section adopt the loss-based pruning strategy tailored for medium-size language models. These approaches align with a more nuanced evaluation based on the performance. 
Given that the model's training process is inherently geared towards minimizing this loss, the loss undoubtedly stands out as the most reliable measure of the model's performance.

The first major category of loss-based pruning methods integrates information about the gradients within the specific metrics.
The universal expression by which these methods evaluate the importance of weights can be articulated through the negative gradient-weight product, expressed as follows:
\begin{equation}
\mathbf{I}=-\mathbf{w}\nabla\mathcal{L} (\mathbf{w})
\end{equation}
The first interpretation of this expression pertains to the weight change. The negative gradient direction of the weights signifies the direction in which the weights are intended to increase. Consequently, if the weight direction aligns with the direction of weight growth, it indicates the importance of that weight in the specific task, as the task necessitates the continued increase in its magnitude. 
Alternatively, the second interpretation of this expression can be simplistically conceived as the first-order term of the Taylor expansion of loss alteration, with higher-order terms being disregarded.

Many methods have implemented their improvements based on this universal expression. 
Movement Pruning\cite{sanh2020movement} accumulates multiple updates of the negative gradient-weight product. Accumulating such information aids in minimizing fluctuations during pruning. 
Among the first-order methods, Movement Pruning stands as a pioneering one, upon which many extensions have been developed\cite{jiang2022pruning,ren2023low}.
To mitigate the substantial variability and uncertainty introduced by mini-batch sampling and intricate training dynamics, PLATON \cite{zhang2022platon} employs a weight pruning strategy that considers both the importance and uncertainty associated with individual weights. The uncertainty originates from changes in importance. To enhance stability, both importance and uncertainty undergo exponential moving averaging. The final importance score of each weight is determined by the product of smoothed importance and uncertainty.
Parameter-efficient Sparse Training  (PST)\cite{li2022parameter} and LoRAPrune \cite{zhang2023pruning} add the magnitude of weight and the accumulated negative gradient-weight product to derive the final importance score.

The second major category of loss-based pruning methods integrates information about the second-order derivative within the specific metrics.
The variation of the loss, when employing the Taylor expansion and expanding up to the second-order term while neglecting higher orders, can be expressed in the following manner:
\begin{equation}
\begin{aligned}\mathcal{L} (\mathbf{w})-\mathcal{L} (\mathbf{w}^*)&\simeq (\mathbf{w}-\mathbf{w}^*)^\top\nabla\mathcal{L} (\mathbf{w}^*)\\&+\frac12 (\mathbf{w}-\mathbf{w}^*)^\top\mathbf{H}_\mathcal{L} (\mathbf{w}^*) (\mathbf{w}-\mathbf{w}^*),\end{aligned}
\label{pru: 2st}
\end{equation}
where $\mathbf{H}_\mathcal{L} (\mathbf{w}^*)$ is the Hessian matrix.
These methods are post-training pruning methods and always prune a well-trained network $\mathbf{w}^*$. Therefore, the gradient $\nabla\mathcal{L} (\mathbf{w}^*)$ can be neglected, getting a universal expression to represent the importance of weights:
\begin{equation}
\mathbf{I}=\frac12 (\mathbf{w}-\mathbf{w}^*)^\top\mathbf{H}_\mathcal{L} (\mathbf{w}^*) (\mathbf{w}-\mathbf{w}^*).
\label{pru: 2st2}
\end{equation}

The Optimal Brain Damage  (OBD) \cite{lecun1989optimal} and the Optimal Brain Surgeon  (OBS)\cite{hassibi1993optimal} represent the second-order pruning approaches in early works. Both of them utilize the Hessian of the loss function to selectively eliminate specific parameters while minimizing the impact on loss. 
To streamline calculations, both methods simplify the computation of the Hessian matrix to some extent. However, OBD computes solely the diagonal entries of the Hessian matrix, whereas OBS also considers the impact of off-diagonal entries.
These methodologies have served as inspiration for numerous subsequent approaches. The Optimal BERT Surgeon \cite{kurtic2022optimal} extends the principles of the OBS to the context of BERT, yielding better results when compared to some magnitude-based pruning methods\cite{han2015learning,zafrir2021prune,kurtic2022gmp}  and the first-order pruning methods\cite{sanh2020movement,zhang2022platon}.

\textbf{3) Regularization}

In addition to the aforementioned methods, regularization techniques find many applications in medium-size language models. $L_1$ and $L_2$ regularization are popular methods employed to counteract network overfitting. Both introduce a regularization term into the loss function. Besides, $L_1$ regularization has the additional effect of inducing sparsity in weights.
However, when it comes to directly pruning the network, $L_1$ regularization is not always the most suitable choice. This is attributed to the fact that $L_1$ regularization imposes more substantial penalties on larger weights, deviating from the original pruning objective of eliminating unimportant connections, where smaller weights are often situated.

Instead, $L_0$ regularization\cite{louizos2017learning} is a  more versatile pruning method than $L_1$ and $L_2$ regularization.
$L_0$ regularization incorporates the $L_0$ norm of weights into the loss function. Similar to $L_1$ and $L_2$ regularization, $L_0$ regularization penalizes non-zero weights. However, it distinguishes itself by applying equal penalties to all non-zero weights, aligning precisely with the pruning objective of equitably penalizing all the existing connections.

The training objective of the pruning process for all three of these regularizations can be expressed by the following formula:
\begin{equation}
\begin{aligned}
  \label{pru: eq-3}
  & \mathop{\min}_{\textbf{w},\textbf{z}}\frac{1}{N} \sum\limits_{i=1}^{N}\ell (\textbf{w}\odot\textbf{z}; (x_i,y_i))+\lambda\Vert \textbf{w} \Vert_p,\\
\end{aligned}
\end{equation}
where $\lambda$ represents the regularization factor, $\Vert \textbf{w} \Vert_p$ denotes the $L_p$ norm of the weights, and $\textbf{z}$ is a binary mask indicating whether the weights are pruned. Consequently, the $L_0$ norm of the weights can be equivalently represented by the summation of binary masks, \ie, $\Vert \textbf{w} \Vert_0 = \sum_{i=1}^{M} z_i$.

\begin{figure}
    \centering
    \includegraphics[width=0.75\linewidth]{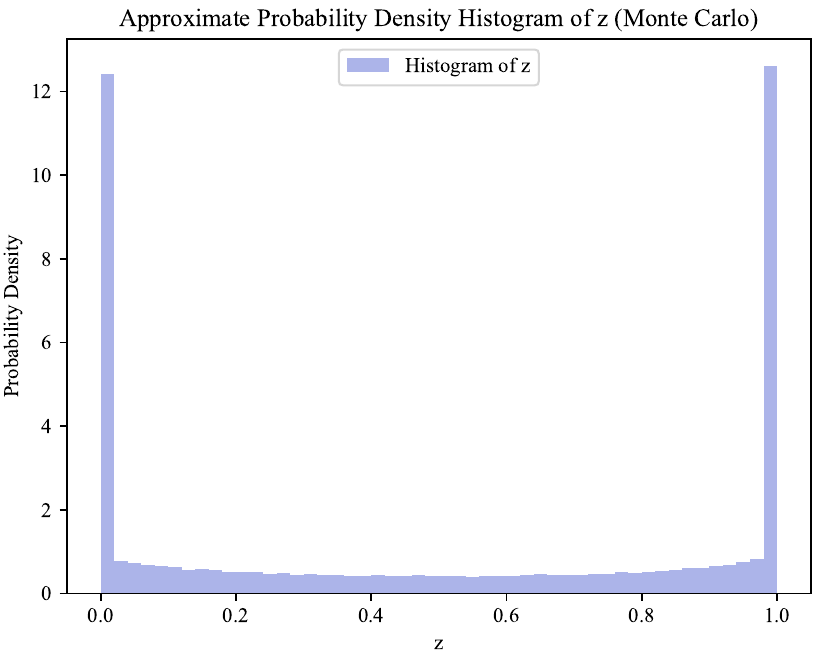}
    \caption{The approximate probability density histogram of hard concrete distribution by using Monte Carlo simulation. The parameters of this hard concrete distribution are $\log{\alpha}=0$, $\beta=0.5$, $\gamma=-0.1$, and $\zeta=1.1$. Under this specification the hard concrete distribution assigns, roughly, half of its mass to  \{0, 1\} and the rest to  (0, 1). }
    \label{pru:L0}
\end{figure}

However, the discrete nature of $\textbf{z}$ poses challenges for efficient gradient-based optimization. 
To this end, the hard concrete distribution serves as an approximation to the binary masks, allocating approximately half of its mass to \{0, 1\} and the remaining half to the interval  (0, 1), thereby bridging the gap between discrete values 0 and 1 with continuous probability mass, as shown in Fig.~\ref{pru:L0}. The formulation of the hard concrete distribution is as follows:
\begin{equation}
\begin{aligned}
  \label{pru: eq-4}
  &\mathnormal{u}\sim\mathcal{U} (0, 1),\quad \\
  &s=\rm{Sigmoid} ( (\log{\mathnormal{u}}-\log{\mathnormal{ (1-u)}} +\log{\alpha})/\beta)\\
  &\bar{s}=s (\zeta-\gamma)+\gamma,\\
  &z=\min (1, \max (0, \bar{s})),
\end{aligned}
\end{equation}
where $\mathcal{U} (\cdot)$ is a uniform distribution, $\log{\alpha}$ is the location parameter, $\beta$ is the temperature parameter, $ (\gamma, \zeta)$ is the “stretch” interval with $\gamma<0$ and $\zeta>1$. Given that the reparameterized variable $z$ is not strictly binary after training, many pruning methods adopt a threshold to discretize $z$ into binary values in the end. 
For values of $z$ below the threshold, the value is set to 0, while for values above the threshold, the value is set to 1.

While $L_0$ regularization finds broader applications in pruning, $L_1$ regularization also has some pertinent use cases, and certain methods strive to enhance $L_1$ regularization. For example, Reweighted Proximal Pruning  (RPP) \cite{guo2019reweighted} builds upon $L_1$ regularization and introduces improvements and refinements.
RPP comprises reweighted $L_1$ regularization and the proximal operator. 
The reweighted $L_1$ regularization dynamically reallocates penalty factors, assigning greater penalties to weights approaching zero. The proximal operator facilitates the separation of the sparsity pattern search and the back-propagation-based gradient update of the training loss, enabling an easier sparse pattern search. 

\textbf{4) Others}

Among the unstructured pruning methods discussed above, numerous approaches demonstrate an ability to uphold satisfactory model performance even with high sparsity. However, they encounter challenges in achieving efficient inference speedup due to the irregular nature of the sparse matrices they generate. To address this predicament, unstructured pruning methods can be integrated with N:M sparsity\cite{mishra2021accelerating}.

The principle underlying N:M sparsity mandates that within each group of $M$ consecutive weights in the neural network, no more than $N$ weights should exhibit non-zero values. 
This implies that within each group of $M$ consecutive weights, there are $N - M$ weights with zero values.
Thus the underlying hardware can compress the regularly occurring zero values within it. 
This kind of compression relies on unique architectures of the hardware, such as sparse tensor cores. For instance, the Nvidia Ampere A100 is equipped with sparse tensor cores to accelerate the 2:4 sparsity.
 
For N:M sparsity, the pruning metric is not a restricted factor. It can be seamlessly integrated with unstructured pruning methods, providing the inference speedup that pure unstructured methods may lack. For example, the determination of the sparsity pattern can be initially predicated on the magnitudes of the weights\cite{zhou2021learning}. 
Serving as a generic sparsity methodology, 2:4 sparsity demonstrates a notable twofold acceleration in computational speed without compromising performance.

\textbf{5) Discussion}

Among all these unstructured pruning methods for medium-size models, the Optimal BERT Surgeon \cite{kurtic2022optimal} demonstrates superior performance compared to various magnitude-based pruning methods\cite{han2015learning,zafrir2021prune,kurtic2022gmp}  and the first-order pruning methods\cite{sanh2020movement,zhang2022platon}  in the conducted experiments\cite{kurtic2022optimal,kurtic2022gmp}.

Nonetheless, Magnitude Pruning\cite{han2015learning} remains the most widely adopted pruning method. Because it is simple to implement, yet achieves competitive results with many intricate methods\cite{nordstrom2022unstructured}. Crucially, the pruning process of Magnitude Pruning operates independently of any specific dataset, thereby addressing challenges in some scenarios where datasets may be unavailable.

\subsubsection{Structured Pruning for Medium-Size Language Models}
Indeed, numerous unstructured pruning methods have demonstrated the capability to achieve a high sparsity ratio while maintaining performance levels comparable to their dense counterparts. However, it's noteworthy that unstructured sparse patterns do not necessarily lead to inference speedup on normal hardware. Consequently, there is an increasing focus on research dedicated to structured pruning. 

In the context of structured pruning methodologies applied to medium-size language models, the selection of appropriate pruning units assumes significance alongside the choice of pruning metric. Pruning units commonly considered encompass attention heads, FFN neurons, hidden dimensions, and \etc. Notably, employing architecture-related structures as pruning units tends to yield more favorable outcomes compared to structures unrelated to model architecture, such as weight blocks.

This observed superiority may be attributed to the preservation of fundamental principles inherent in the model's construction when reducing architecture-related structures. For instance, after the pruning of attention heads, the resultant model retains the essential characteristics of a transformer-based model, featuring a reduced number of attention heads.

In the following, we will delve into the realm of structured pruning, encompassing magnitude-based pruning, loss-based pruning, and regularization techniques.

\textbf{1) Magnitude-based Pruning}

Intuitively, the aggregation of weight magnitudes for pruning units serves as a meaningful representation of importance, which is widely applicable to convolutional kernels in CNNs.
Similarly, it can be extended to medium-size language models.
For instance, the weight magnitudes could be aggregated with $L_2$ norm to represent the corresponding importance in attention heads, FFN neurons\cite{cui2021joint}, and weight blocks\cite{li2020efficient}. 
The less important structures are then removed based on the order of their importance scores.


\textbf{2) Loss-based Pruning}

Within loss-based pruning methodologies, considerable attention has been directed towards the exploration and analysis of attention heads\cite{michel2019sixteen,li2021differentiable,yang2022gradient,wang2023task}. This focus emanates from their proclivity to become redundant, and that the rest of the heads frequently could demonstrate an aptitude for assuming the functional roles previously carried out by the pruned heads.
Michel et al. \cite{michel2019sixteen}  proposed an iterative pruning approach based on head importance scores. 
The attention heads are covered with binary mask variables. Thus, the head importance scores are computed through the examination of gradients on the binary mask variables.
The results indicated that  20-40\% of heads Transformer heads could be pruned without significantly compromising test accuracy on the target task.

However, Differentiable Subset Pruning  (DSP) \cite{li2021differentiable} demonstrated that Michel et al. \cite{michel2019sixteen} significantly underestimated the number of Transformer heads that could be pruned. The experiments showed that DSP could prune up to 90\% of heads without causing much degradation in test performance.  (Pruning up to 90\% of heads means around 20\% of the model size shrinkage as the parameters of heads are just part of the whole model.)
DSP treats Transformer head pruning as a subset selection problem. To ensure the differentiability of the subset pruner, the Gumbel–Softmax trick \cite{maddison2016concrete} and its extension to subset selection is applied to DSP. 
The results indicated superior accuracy and inference speedup of DSP compared to other head pruning methods \cite{michel2019sixteen,voita2019analyzing}.

In addition to attention head pruning, Block Movement Pruning\cite{lagunas2021block} is a block pruning method. 
It extends structured methods by considering blocks of any size and integrates these structures into the Movement Pruning\cite{sanh2020movement}. 
The matrix within the model undergoes partitioning into fixed-sized blocks, with larger block sizes yielding greater inference speedup. 
Furthermore, the combination of this approach with the pruning of neurons in the FFNs results in the best overall performance. 
Similarly, numerous methodologies prune neurons in the FFNs and attention heads simultaneously
\cite{xu2022dense,liu2021ebert,khetan2020schubert,kurtic2023ziplm,klein2023structural,park2023knowledge}. 

In addition to the above methods designed for the Transformer structures, some structured pruning methods can be generalized because the pruning units in them are neurons\cite{li2023losparse,santacroce2023matters,yang2023task}. For instance, Low-Rank and Sparse approximation  (LoSparse)\cite{li2023losparse} prunes the weight matrix in neuron level (\ie, the columns of weight matrix). 
Considering the sensitivity of parameters defined in PLATON\cite{zhang2022platon}, the importance of each neuron is defined by the cumulative sensitivity of parameters within a given column. 

\textbf{3) Regularization}

In addition to the loss-based pruning methods, regularization methods constitute another category within the spectrum of structured pruning techniques applicable to medium-size language models. 
Diverging from unstructured pruning approaches, the regularization term in structured pruning encompasses binary masks associated with specific structural components, as opposed to individual weights. 
Except for the pruning units, other details closely resemble those in unstructured pruning.

Nevertheless, among these regularization methods, $L_0$ regularization stands out as the most extensively employed technique. 
The main variability among these $L_0$ regularization methods resides in their respective approaches to the selection of pruning units.
Voita et al. \cite{voita2019analyzing} introduced $L_0$ regularization to attention head pruning, specifically selecting a subset of attention heads.
McCarley et al. \cite{mccarley2019structured} incorporated $L_0$ regularization to prune attention heads and FFN neurons. 
Factorized Low-rank Pruning  (FLOP) \cite{wang2019structured} integrates Low-Rank Factorization with $L_0$ regularization.  
This methodology involves the reparameterization and factorization of the matrix $\textbf{W}$ into the product of two smaller matrices, denoted as $\textbf{W} = \textbf{PQ}$, where $\textbf{p}_k$ and $\textbf{q}_k$ represent the $k$-th column of $\textbf{P}$ and $k$-th row of $\textbf{Q}$ respectively. The pruning unit is the combination of $\textbf{p}_k$ and $\textbf{q}_k$.
Additionally, an augmented Lagrangian method is introduced to regulate the sparsity ratio in the context of FLOP.
Coarse- and Fine-grained Pruning  (CoFi)\cite{xia2022structured} jointly prunes coarse-grained and fine-grained modules using $L_0$ regularization, including attention and FFN layers, individual attention heads, FFN neurons, and hidden dimensions for Transformer-based models. Notably, the mask over the hidden dimension is shared across all Transformer layers and an augmented lagrangian method is adapted. By combining with a layerwise distillation approach, CoFi achieves models with more than 10× speedups while exhibiting only a marginal decrease in accuracy in the conducted experiments.

In addition to $L_0$ regularization, $L_1$ regularization also gets relevant research. 
SIMPLE \cite{tao2023structured} introduces $L_1$ regularization to structured pruning, encompassing attention heads, intermediate neurons of the FFN, and the hidden dimension as compressible components. The mask over the hidden dimension is shared across layers, akin to the approach employed in CoFi\cite{xia2022structured}. Through the learning of masks for these compressible components via a sparsity-induced objective, various-sized pruned models can be obtained. These pruned models can subsequently be fine-tuned with a causal distillation objective to enhance performance. 

\textbf{4) Others}

Beyond the classification based on metrics, certain structured pruning methods exhibit notable similarities when their designated pruning units are identical.

The first class among other structured pruning is layer pruning\cite{fan2019reducing,zhang2020accelerating,sajjad2023effect}. The aforementioned pruning units, such as attention heads and neurons, are characterized by their relatively diminutive scale, necessitating a more detailed pruning scheme to determine which should be pruned. Conversely, when dealing with substantially larger pruning units, such as entire layers, numerous methodologies tend to engage in direct experimentation with multiple pruning schemes before determining the most effective network configuration. This practice stems from the lower testing costs associated with a smaller number of layers.

In addition to layer pruning, there is a body of research dedicated to token pruning \cite{goyal2020power, kim2022learned, wang2021spatten}, which does not alter the underlying network architecture. Token pruning involves the removal of unimportant tokens from a sequence during inference to reduce computational requirements.
Learned Token Pruning  (LTP) \cite{kim2022learned} represents a straightforward and effective approach to adaptively remove unimportant tokens as an input sequence traverses through transformer layers. The pruning metric for each token is determined by the sum of normalized attention probability from the Transformer block. 

Extending beyond the pruning units previously mentioned, structured pruning encompasses a myriad of diverse units. For instance, Spectral-Normalized Identity Prior  (SNIP) \cite{lin2020pruning} employs a strategy to prune attention and FFN sublayers by transforming residual connections into strict identity mappings. 
SNIP sets specific thresholds for activation vectors, and those falling below the thresholds result in the pruning of residual blocks  (\ie, the attention and FFN sublayers). 

\subsection{Pruning Methods for LLMs}
\label{pru: llm}

In the last section,  we introduced the pruning methods for medium-size language models with parameters numbering less than 1 billion.
Most of these methods adopt full fine-tuning after pruning to improve the performance. 
However, as the parameters increase, full fine-tuning becomes more difficult or even infeasible.
This discrepancy underscores a significant challenge in the field of research dedicated to pruning techniques tailored specifically for LLMs.
To handle this problem, on the one hand, certain pruning methodologies opt to incorporate parameter-efficient tuning techniques to reduce fine-tuning costs. 
On the other hand, alternative approaches abandon the fine-tuning process, relying on an optimized pruning procedure will inherently lead to retained model performance. 
The viability of these alternative approaches is partly attributed to the huge number of parameters in LLMs. The higher number implies a higher likelihood of redundancy within the model.

In this section, we will introduce the pruning methods for LLMs, mirroring the sequence established in the section devoted to pruning methods for medium-size language models. Pruning methods for LLMs adhere to a parallel approach to those employed for medium-size language models, with some distinctions in certain methods primarily arising in omitting the fine-tuning process.
To facilitate a more comprehensive comparison of these methods, we consolidate the characteristics of these pruning methods, as shown in TABLE~\ref{pru_table}.

\begin{table*}[!t]
\renewcommand{\arraystretch}{1.3}
\caption{
A summary of various pruning methods for LLMs.
}
\label{pru_table}
\centering
{\footnotesize
    \begin{tabularx}{\textwidth}{
    | 
    >{\setlength{\hsize}{.4\hsize}\centering\arraybackslash}X
        | >{\setlength{\hsize}{.25\hsize}\centering\arraybackslash}X
        | >{\setlength{\hsize}{.35\hsize}\centering\arraybackslash}X
        | >{\setlength{\hsize}{.4\hsize}\centering\arraybackslash}X
        | >{\setlength{\hsize}{.3\hsize}\centering\arraybackslash}X
        | >{\setlength{\hsize}{.3\hsize}\centering\arraybackslash}X
        |}
    \hline
    \textbf{Methods} & \textbf{Unit} & \textbf{Metric} & \textbf{Iterative/One-shot} & \textbf{Finetuning} & \textbf{Global/Local}  \\
    \hline
    Wanda\cite{sun2023simple}  & Unstructured & Magnitude-based &  One-shot & No & Local \\
    \hline
     RIA\cite{zhang2023efficient} & Unstructured & Magnitude-based &  One-shot & No & Local \\
    \hline
     E-Sparse\cite{li2023sparse}&Unstructured &  Magnitude-based &  One-shot & No & Local \\
    \hline
    SparseGPT\cite{frantar2023sparsegpt}  & Unstructured  & Loss-based  & One-shot & No & Local\\
    \hline
    ISC\cite{shao2023one}  &  Unstructured  & Loss-based  & One-shot & No & Local\\
    \hline
    GBLM-Pruner\cite{das2023beyond}&  Unstructured  & Loss-based  & One-shot & No & Local\\
    \hline
    PGZ\cite{anonymous2024pushing}&  Unstructured  & Loss-based  & One-shot & No & Local\\
    \hline    FLAP\cite{an2023fluctuation}&Structured&Magnitude-based&One-shot&No&Global\\
    \hline
    SliceGPT\cite{ashkboos2024slicegpt}&Structured&Magnitude-based&One-shot&PEFT&Local\\
    \hline
    LLM-Pruner\cite{ma2023llm}&Structured  &Loss-based  &One-shot   & PEFT  & Global 
    \\
    \hline
    LoRAShear \cite{chen2023lorashear}&Structured  &Loss-based & Iterative & PEFT & Global \\
    \hline
    APT\cite{zhao2024apt}&Structured&Loss-based&Iterative&PEFT&Global\\
    \hline
    Sheared LLaMA \cite{xia2023sheared}  & Structured & Regularization &  One-shot  & Yes & Local \\
    \hline
    Compresso\cite{guo2023compresso}& Structured & Regularization &  Neither  & PEFT & Global\\
    \hline
    LLM Surgeon\cite{van2023llm}&Both&Loss-based&Iterative& PEFT & Global\\
    \hline
    \end{tabularx}
}
\end{table*}

\subsubsection{Unstructured Pruning for LLMs}

Attributed to the greater capacity of unstructured pruning methods to preserve model performance compared to structured alternatives, all of the unstructured pruning methodologies in this section for LLMs adopt an approach of eschewing the fine-tuning process as shown in TABLE~\ref{pru_table}. 
The experiments have demonstrated that these methodologies can attain a sparsity ratio of 50\% with a relatively modest compromise in model performance. 

The two pioneer unstructured pruning methods for LLMs are SparseGPT\cite{frantar2023sparsegpt} and Wanda\cite{sun2023simple}, which become the baselines for many subsequent methods for comparison. 
The subsequent unstructured pruning methods demonstrate their capability to outperform SparseGPT and Wanda across various NLP tasks, thereby attaining superior results.
Though unstructured pruning methods get hardly inference speedup, they can easily be combined with N:M sparsity\cite{mishra2021accelerating} to accelerate inference speed, which is also experimented in SparseGPT and Wanda. 

These unstructured pruning methods require minimal calibration data. 
The minimal calibration data is for a single forward pass of the model, specifically aiming at acquiring activation values or gradients to calculate the importance of weights, which remains a contributing factor to the outcome of the pruning\cite{williams2023does}.

In the following, we will introduce these unstructured pruning methods in LLMs in the order of pruning metrics. In this investigation, no regularization-related methods have been identified, thus this section will be divided into introductions of methods based on magnitude and methods based on loss.

\textbf{1) Magnitude-based Pruning}

When directly applying Magnitude Pruning \cite{han2015learning} to LLMs, the outcomes are not very competitive even with parameter-efficient fine-tuning strategies\cite{zimmer2023perp,gholami2023can}.
Therefore, in magnitude-based pruning methods, compared to only using the magnitude of weights as the pruning metric in medium-size language models, more magnitude-based pruning methods in LLMs combine the magnitude of weights and activate values as the pruning metric. 
For instance, Wanda\cite{sun2023simple} and RIA\cite{zhang2023efficient} use the magnitude of weight and activation metric. In addition to the magnitude of weight and activation, E-Sparse\cite{li2023sparse} also introduces the information entropy into the metric.

Wanda  (Pruning by Weights and activations)\cite{sun2023simple}   introduces a novel pruning metric, considering both the magnitude of weights and activate values. 
The motivation is that the significance of weights should not solely be evaluated in isolation but rather in consideration of its product with the corresponding activation value. 
To illustrate, let's consider a fully connected layer with weights represented by $\mathbf{W}$ with dimensions  $ (C_{out}, C_{in})$. In the context of language models, this linear layer receives input activation $\mathbf{X}$ with dimensions $ (N \times L, C_{in})$, where $N$ and $L$ denote the batch and sequence dimensions respectively. For each weight, its importance is quantified as the product of its magnitude and the corresponding input feature norm. Concretely, the score $\mathbf{S}_{ij}$ for the weight $\mathbf{W}_{ij}$ is defined as:
\begin{equation}
    \mathbf{S}_{ij}=|\mathbf{W}_{ij}|\cdot\Vert \mathbf{X}_j \Vert_2,
\end{equation}
where  $\Vert \mathbf{X}_j \Vert_2$ evaluates the $L_2$ norm of $j$-th features aggregated across $N\times L$ different tokens. 
Remarkably, the results indicate that Wanda achieves comparable performance to SparseGPT but in a significantly shorter time.

Similar to Wanda\cite{sun2023simple}, RIA  (Relative Importance and Activations)\cite{zhang2023efficient} also jointly considers the weight and activation.
The primary distinction lies in its approach to alleviating channel corruption  (\ie, the rows and columns of the weight matrix pruned integrally). RIA replaces the magnitude of weights with relative importance. This relative importance is calculated as the magnitude of individual weights divided by the sum of the magnitude of weights in their corresponding row and column. 
Therefore, the comparison among different rows and columns becomes relatively equitable by utilizing the relative importance, mitigating potential biases introduced by the variations in their magnitudes.
RIA can be further combined with channel permutation, which maximally preserves important weights under N:M sparsity to get practical speed-up on specific hardware. 

In addition to the magnitude of weight and activation as Wanda and RIA, E-Sparse  (Entropy-based Sparsity) \cite{li2023sparse} introduces information entropy from hidden state features into the pruning metric. The entropy serves as a measure of information richness, with higher values indicating richer information. Consequently, entropy is incorporated alongside standard weight magnitude and input feature norm in the pruning metric, enhancing the evaluation of channel information activation. 

\textbf{2) Loss-based Pruning}

In loss-based approaches, it is observed that the pruning metrics involve the first or second-order derivatives of weights with respect to the loss. 
The second-order methods discussed in this subsection are all inspired by two earlier second-order loss-based pruning methods, namely, Optimal Brain Damage  (OBD)\cite{lecun1989optimal} and Optimal Brain Surgeon  (OBS)\cite{hassibi1993optimal}.

SparseGPT\cite{frantar2023sparsegpt}, a second-order pruning method,  incorporates OBS\cite{hassibi1993optimal} technique into the GPT-family models. It is the first pruning method that works efficiently at models with 10-100+ billion parameters. 
The SparseGPT pruning methodology is delineated by two main components: mask selection and weight reconstruction processes.
Initially, the mask selection identifies weights for pruning based on a metric, such as weight magnitude. Subsequently, the unpruned weights undergo optimization using the OBS method to reconstruct the compressed model  (\ie, update the remaining parameters) to compensate for the pruned weights.
The pruning procedure in SparseGPT requires minimal calibration data.
These data undergo a single forward propagation, during which the unpruned weights are updated only once. 
The results of this approach demonstrate that LLMs can be compressed to high sparsity through weight pruning in a single pass, without necessitating the fine-tuning process. Importantly, this compression is achieved with a low loss of accuracy, as assessed by perplexity and zero-shot performance metrics. Similarly, the LLM Surgeon\cite{van2023llm} extents OBS but is generic for unstructured and structured pruning.

Building upon the concepts of OBS and OBD, Shao et al. \cite{shao2023one} introduced a novel pruning metric termed the Improved Saliency Criterion  (ISC). ISC is devised by adding the metrics derived from OBS and OBD directly. This new metric aims to provide a comprehensive and refined assessment of the importance of model parameters for the pruning process.
In addition to proposing ISC, Shao et al. put forward to allocate sparsity ratio individually to each matrix. In this way, pruning targets are selected
adaptively within each weight matrix. 


In addition to the aforementioned second-order methods, there has been corresponding research into first-order methods\cite{das2023beyond,anonymous2024pushing}. 
Gradient-based Language Model Pruner  (GBLM-Pruner)\cite{das2023beyond} is a first-order pruning method. 
The importance of weights is defined by the product with the magnitude of weights and the normalization of the corresponding gradients across different samples, which can be seen as an extension of the traditional first-order method  (\ie, gradient-weight product).
Furthermore, the feature activations can be integrated into the pruning metric to enhance performance. 



\subsubsection{Structured Pruning for LLMs}
In contrast to unstructured pruning, structured pruning is not constrained by hardware limitations, enabling the realization of inference acceleration on conventional hardware following the pruning process.
However, these methods might result in more performance degradation than unstructured ones due to the alteration of network structures, necessitating a fine-tuning process to recover performance. 
Therefore, while fine-tuning is abandoned in unstructured pruning for LLMs, it is widely employed in structured pruning for LLMs but in a parameter-efficient way. 
Similar to unstructured pruning, structured pruning for LLMs has its pioneer method, LLM-Pruner \cite{ma2023llm}, which serves as a baseline for subsequent methods and facilitates meaningful comparisons. 

The discussion of these structured pruning methods for LLMs will be presented in the following section.
Similarly, we will introduce these structured pruning methods in LLMs in the order of pruning metrics, including magnitude-based pruning, loss-based pruning, and regularization.

\textbf{1) Magnitude-based Pruning}

Magnitude-based pruning methods for LLMs consider rows or columns as pruning units\cite{an2023fluctuation,ashkboos2024slicegpt,valicenti2023mini}.
For instance, the pruning units of FLuctuation-based Adaptive Structured Pruning (FLAP)\cite{an2023fluctuation} are columns. The importance score of each column of the weight matrix is measured by the "fluctuation metric". 
This metric is the sample variance of each input feature which is weighted with the squared norm of the corresponding column of the weight matrix.
Furthermore, in its pursuit to obviate the necessity for fine-tuning, FLAP incorporates bias compensation mechanisms aimed at mitigating the adverse effects stemming from the removal of components.

\textbf{2) Loss-based Pruning}

In the realm of loss-based structured pruning methods applied to LLMs, gradients remain pivotal information, akin to their significance in medium-size models.
The following methods utilize gradient information in different ways\cite{ma2023llm,chen2023lorashear,ji2023pruning,zhao2024apt}, such as defining pruning structures, selecting pruning targets, and \etc. 
The most notable departure of these methods from traditional approaches lies in their avoidance of predefined pruning units  (\eg, attention heads, neurons). Instead, some of these methods dynamically identify and designate pruning units.

For instance, LLM-Pruner \cite{ma2023llm} removes non-critical coupled structures during the pruning process. 
These coupled structures are automatically identified and extracted through the definition of structure dependency  (\ie, connection dependencies between neurons). 
A coupled structure comprises a group of weights. The importance of individual weights is formulated as the change in loss, expanded using Taylor expansion to the second order. The diagonal of the Hessian matrix in the second-order term is approximated by the Fisher information matrix using first-order information. Ultimately, the importance of a group of weights is aggregated through summation, production, or other methods to determine the group's overall importance. After evaluating the importance of each group, those with lower importance are pruned based on a predefined pruning ratio. 
The fine-tuning process in LLM-Pruner applies some parameter-efficient tuning techniques, such as LoRA. 
This facilitates rapid and effective fine-tuning of pruned models using a small amount of data.
The experimental results showcase that when 20\% of the parameters are removed, the pruned model maintains the performance of the majority of the original model. However, a more aggressive pruning strategy, involving the removal of 50\% of the parameters, results in a substantial decline in model performance. This observation also underscores the difficulty of achieving high sparsity ratios through structured pruning while maintaining model performance. 

Similar to LLM-Pruner, LoRAShear \cite{chen2023lorashear} discovers the minimal removal structures in the dependency graph. However, LoRAShear specifically constructs dependency graphs over LoRA modules, considering their learnable nature. The analysis of knowledge distribution is then utilized to identify crucial structures, marking them as unprunable.
A distinctive feature of LoRAShear is the introduction of LoRA Half-Space Projected Gradient  (LHSPG) for progressive structured pruning. LHSPG leverages information from LoRA modules to identify and remove redundant structures while preserving the knowledge stored in the important structures. This is achieved through the projection of redundant structures onto zero, transferring the knowledge to the crucial structures.

In contrast to the manual design of pruning features, Ji et al. \cite{ji2023pruning} proposed a novel approach by employing a non-neural model, specifically a gradient boosting decision tree  (GBDT), as an accuracy predictor. The use of this accuracy predictor enables further optimization of the search space and search process for identifying the optimal pruned model automatically.
By training the GBDT as an accuracy predictor, the model gains the ability to assess and predict the impact of different pruning configurations on the accuracy of the neural network, facilitating more efficient and automated selection of the optimal pruned model. 

\textbf{3) Regularization}

In the context of regularization methods applied to LLMs, contemporary approaches predominantly adhere to the principles established for earlier medium-size language models, incorporating some generic refinements and optimizations.

Sheared LLaMA \cite{xia2023sheared} can be viewed as an extension of CoFi \cite{xia2022structured}. This approach involves the joint pruning of coarse-grained and fine-grained modules in Transformer-based models using $L_0$ regularization. The modules subjected to pruning include layers, individual attention heads, FFN neurons, and hidden dimensions as in CoFi. 
Sheared LLaMA introduces two novel and significant components. The first component is targeted structured pruning, which frames pruning as a constrained optimization problem. This formulation aims to learn pruning masks that search for a subnetwork matching a pre-specified target architecture while maximizing performance.
The second component is dynamic batch loading, a strategy that loads training data from each domain in proportion to its rate of loss reduction. This approach efficiently utilizes data and accelerates overall performance improvement during training.
In a full-resource setup, Sheared LLaMA achieves compact counterparts that outperform models of equal sizes trained from scratch. 

Compresso \cite{guo2023compresso} integrates LoRA into the $L_0$ regularization. The $L_0$ regularization is employed to optimize binary masks that cover modules including heads, FFN intermediate neurons, and hidden dimensions. Simultaneously, model parameters are updated through LoRA in the instruction tuning process.
An innovative aspect of Compresso is the introduction of a collaborative pruning paradigm where the pruning algorithm and target LLM work together through a collaborative prompt to learn the optimal pruning decisions during the instruction tuning process. 
The prompt explains the concept of pruning and its purpose, informs the LLM that it is undergoing pruning, and encourages the LLM to better adapt to the pruning process. 
By incorporating this informative prompt, Compresso aims to enhance the LLM's understanding and cooperation during pruning, contributing to improved performance and adaptation to the modified model structure.


\subsection{Other Topics for LLM pruning}
\label{pru: topic}
\subsubsection{Enhancing Pruning Efficacy for LLMs}
Several auxiliary techniques have been developed to enhance the efficacy of pruning methods tailored for LLMs, including the sparsity ratios tailored for subregions\cite{anonymous2023outlier,anonymous2024besa}, post-pruning fine-tuning methods\cite{syed2023prune,kurtic2023sparse,zhang2023dynamic,zimmer2023perp,boža2024fast}, and hardware optimization\cite{xia2023flash,srinivasan2023training}
While not constituting a novel pruning method, these auxiliary techniques can readily be integrated with existing pruning methods for LLMs to enhance overall pruning outcomes.

One such method of tailored sparsity ratios is Outlier Weighed Layerwise sparsity  (OWL) \cite{anonymous2023outlier}.
The experiments in OWL indicate that the appropriate layerwise sparsity ratios have a strong correlation with the emergence of outliers.
Therefore, 
the sparsity ratio of OWL is directly proportional to the outlier ratio observed within each layer.
Consequently, in contrast to the prevailing LLM pruning strategies that uniformly apply sparsity levels across all layers, OWL introduces a customized set of non-uniform layerwise sparsity ratios. 
Another approach of post-pruning fine-tuning methods is Dynamic Sparse No Training \cite{zhang2023dynamic}, which introduces a training-free fine-tuning method for sparse LLMs. This allows for slight updates to sparse LLMs, enabling further refinement without the need for a complete fine-tuning process.
Without the expensive backpropagation, Dynamic Sparse No Training minimizes the reconstruction error between the dense and sparse LLMs, in the fashion of performing iterative weight pruning and growing on top of sparse LLMs.

The experimental results demonstrate that these techniques can significantly improve the performance of existing pruning methods, such as Wanda and SparseGPT. 
These findings suggest that 
the potential enhancements to the performance of pruning can be achieved through various means unrelated to the cores of the pruning methods.

\subsubsection{Future Works of Pruning for LLMs}

While the field of pruning for LLMs has yielded fruitful results, it continues to grapple with significant challenges.
Two primary issues stand out as particularly crucial.

Firstly, the integration of pruning with other methodologies, such as quantization\cite{frantar2023sparsegpt} and knowledge distillation\cite{xia2023sheared}, is essential for achieving competitive performance. 
Relative to the achievements of pruning in the domain of visual models in the past, the current outcomes in LLM pruning are comparatively less satisfactory. Therefore, a pivotal challenge lies in augmenting the inherent effectiveness of the pruning method, ensuring its proficiency even when employed independently.

Secondly, the fine-tuning cost is a significant challenge in the pruning of LLMs. Many pruning methods for LLMs adopt one-shot pruning without fine-tuning to minimize the computational burden. Alternatively, some approaches incorporate parameter-efficient tuning techniques to reduce training costs. However, such strategies inevitably compromise the performance of the pruned model.
Researchers and practitioners in the field must persist in addressing the challenge of the inability to execute full fine-tuning, particularly when dealing with LLMs aiming to enhance the performance of pruning.

In conclusion, addressing these challenges is imperative for advancing the effectiveness and practicality of pruning techniques.

%% file: knowledge_distillation.tex
Knowledge Distillation (KD) is a common technique for compressing and speeding up models. The specific implementation process involves transferring the knowledge acquired by a \textit{complex teacher model} to a \textit{simpler student model}, thereby enabling a more concise and efficient representation of the teacher model's knowledge.

In Section~\ref{basicKD}, we will introduce some fundamental concepts of knowledge distillation and provide a brief classification of knowledge distillation methods. Then we will summarize various knowledge distillation methods employing medium-size language models (the language models with around 1 billion parameters) in Section~\ref{mediumKD}, and we will classify them into three groups based on whether distillation occurs during the pretraining phase, the finetuning phase, or both. We finally provide a detailed overview of knowledge distillation for large language models (the language models with over 1 billion parameters), categorizing them as black-box distillation and white-box distillation in Section~\ref{largeKD}.

\subsection{Basic Concepts}
\label{basicKD}

Understanding the core of knowledge distillation involves answering three questions: what is knowledge, between whom is knowledge transmitted, and how is knowledge transmitted. Knowledge, in simple terms, is summarized as the abilities the models possess (classification, reasoning, \etc). In the distillation process, the source of knowledge is the teacher model, and the recipient of knowledge is the student model. In other words, a well-trained teacher is essential, and our goal is to enable the student to acquire or reinforce the abilities the teacher possesses. However, the key lies in how knowledge is transmitted. The pioneers of knowledge distillation, Hilton et al.\cite{hinton2015distilling}, first used the outputs of the teacher and student's softmax layers to transmit knowledge. They designed the following loss function to train the student model, thereby achieving the transfer of knowledge:
\begin{equation}
  L = \alpha \cdot L_{D}(p(z_{t},T),p(z_{s},T)) + (1-\alpha) \cdot L_{S}(y,p(z_{s},T))
\end{equation}
where $L_{D}(p(z_{t},T),p(z_{s},T))$ represents the difference in output of the softmax layers between the student and teacher, $L_{S}(y,p(z_{s},T))$ represents the difference between the output of the student's softmax layers and the ground-truth labels. Both of them utilize the cross-entropy loss. $\alpha$ represents the weight coefficient, and the specific expression for $p_{i}$ is as follows:
\begin{equation}
  p_{i} = \frac{exp(z_{i}/T)}{\Sigma_{j}exp(z_{j}/T)}
\end{equation}
where T is employed to amplify the impact of incorrect labels on the transmission of knowledge, thereby enabling the student model to acquire more knowledge from a single sample.

\begin{figure}[!t]
    \centering
    \includegraphics[width=3in]{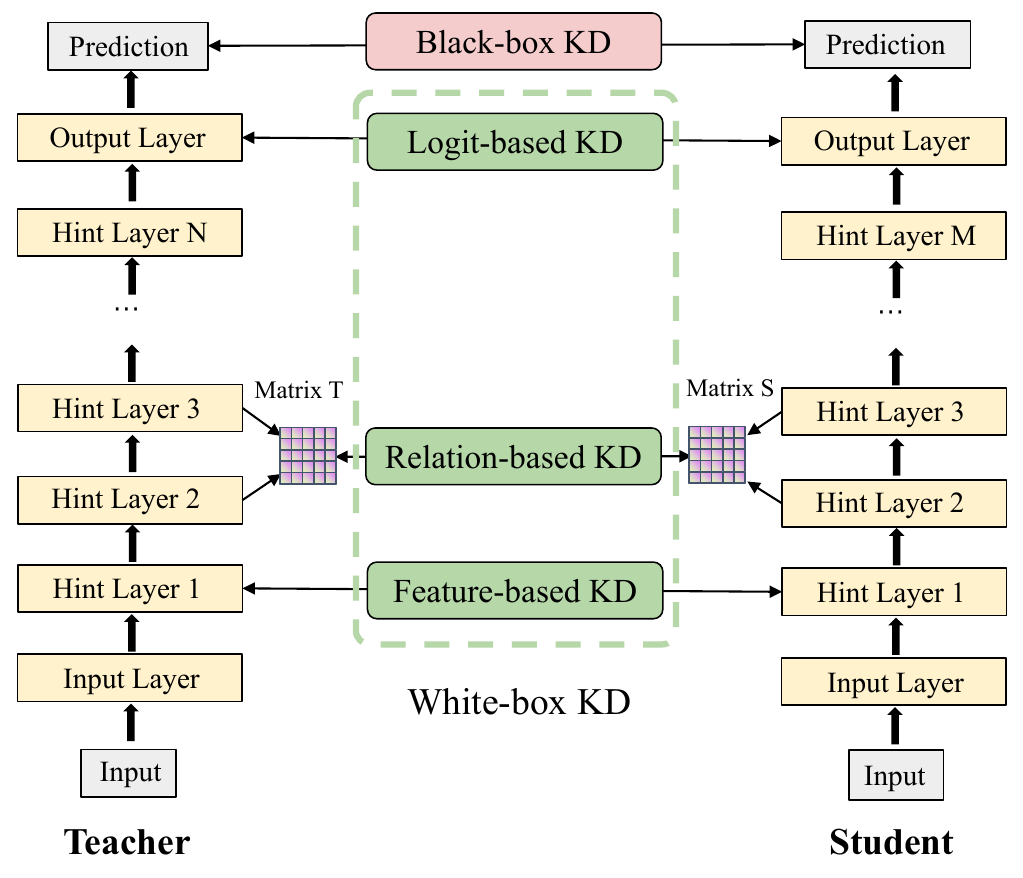}
    \caption{Taxonomy of knowledge distillation}
    \label{KD classfication}
\end{figure}

Subsequent researchers have employed a variety of methods to achieve knowledge transfer, primarily falling into the following four categories: logit-based KD, feature-based KD, relation-based KD and black-box KD. In Fig.~\ref{KD classfication}, we also provide a brief overview of these distillation methods and their relationships.

1) \textit{Logit-based KD}. As the name suggests, logit-based KD is a distillation paradigm that involves the transfer of knowledge using the logits from the teacher model. We can write down the general form of the logit-based knowledge distillation loss function:
\begin{equation}
  {L}_{logit} = \mathcal{L}(p(z_{t}),p(z_{s}))
\end{equation}
where $\mathcal{L}(\cdot)$ indicates the cross-entropy loss\cite{hinton2015distilling}, Kullback-Leibler divergence (KLD) loss\cite{zhang2018deep} and so on.

Clearly, Hilton et al.'s method is a example of logit-based knowledge distillation.

2) \textit{Feature-based KD}. Due to the limited knowledge acquired by the student in logit-based knowledge distillation, researchers aim for better emulation of the teacher's behavior. Hence, they introduced feature-based knowledge distillation. Specifically, this involves matching the outputs of intermediate layers in both the student and teacher models, requiring the student not only to know the results but also to understand the underlying processes. The following is the general form of the loss function for feature-based knowledge distillation: 

\begin{equation}
  {L}_{feature} = \mathcal{L}((f_{t}(x),r(f_{s}(x)))
\end{equation}
where $f_{t}(\cdot)$ and $f_{s}(\cdot)$ represent the feature maps of the teacher model and the student model. $\mathcal{L}(\cdot)$ is the function used to fit features, and $r(\cdot)$ is applied to make feature maps of the teacher model and the student model have the same shape.

For example, FitNet\cite{romero2014fitnets} leverages feature maps from intermediate layers of both the teacher and student models to adjust the parameters of the student model. It also uses mean squared error (MSE) along with a learnable matrix as $\mathcal{L}(\cdot)$ and $r(\cdot)$. 

3) \textit{Relation-based KD}. Furthermore, researchers aim for the student to learn how the teacher handles relationships between different data, leading to the proposal of relation-based knowledge distillation. This relationship is primarily manifested in two aspects: the relationship between outputs at different layers for the same sample and the relationship between outputs for different samples. The general form of its loss function is as follows:
\begin{equation}
  {L}_{response} = \mathcal{L}((f_{t}(t_{i},t_{j}),f_{s}(s_{i},s_{j}))
\end{equation}
where $t_{i}$, $t_{j}$ and $s_{i}$, $s_{j}$ are feature representations from the teacher model and the student model. They can represent outputs from different layers or outputs from different samples. $f_{t}(\cdot)$ and $f_{s}(\cdot)$ represent the similarity functions.

For example, FSP\cite{Yim2017AGF} uses feature maps of the same size as feature representations, and employs Gram matrix and MSE as $f(\cdot)$ and $\mathcal{L}(\cdot)$.

4) \textit{Black-box KD}. The three distillation methods mentioned above rely on the premise that internal information of the teacher model is accessible, so they all fall under the category of \textit{white-box distillation}(distillation method that requires access to internal data of the teacher model during the training process). However, many contemporary closed-source large models have inaccessible internal information, and we can only obtain the model's predictions. The distillation pattern where knowledge is transmitted through the predictions of the teacher model is referred to as black-box knowledge distillation.

\subsection{KD for Medium-Size Language Models}
\label{mediumKD}

\begin{table*}[!t]
\renewcommand{\arraystretch}{1.3}
\caption{A summary of various KD methods for BERT. Embed., Attn., Hidden., and Pred. represent knowledge is from embeddings, attentions, hidden layers, and model's prediction, repectively.}
\label{KD_table}
\centering
{\footnotesize
    \begin{tabularx}{\textwidth}{| >{\setlength{\hsize}{.4\hsize}\centering}X
        | >{\setlength{\hsize}{.5\hsize}\centering}X
        | >{\setlength{\hsize}{.12\hsize}\centering}X
        | >{\setlength{\hsize}{.12\hsize}\centering}X
        | >{\setlength{\hsize}{.12\hsize}\centering}X
        | >{\setlength{\hsize}{.12\hsize}\centering}X
        | >{\setlength{\hsize}{.62\hsize}}X
        |}
    \hline
    \textbf{KD Method} & \textbf{Training stage} & \textbf{Embed.} & \textbf{Attn.} & \textbf{Hidden.} & \textbf{Pred.} & \textbf{New Knowledge Source} \\
    \hline
    Distilled BiLSTM \cite{tang2019distilling} & Finetuning  &  &  &  & MSE &  \\
    \hline
    PKD \cite{sun2019patient} & Finetuning &  &  & MSE & CE &  \\
    \hline
    DynaBERT \cite{hou2020dynabert} & Finetuning & MSE &  & MSE & CE & \\
    \hline
    Metadistil \cite{zhou2021bert} & Finetuning &  &  &  & CE &  \\
    \hline
    AD-KD \cite{wu2023ad} & Finetuning &  &  &  & CE & Attribution map (MSE) \\
    \hline
    AdaBERT \cite{chen2020adabert} & Finetuning &  &  & CE & CE & Model efficiency \\
    \hline
    MixKD \cite{liang2020mixkd} & Finetuning &  &  &  & CE & MixUp data (CE/MSE) \\
    \hline
    Meta-KD \cite{pan2020meta} & Finetuning & MSE & MSE & MSE & CE & Transferrable knowledge (MSE) \\
    \hline
    ReAugKD \cite{Zhang2023ReAugKDRK} & Finetuning &  &  &  & CE & Similarity matrix (KL) \\
    \hline
    DistilBERT \cite{sanh2019distilbert} & Pretraining &  &  & COS & CE &  \\
    \hline
    MiniLM \cite{wang2020minilm} & Pretraining &  & KL &  &  & Value-relation (KL) \\
    \hline
    MobileBERT \cite{sun2020mobilebert} & Pretraining &  & KL & MSE & MSE &  \\
    \hline
    HomoBERT \cite{liang2023homodistil} & Pretraining & MSE & MSE & MSE & KL &  \\
    \hline
    TinyBERT \cite{jiao2019tinybert} & Finetuning and pretraining & MSE & MSE & MSE & CE &  \\
    \hline
    TED \cite{liang2023less} & Finetuning or Pretraining &  &  &  & KL & Filters (MSE) \\
    \hline
    \end{tabularx}
}
\end{table*}

With the emergence of the transformer architecture, various medium-size language models based on the transformer structure (e.g. BERT, GPT-2, \etc), have been proposed. These language models are trained through two training processes: pretraining and finetuning. Specifically, in the pretraining phase, we train the model on a large-scale unlabeled dataset to learn the general features and structure of language. Subsequently, during the finetuning process, we further train the model on labeled data to adapt it to the specific features and requirements of the given task. Consequently, unlike previous distillation methods, distillation for these models is categorized into two classes: finetuning distillation and pretraining distillation. The student model can receive knowledge transmitted from the pretrained teacher during the pretraining period or from the teacher fine-tuned for a specific task during the finetuning period. We will separately introduce these two distillation paradigms. Additionally, we have created the Table~\ref{KD_table} to illustrate the training stage, knowledge source and loss function for various medium-size model distillation methods mentioned below.

\subsubsection{Finetuning Distillation}

Finetuning distillation is primarily aimed at compressing models for specific tasks. Generally, teachers in finetuning distillation are models that have been fine-tuned for specific tasks. For example, Distilled BiLSTM\cite{tang2019distilling} is the earliest method to employ knowledge distillation on BERT. It transfers the knowledge of fine-tuned BERT to BiLSTM by learning from logits. Therefore, this is a successful implementation of logit-based knowledge distillation on 
medium-size models. Subsequently, many feature-based knowledge distillations\cite{sun2019patient,hou2020dynabert} have also been implemented on medium-size models. They distill knowledge in the embedding layer, transformer layers, and prediction layer, allowing the student model to learn the knowledge mastered by the teacher model from various aspects. For example, PKD\cite{sun2019patient} introduced a hidden state loss. It selects a subset of outputs from the intermediate transformer blocks of both the teacher and student for distillation. Additionally, it designed two alignment modes, namely PKD-Skip (the student learns from every k layers of the teacher) and PKD-Last (the student learns from the last k layers of the teacher), with experimental evidence demonstrating the superiority of the former. DynaBERT\cite{hou2020dynabert} also takes into account the width of the model, and it incorporates the idea of pruning. To be specific, It sets a parameter, the width multiplier $m_{w} \in (0,1)$, and retains the most important $m_{w}$ attention heads in the Multi-Head Attention (MHA) layer of the transformer, as well as the most important $m_{w}$ neurons in the Feed-Forward Network (FFN), to initialize the student model $\rm DynaBERT_{w}$. Then it transfers knowledge from the teacher model to the width-adaptive $\rm DynaBERT_{w}$ through the embedding layer, hidden states, and the prediction layer.  Following that, it uniformly selects transformer layers from $\rm DynaBERT_{w}$ using the depth multiplier $\rm m_{d}$ (similar to PKD-skip) to initialize the student model $\rm DynaBERT$. Knowledge is then transferred from $\rm DynaBERT_{w}$ to both the width-adaptive and depth-adaptive $\rm DynaBERT$ using the same knowledge source as in the width-adaptive process. Metadistil\cite{zhou2021bert} points out two common issues in general distillation: the teacher cannot perceive the student's abilities, and a strong teacher may not necessarily be effective in teaching good students. To address these problems, it proposes a novel distillation approach: first distill a copy S' of the student S on training data, then use the updated S' to update the teacher model on quiz data, allowing it to learn to teach. Finally, use the updated teacher model to distill S on training data. AD-KD\cite{wu2023ad} focuses on the importance of each token to the prediction results. It aims for the student model to understand which tokens the teacher model prioritizes when generating predictions, thus learning the rationale behind the teacher model's reasoning.

Some methods\cite{sun2019patient,hou2020dynabert} mentioned above can be applied to pretraining distillation from the perspective of operational feasibility, but Turc et al.\cite{turc2019well} has demonstrated that simple pretraining distillation methods result in significant distillation losses. Therefore, the effectiveness of the student models distilled using these methods may not be ideal. Besides, some methods\cite{zhou2021bert,wu2023ad} have not been utilized in pretraining distillation. The applicability of these methods to pretraining distillation remains to be explored.

Considering the fact that finetuning distillation is tailored for specific tasks, many other methods also utilize proprietary knowledge sources, enabling students to acquire knowledge these specific tasks need from the teacher model more efficiently. So, these methods cannot be applied to pretraining distillation. 

For instance, AdaBERT\cite{chen2020adabert} employs a search space to enable adaptive changes in the student's structure. Specifically, the search space consists of multiple layers, with each layer comprising input nodes, output nodes, and hidden internal nodes that form a directed graph. The edges of this graph represent candidate operations selected from a series of lightweight operations based on CNN. Considering the size and efficiency of the student model, AdaBERT incorporates not only soft and hard targets for distillation but also includes the normalized parameter size and number of floating-point operations of the student model in the loss function. Ultimately, this loss function is used to choose appropriate CNN-based operations. However, MixKD\cite{liang2020mixkd} starts with the dataset and applies MixUp\cite{zhang2017mixup} to KD in order to address the issue of limited training samples leading to insufficient knowledge acquisition by the student. It uses zero padding to make all sentences the same length, and then interpolates the word embeddings and labels of two training samples to obtain an augmented sample. Then it incorporates the loss of mixup samples into the loss function. Meta-KD\cite{pan2020meta} recognizes that when a student is learning in one domain, they may benefit from auxiliary knowledge in other domains. For example, a physics student may find it easier to grasp physics equations under the guidance of a teacher proficient in both physics and mathematics. Hence, training an "all-purpose teacher" model for domain-specific student models becomes essential. More precisely, it constructs a learnable sub-network using the output of the last hidden layer for each instance. This sub-network is capable of distinguishing the domain of each instance, making the knowledge transferable and not restricted by domain limitations. During the distillation process, the teacher is tasked not only with conveying knowledge encompassed by input embeddings, hidden states, attention matrices, and output logits but also with transmitting this transferable knowledge. ReAugKD\cite{Zhang2023ReAugKDRK} take the inference phase into consider. It uses an external memory derived from relevant task-specific knowledge of the teacher to enhance the effective capacity of the student. In the distillation phase, it adds a linear projection head, which has been fine-tuned for downstream tasks, on top of the teacher model's encoder to generate the teacher embedding and obtains the student embedding from the last transformer. Then it trains with a relational KD loss that minimizes the divergence between teacher-teacher and teacher-student embedding distributions. They found that this distillation method can effectively enhance the student model's ability to retrieve external information. In the inference phase, it constructs a knowledge base with the teacher’s soft labels and predictions. Then, it processes the top-k data entries from the knowledge base that are most similar to the student embedding. The final prediction is obtained by weighting and combining the student's predictions with these processed entries from the knowledge base. 

Besides, Enhanced KD\cite{Dasgupta2023CosteffectiveDO} proposes a new distillation loss function by expanding the loss in a Taylor series, which allows for effective distillation even when the teacher model is not fine-tuned for a specific task. This approach reduces a significant amount of training cost and architecture-agnostic.

\subsubsection{Pretraining Distillation}

The primary objective of pretraining distillation is to obtain a pretrained model with fewer parameters and good generalization capabilities. So some of them\cite{sanh2019distilbert,sun2020mobilebert,liang2023homodistil} utilize the loss function employed during the training of BERT. DistilBERT\cite{sanh2019distilbert} is the first to introduce pretraining distillation for BERT. It transfers the idea of PKD-skip\cite{sun2019patient} (the student learns from every k layers of the teacher) to pretraining distillation and employs the cosine similarity loss function to facilitate the transfer of knowledge within hidden states. MiniLM\cite{wang2020minilm} places the emphasis of distillation on the last transformer layer. It utilizes the self-attention distributions and self-attention value-relation (dot-product of the value matrix with itself) from this layer to acquire knowledge and perform distillation. This approach cleverly allows the student to have more flexible layer numbers and hidden dimensions. Hence, it can straightforwardly distill the teacher into a teacher assistant\cite{mirzadeh2020improved} with smaller hidden dimensions and then distill the teacher assistant into a student model with fewer layers, thereby enhancing the performance of the student model. MobileBERT\cite{sun2020mobilebert} and HomoBERT\cite{liang2023homodistil} put emphasis on model width as DynaBERT\cite{hou2020dynabert}, but they just alter models' width while preserving their depth because Turc et al.\cite{turc2019well} proves that the impact of depth on model performance is more significant. MobileBERT adds bottleneck and inverted-bottleneck to both the teacher and student models to alter the hidden dimensions. However, the practical implementation of this approach may disrupt the balance of parameters between Multi-Head Attention (MHA) and Feed-Forward Network (FFN). Therefore, the authors address this issue by adopting a stacked FFN approach. Then it distills knowledge through the attention and hidden states of transformer layers. HomoBERT utilizes the concept of pruning as DynaBERT. But it initializes the student with the teacher model so that it can maintain small discrepancy compared to the teacher model. Then it derives the distillation loss function using input embeddings, hidden states, attention matrices, and output logits as the pruning objective. In each iteration, it removes the least important neurons from the student based on importance scores and guides the student's training using the distillation loss. This process is iteratively repeated throughout the entire training until the student reaches the target size. TinyBERT\cite{jiao2019tinybert} combines pretraining distillation and finetuning distillation so that TinyBERT can capture the general-domain as well as the task-specific knowledge in BERT. It also distills various knowledge from the embedding layer, hidden states and attention matrices of transformer layers, and the prediction layer. But the ablation studies show that finetuning distillation has a more significant impact than pretraining distillation. TED\cite{liang2023less} equips each layer with a task-aware filter (a neural network with a task-specific head) to extract knowledge from the hidden representation of this layer. It has achieved promising results in both pretraining and finetuning scenarios.

\subsubsection{Discussion}

Finetuning distillation is computational costly because switching to a new task always requires the training of a task-specific teacher. So many finetuning knowledge distillation methods\cite{pan2020meta,Zhang2023ReAugKDRK,Dasgupta2023CosteffectiveDO} are proposed to reduce the computational cost of the finetuning process. But in pretraing distillation, student is distilled from a teacher pretrained on open-domain data and can be efficiently fine-tuned on various downstream tasks, which reduces the computational cost associated with distillation for multiple specific tasks to a certain extent. However, pretraining distillation also comes with many new challenges. For example,  teacher models have larger capacity and stronger representation capabilities than student models, it is challenging for students to produce predictions that match the teacher's on a large amount of open-domain training data. Therefore, for general methods, the choice between pretraining distillation and finetuning distillation depends on the trade-off we make between model size and performance.

\subsection{KD for Large Language Models}
\label{largeKD}

Recently, an increasing number of large language models(LLMs) have been developed. However, many of these large models are closed-source, which imposes significant limitations on knowledge distillation for such models. While the student model cannot acquire knowledge from internal information, we can still use the teacher model's responses, the remaining source of knowledge, to transfer information to the student model. Depending on whether the source of knowledge for the student model is solely the answers provided by the teacher model, distillation for large language models can be categorized into black-box distillation and white-box distillation.

\subsubsection{Black-box Distillation}

Even though conventional distillation methods may no longer apply, some unique properties of LLMs allow us to find a breakthrough. Researchers have found that when the models' parameter is large enough, they exhibits surprising emergent abilities, enabling it to tackle intricate tasks. Many black-box distillation methods leverage that abilities, and there are typically three methods commonly in use: Instruction-Following, Chain-of-Thought (CoT) and In-Context Learning.

\textbf{1) Instruction-Following}

Instruction-following capability means that the LLMs can generate corresponding outputs based on a specific instruction (directing the model on what task to accomplish) and the input (data required to fulfill that instruction). Due to the fact that black-box distillation can only transfer knowledge through datasets, it necessitates a sufficiently comprehensive dataset. Therefore, the common effort in this method\cite{wang2022self,peng2023instruction,wu2023lamini,jiang2023lion} involves constructing a large dataset (comprising instructions, inputs, and outputs) to enable the student models to learn as much as possible from the teacher models. Specifically, SELF-INSTRUCT\cite{wang2022self} employs a self-distillation approach, where the model serves both as the teacher and the student. It starts by obtaining a manually curated small-scale task pool, where each task consists of an instruction and a corresponding input-output pair. Subsequently, it selects a subset of instructions from the task pool as in-context examples for the model to generate new instructions and matching inputs and outputs for them. Finally, it filters out data with excessive redundancy or content that cannot be handled by language models, placing the qualified data back into the task pool. This iterative process continues to generate an extensive dataset for finetuning the student model. This has become a paradigm for instruction-following distillation, and the 13B open-source models Alpaca\cite{alpaca}, Vicuna\cite{vicuna2023} and GPT4All\cite{anand2023gpt4all} were trained with some adjustments based on this paradigm. Also, following this idea, LLaMA-GPT4\cite{peng2023instruction} and LaMini-LM\cite{wu2023lamini} construct their respective instruction sets and fine-tune smaller models. Compared to SELF-INSTRUCT, their breakthroughs are as follows: LLaMA-GPT4 generates a 52K instruction-following dataset in both English and Chinese using GPT-4, and fine-tunes two student models, LLaMA-GPT4 and LLaMA-GPT4-CN. Additionally, it trains a reward model specifically for evaluating the quality of model responses. LaMni-LM enriches the types of models used for generating instructions and the topics of instructions, constructing a massive dataset of 2.58M for finetuning smaller-parameter student models, which achieves good results. However, in the methods mentioned above, the student model is not involved in the selection of the dataset, so the teacher model cannot receive timely feedback from the student model during the dataset generation process. In response to this issue, Lion\cite{jiang2023lion} adopts adversarial knowledge distillation, where the student model not only learns from the teacher model's responses but is also evaluated by a referee to assess its difference compared to the teacher model. This helps to identify "hard" instructions where the student model’s performance falls short, thus generating new "hard" instructions so that teacher models can achieve feedback in the learning process. PERsD\cite{chen2023personalised} evaluates
the student’s attempt with unit test cases and gets execution feedback. Then it prompts the teacher
model to refine the student’s attempt so that the student can be trained on personalized data.

Some work focuses on task-specific instruction-following distillation. For instance, UniversalNER\cite{zhou2023universalner} conducts in-depth research on Named Entity Recognition (NER) tasks. So, unlike the methods mentioned above that increase the diversity of instructions, its emphasis is on enhancing the diversity of inputs to improve the model's generalization across multiple domains. To be specific, it directly samples inputs from a large corpus across diverse domains, and then uses a LLM to generate outputs. After obtaining the data, it trains the student model using a conversation-style tuning format, enabling it to identify entities of each entity type contained in the input text.

Furthermore, this approach of using large language models to construct reinforced datasets for finetuning student models is not unique to instruction-following but rather a common method of black-box distillation.

\textbf{2) Chain-of-Thought}

Chain-of-Thought capability refers to the ability of a large language model to provide better answers to questions based on the rationale within the given prompts. The typical paradigm of CoT \cite{li2022explanations,magister2022teaching,hsieh2023distilling,wadhwa2023revisiting} distillation utilizes large models to generate reinforced datasets containing rationales, which are then used to fine-tune the student model. Hence, the issues of interest revolve around how to generate high-quality rationales for training\cite{li2022explanations,ho2022large,shridhar2023distilling,wang2023scott,kang2023knowledge,jie2023leveraging,zhu2023pad,li2023symbolic,chen2023mcc,chae2023dialogue,wang2023democratizing} and how to ensure that students effectively leverage these rationales\cite{li2022explanations,hsieh2023distilling,shridhar2023distilling,wang2023scott,ma2023sci,kang2023knowledge,wang2023democratizing}.

Li et al. \cite{li2022explanations} systematically explores three explanation generation approaches from LLMs and three multi-task learning with explanations methods. Finally it finds that CROP (Chain of Thought with Rationalization Prompting backup) and MT-CoT (Multi-task Learning with Chain of Thought) are outstanding methods. In detail, CROP refers to a process where, for a dataset containing questions and answers, the teacher model first produces an explanation and an answer based on the question. If the answer is correct, the explanation is retained. If the answer is incorrect, the teacher model generates an explanation based on the question and the correct answer. Ultimately, a dataset is obtained with questions, explanations, and answers for finetuning the student model. MT-CoT refers to a training process for the student model with two tasks. The model is not only required to learn predicting answers but also to provide explanations. Moreover, in the task of providing explanations, the model needs to arrive at the correct answer through the reasoning steps it takes. Further, Distilling Step-by-Step\cite{hsieh2023distilling} demonstrates that good results can be achieved even when the original dataset only consists of questions without answers. Fine-tune-CoT\cite{ho2022large} applies existing zero-shot CoT prompting to generate rationales from large teacher models, and uses them to fine-tune smaller student models. It also proposes diverse reasoning to augment the training data for student models so that student models can have better performance. Besides, SCoTD\cite{li2023symbolic} and MCC-KDc\cite{chen2023mcc} also conducts in-depth explorations on the diversity of rationales. Fu et al.\cite{fu2023specializing} found that it is indeed possible to transfer the student model's capabilities from general tasks to tasks specific by employing CoT distillation. SOCRATIC CoT\cite{shridhar2023distilling} decomposes a question into several sub-questions to guide the generation of rationales. It starts by selecting a subset of data from the dataset, manually decomposing questions, and providing answers for each sub-question. These serve as examples given to the LLM to generate sub-questions and answers for the remaining data. The resulting dataset is reinforced by filtering based on the correctness of the final results. Two student models are then trained using this dataset, one for questioning and one for answering questions. SCOTT\cite{wang2023scott} takes into account two issues in CoT. Firstly, the rationale generated by the teacher model may not match the answer or be meaningful. Secondly, the student model may struggle to connect rationale and answer during learning. To address these challenges, SCOTT employs contrastive decoding during the rationale generation process to make the model pay more attention to the answer. This requires the teacher model's decoding process to be adjustable. In the training process of the student model, SCOTT introduces counterfactual rationales to guide the student in obtaining different answers, thereby establishing a closer relationship between rationale and answer. KARD\cite{kang2023knowledge} addresses the issue of limited memory capabilities in small models by retrieving information from external knowledge bases. Program Distillation\cite{jie2023leveraging} and PaD\cite{zhu2023pad} both leverage programs as rationales and have achieved promising results on math word problems. DOCTOR\cite{chae2023dialogue} utilizes a teacher model to generate question-answer-style rationales containing commonsense knowledge, and then filters and selects high-quality multi-hop reasoning for training students. Wang et al.\cite{wang2023democratizing} build an interactive multi-round learning paradigm, where the student first provides its learning status to the teacher LLM who then can provide customized rationales as the feedback to the student. They also exploit the reasoning potential of smaller LM by eliciting it to take self-reflection on the mistakes.

\textbf{3) In-Context Learning}

In-context learning (ICL) is also a manifestation of the emergent capabilities of large models, referring to the capacity of large models to generate correct outputs for new inputs based on some input-label examples without updating model parameters. Based on it, In-context Learning Distillation\cite{huang2022context} utilizes two few-shot learning paradigms, namely Meta In-context Tuning (Meta-ICT) and Multitask In-context Tuning (Multitask-ICT), to transfer the in-context learning capabilities of teacher models to student models by distillation. In Meta-ICT, it enables the student model to adapt to unseen tasks through in-context learning and assistance from the teacher. But in Multitask-ICT, it treats all target tasks as training tasks and directly employs examples from target tasks in in-context learning distillation. The results demonstrate that multi-task in-context tuning is more effective, although it comes with higher computational costs. LLM-R\cite{wang2023learning} initially trains a reward model based on LLM feedback to evaluate the quality of candidate examples, followed by knowledge distillation to train a retriever that can identify high-quality in-context examples for LLMs.

\textbf{4) Others}

In addition to the three paradigms mentioned above, there are other methods that generate specific reinforcement datasets to enable the student model to acquire specific capabilities. For instance, Symbolic Knowledge Distillation\cite{west2021symbolic} utilizes a LLM to gather data and filter it, thereby obtaining high-quality Commonsense Knowledge Graphs for training a Commonsense Model. DISCO\cite{chen2023disco} uses a LLM to obtain counterfactual data and employs a large teacher NLI model for filtering, thus obtaining a high-quality dataset to improve students' abilities in natural language inference (NLI) tasks. PubMedBERT\cite{gu2023distilling} conducts a case study on adverse drug event (ADE) extraction and proposes a novel framework that simultaneously handles adverse event (AE) entity extraction and ADE relation extraction to reduce computational requirements. Promptmix\cite{sahu2023promptmix} utilizes LLMs to mix and relabel text data for classification problems in proportion, aiming to obtain a stronger dataset for training.

However, Gudibande\cite{gudibande2023false} demonstrates that continually increasing imitation training data can lead to the model simply imitating without understanding, thus enhancing the capabilities of the base model is also an indispensable aspect of black-box distillation.

\subsubsection{White-box Distillation}

Compared to black-box distillation, the work on white-box distillation is relatively limited, but there is still some exploration. For example, MINILLM\cite{gu2023knowledge} and GKD\cite{agarwal2023generalized} both focus on the loss function and they find that forward KL divergence overestimates the void regions of the teacher distribution in language generation tasks when the student model distribution is insufficiently expressive to cover all the modes of teacher distribution. But reverse KL divergence focuses on the major modes, allowing the student to learn the main part of the teacher's distribution. Furthermore, it doesn't force the student to exactly match the teacher's distribution but aims to leverage the information provided by the teacher to assist in the student's training. So MINILLM samples from the student distribution and uses the Policy Gradient Theorem to calculate the reverse KL divergence. Also due to the high variance and reward hacking policy gradient suffers from, it comes up with the single-step regularization, teacher-mixed sampling and length normalization to solve these problems. Similar to MINILLM, GKD utilizes reverse KLD and Jensen-Shannon divergence (JSD) to enhance the student's expressive capacity. But it uses on-policy KD to alleviate the distribution mismatch between training and evaluation, which involves sampling from the student distribution without backpropagating through student's sampling process—something that MINILLM requires. It's proved that this gradient handling approach is relatively simple yet effective. Padmanabhan et al.\cite{padmanabhan2023propagating} generate a transfer set by prompting a language model to generate continuations from the entity definition and then update the model parameters so that the distribution of the student matches the distribution of the teacher on the transfer set. TSLD\cite{kim2023token} utilizes logit distillation to reform intermediate representations and applies token-wise logit scaling, reducing the errors introduced when QAT is applied to generative language models. MiniMA\cite{zhang2023towards} finds that the optimal distillation effect occurs when the student model is approximately 40\% of the size of the teacher model's parameters. It utilizes LLaMA2-7B for structured pruning and logit-based knowledge distillation to train a 3B MiniMA model.

Due to the limitations imposed by the closed-source nature of large language models, white-box distillation has faced constraints. However, with the emergence of increasingly diverse open-source large language models (e.g. Alpaca, Vicuna), white-box distillation holds significant promise for the future.

%% file: operators.tex

  

Compact architecture design is a philosophy that pursues efficiency and streamlining, and it aims to achieve a significant increase in model efficiency by optimizing the network structure and algorithms while reducing the consumption of computational resources and memory usage. Specifically, it can be divided into two levels of research: micro and macro. This section will focus on optimizing the attention computation and the Transformer architecture design. Since the Transformer layer is currently the main component of the LLM, and it makes no difference for large and medium-size models, so we will not specifically categorize methods by model size here.

\subsection{Efficient Attention}
The standard self-attention mechanism of the Transformer has a time and space complexity of $O(N^2)$ for sequence length $N$, which significantly limits its further expansion in various fields and prevents it from handling long-sequence problems. To solve this problem, many works have emerged to improve attention, many of which have focused on improving computational and memory efficiency. We refer to these works as \textbf{Efficient Attention}. Based on the starting point and method characteristics, we divide these works into three categories: \textbf{Sparse Attention}, \textbf{Linear Approximate Attention}, \textbf{and Flash Attention}.
There are also some unique works, such as Transformer-XL\cite{Transformer-XL}, which do not improve within the attention operator and, therefore, will not be discussed here.

\subsubsection{Sparse Attention}
\begin{figure}[!t]
    \centering
    \includegraphics[width=3.5in]{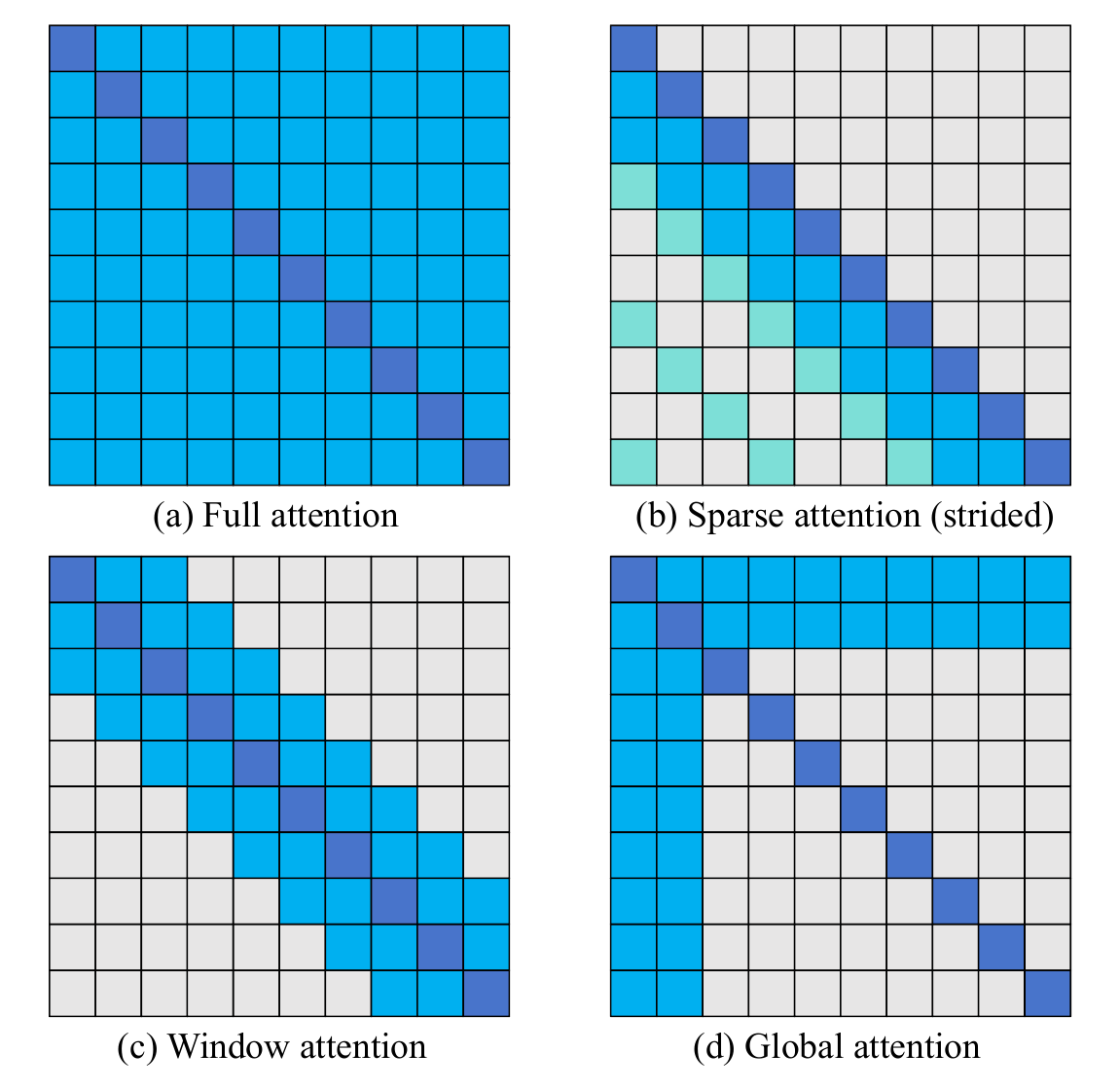}
    \caption{Comparing the sparse attention patterns. (a) full self-attention, (b) strided attention, (c) window attention, (d) global attention.}
    \label{SparseAttention}
\end{figure}
The Sparse Attention approaches\cite{SparseTransformers,AdaptiveAttentionSpan,AdaptivelySparseTransformers,Bp-transformer,BlockwiseAttention,Longformer,Reformer,BigBird,SparseSinkhornAttention,SAC,Combiner,RoutingTransformers} allow each token to attend only locally or predominantly relevant items to implement the sparse attention pattern, thus reducing computational and memory complexity. 
Based on the characteristics of these methods, we categorize them into stride-based, window-based, and data-based methods.

\textbf{1) Stride-based Methods}

The stride-based methods\cite{SparseTransformers, AdaptiveAttentionSpan, AdaptivelySparseTransformers} reduce computational complexity by having each token attend to several preceding tokens of length stride to achieve sparse attention patterns.

\cite{SparseTransformers} is an earlier work. It offered two ways to decompose attention: strided (Fig.~\ref{SparseAttention} (b)) and fixed attention patterns. These ways allowed each query to attend only to preset positions, reducing the complexity of self-attention to $O(N\sqrt{N})$. 
However, this method has limited applicability unless we can design appropriate sparse attention kernels for various scenarios.
By observing the distribution of attention across different heads in a standard Transformer, \cite{AdaptiveAttentionSpan} found that not all heads attend to the entire context (some heads only focus on the nearest tokens) and proposed learning dynamic attention spans for each head to reduce computational and storage costs. 
However, both previous works can only attend to past consecutive spans of tokens. To solve this problem, \cite{AdaptivelySparseTransformers} introduces $\alpha$-entmax to replace softmax, allowing low-rated words to receive precisely zero weight, thus enabling more flexible sparse attention. 

\textbf{2) Window-based Methods}

Unlike the above approaches, the window-based methods divide the input into individual localized windows so that each token only attends to items inside the window (Fig.~\ref{SparseAttention} (c)), thus reducing the computational complexity. 

An early method\cite{BlockwiseAttention} achieves a complexity of $O(\frac{N^2}{n})$ by dividing the $Q$, $K$, and $V$ matrices into $n$ blocks (padding is used if not divisible) and calculating attention within each block by shifting one position. This method is more straightforward to implement than Sparse Transformer. However, it is worth noting that $n$ is typically set to $2$ or $3$ in practice to maintain performance, which results in poor actual acceleration.
To solve this problem, \cite{Longformer} achieves $O(N(k+m))$ complexity by using a dilated sliding window to increase the receptive field without increasing the computation and improving the performance by adding "Global Attention" (Fig.~\ref{SparseAttention} (d)) to the pre-selected input positions. Here, $k$ represents the size of the sliding window, and $m$ represents the number of pre-selected positions.
However, its extended implementation requires efficient banded matrix multiplication support, and using the naive CUDA kernels can only have the running speed as standard self-attention.
Therefore, in practice, it only has a consistent memory footprint with theoretical complexity, but there is a gap between actual running speed and theoretical complexity. 
Similarly, \cite{BigBird} established a sparse attention pattern consisting of three main components:
\begin{itemize}
  \item Global attention: A global token set $g$ where tokens within the set attend to the entire sequence, and all tokens in the sequence attend to set $g$.
  \item Local attention: All tokens attend to a set $w$ of surrounding windows around themselves.
  \item Random attention: All tokens attend to a random token set $r$.
\end{itemize}
It is worth noting that there is also a gap between BigBird's actual running speed and theoretical complexity, similar to  \cite{Longformer}.

In addition to the methods described above, some methods\cite{Bp-transformer, Combiner} let each token attend directly to close and indirectly to distant locations. This method is very similar to the window-based methods introduced above but with a different indirect way to implement "Global Attention." Therefore, we present them here.

\cite{Bp-transformer} proposes a new architecture, BP-transformer (BPT), based on the prior observation that elements closer together have higher attention scores, while elements further away have lower attention scores. 
It treats attention calculation as a graph neural network and partitions the input sequence into different multi-scale spaces through binary partitioning (BP), constructing a binary tree-based attention pattern. Each leaf node represents a token, and each token focuses on different scale nodes based on the target distance, thereby reducing the complexity of attention to $O(kNlog(N/k))$ where $k$ is a hyperparameter controlling the density of attention.
The core idea of \cite{Combiner} is very similar to BPT in that it enables each token to attend to all other items directly or indirectly. The difference is that it treats the attention mechanism as a conditional expectation problem.

\textbf{3) Data-based Methods}

Unlike the above methods that need to design sparse patterns manually, data-based methods\cite{Reformer, RoutingTransformers, SAC, SparseSinkhornAttention} make each token automatically and quickly find the most relevant items to compute attention using appropriate algorithms. The most significant advantage of these methods is data-awareness, which effectively avoids the disadvantage of having to re-design the sparse patterns manually in the case of different tasks and data, and it isn't easy to obtain the optimal solution.

Reformer\cite{Reformer} achieves efficient sparse attention computation by using locally sensitive hashing to find similar vectors quickly, reducing the complexity to $O(Nlog(N))$. 
At the same time, Reformer also uses techniques such as reversibility layers and chunking in FFN layers to significantly reduce memory usage during training. However, this trade-off may also slow down the training speed. In addition, to avoid hash errors, Reformer requires multiple rounds of hashing, weakening its final efficiency benefits. 
Similarly to Reformer, \cite{RoutingTransformers} views self-attention as a routing problem. Specifically, it is based on k-means clustering, which allows queries and keys to cluster on the same set of cluster center-of-mass vectors by letting the model learn to select sparse clusters of word examples. So that each query $Q_i$ attends only to the keys that belong to the same cluster as it does. To ensure performance, it sets the number of clusters to $\sqrt {N}$, which reduces the attention complexity to $O(N\sqrt{N})$.

Other works related to sparse attention based on input are SAC\cite{SAC} and SSA\cite{SparseSinkhornAttention}. 
Among them, SAC regards the input as a graph and uses an LSTM edge predictor to learn the edges between tokens. The nodes in the graph represent tokens, and the edges represent attention relations. It also uses reinforcement learning to train this edge predictor. However, LSTM has limitations, such as a lack of parallelism and a limited ability to express long-term dependencies. There may be better methods available for building an edge predictor. 
On the other hand, SSA is based on the differentiable sorting of internal representations and introduces a meta-sorting network that can learn to generate potential orderings on sequences. It allows us to use only local windows for quasi-global attention after a given ordering sequence, improving the memory efficiency of the attention module.

\subsubsection{Linear Approximate Attention}
The standard attention can be represented as:
\begin{equation}
    {\rm Attention}(Q,K,V) = {\rm softmax}(QK^T)V
\end{equation}
Since $QK^T$ is quadratic in sequence length and memory complexity, this severely limits applying attention to long sequence scenarios.
Therefore, several methods devoted to linearized attention computation have been proposed to address this dilemma. Based on the characteristics of these methods, we categorize them into associativity-based and low-rank-based methods.

\textbf{1) Associativity Based Methods}

The natural idea is that if we can calculate $K^TV$ first utilizing the associativity of matrix multiplication, we can achieve linear complexity in attention computation.
However, due to the presence of softmax, we cannot easily implement this.
For each row $i$ in the attention result, we can equivalently represent it as:
\begin{equation}
    {\rm Attention}(Q,K,V)_i = \frac{\sum_{j=1}^{n}{\rm sim}(q_i,k_j)v_j}{\sum_{j=1}^{n}{\rm sim}(q_i,k_j)}
\end{equation}
Where $sim(q_i, k_j) = e^{q_ik_j^T}$, it is actually a weighted average of $v_j$ with weights given by $e^{q_ik_j^T}$.
A natural thought is that if we can find two functions $\phi_1(x)$ and $\phi_2(x)$ such that: 
\begin{equation}
    {\rm sim}(q_i, k_j) = \phi_1(q_i)\phi_2(k_j)^T
\end{equation}
and satisfy $sim(q_i, k_j)>=0$ all the time, and also satisfy:
\begin{equation}
    \left(\phi_1(q_i)\phi_2(k_j)^T\right)v_j = \phi_1(q_i)\left(\phi_2(k_j)^Tv_j\right)
\end{equation}
Then, we can achieve linear attention. Building on this idea, many different approaches to linear attention have been proposed\cite{TransformersAreRnns, Performer, Nystrmformer, TransformerQuality, hyperAttention, EfficientAttention}.

Specifically, \cite{TransformersAreRnns} achieves this by constructing:
\begin{equation}
    \phi_1(x) = \phi_2(x) = elu(x) + 1 = 
    \left\{ 
        \begin{array}{lc}
            1+x & x \geqslant 0 \\
            e^x & x<0\\
        \end{array}
    \right.
\end{equation}
Performer\cite{Performer} also achieves linear attention through a kernel function method. It proposes a FAVOR+ method that cleverly uses random projection to project the input features orthogonally. Without relying on any prior
and without loss of accuracy, it successfully realizes the linear attention.
Specifically, by taking $\phi$ of the following form for functions $f_1,...,f_l:\mathbb{R}\to\mathbb{R}$, function $g:\mathbb{R}^d\to\mathbb{R}$ and deterministic vectors $\omega_i$ or $\omega_1,...,\omega_m\stackrel{\mathrm{iid}}{\sim}\mathcal{D}$ for some distribution $\mathcal{D}\in\mathcal{P}(\mathbb{R}^d)$:
\begin{equation}
    \phi(\mathbf{x})=\frac{h(\mathbf{x})}{\sqrt{m}}(f_1(\omega_1^\top\mathbf{x}),...,f_1(\omega_m^\top\mathbf{x}),...,f_l(\omega_1^\top\mathbf{x}),...,f_l(\omega_m^\top\mathbf{x}))
\end{equation}
To better describe $f_i$, $h$ and $\omega_i$ in $\phi$, for the element $A(i,j)=exp(q_ik_j^T) $ in the $i$th row and $j$th column of the original attention matrix $A$, we give it a generalized definition:
\begin{equation}
    \mathrm{SM}(\mathbf{x},\mathbf{y})\stackrel{\mathrm{def}}{=}\exp(\mathbf{x}^\top\mathbf{y})
\end{equation}
In fact, as early as \cite{RandomFeatures} there was an approximate expression for $\mathrm{SM}(\mathbf{x},\mathbf{y})$ with $h(x)=exp(\frac{||x||^2}2)$, $l=2$, $f_1=sin$, $f_2=cos$. Since the previous methods appear to have sin and cos trigonometric functions and instabilities such as negative numbers may appear in the computed results, Performer proposes another, more stable, approximation:
\begin{equation}
    \begin{aligned}
    &\mathrm{SM}(\mathbf{x},\mathbf{y})=\mathbb{E}_{\omega\sim\mathcal{N}\left(0,\mathbf{I}_d\right)}\left[\exp\left(\omega^\top\mathbf{x}-\frac{\|\mathbf{x}\|^2}2\right)\right. \\
    &\left.\exp\left(\omega^\top\mathbf{y}-\frac{\|\mathbf{y}\|^2}2\right)\right]=\Lambda\mathbb{E}_{\omega\sim\mathcal{N}\left(0,\mathbf{I}_d\right)}\cosh(\omega^\top\mathbf{z})
    \end{aligned}
\end{equation}
where $\Lambda=\exp(-\frac{\|\mathbf{x}\|^2+\|\mathbf{y}\|^2}2), \mathbf{x},\mathbf{y}\in\mathbb{R}^d,\mathbf{z}=\mathbf{x}+\mathbf{y}$ and cosh is hyperbolic cosine. This is equivalent to making:
\begin{equation}
    h(x)=exp(-\frac{||x|||^2}2),l=2,f_1(x)=exp(x),f_2(x)=exp(-x)
\end{equation}
However, to ensure accuracy, the number of random samples $m$ is usually larger than the feature dimension d, which means that when dealing with short sequences, the Performer may not perform as well as the standard Transformer. Only when the sequence is relatively long can its advantage be fully leveraged. Similarly, \cite{EfficientAttention} achieves linear approximate attention through a double softmax approach:
\begin{equation}
    {\rm Attention}(Q,K,V) \approx {\rm softmax_1}(Q){\rm softmax_2}(K)V
\end{equation}
Where, $softmax_1$, $softmax_2$ refer to softmax operations in the first$(N)$ and second$(d)$ dimension, respectively. However, directly softmaxing $Q, K^T$ separately, i.e., without similarity (inner product) computation, gives the impression of running counter to the attention mechanism. 
\cite{Nystrmformer} builds on this by first considering $Q, K$ as $n$ d-dimensional vectors, respectively, and then clustering them into matrices consisting of $m$ cluster centers $\tilde{\boldsymbol{Q}},\tilde{\boldsymbol{K}}\in\mathbb{R}^{m\times d }$. In addition, it inserts a matrix $M \in \mathbb{R} ^{m \times m} $ in the middle such that the final attention computation can be represented as 
\begin{equation}
\begin{aligned}
    {\rm Attemtion}(Q, K, V) \approx{softmax\left(Q\tilde{K}^\top\right)}\\\left({\rm softmax}\left(\tilde{Q }\tilde{K}^\top\right)\right)^{-1}{\rm softmax}\left(\tilde{Q}K^\top\right)V    
\end{aligned}
\end{equation}
which is closer to the standard attention.

Recently, HyperAttention\cite{hyperAttention} simplified the existing algorithm based on Kernel Density Estimation (KDE), identified the main entries in the attention matrix through Hamming-ordered locally sensitive hashing, and proposed a simple linear time attention approximation algorithm. This algorithm can achieve a wide range of linear approximation attentions while ensuring the spectral properties of attention and supporting causal masks. It is worth noting that the acceleration effect of HyperAttention can be tens of times different in two cases of using causal masks and not using causal masks. At the same time, if HyperAttention completely replaces all layers of attention, the model performance will be significantly reduced, so this method still needs to balance speed and performance.

\textbf{2) Low-rank Based Methods}

Other methods\cite{Linformer, Transformer-VQ} to achieve linear attention are through the utilization of low-rank property. Linformer\cite{Linformer} observed that the normalized cumulative singular values of the attention matrices in the Transformer model exhibit low-rank properties across multiple tasks. Based on this observation, Linformer preserves the original Scaled-Dot Attention formulation but projects $K$ and $V$ with two matrices $E, F \in \mathbb{R}^{m\times n}$ before computing attention, enabling linear approximate attention computation, formally expressed as:
$$Attention(Q,K,V) \approx softmax(Q(EK)^T)FV$$
In order to maintain its performance, it's essential to set the value of $m$ high enough. However, this can result in Linformer processing short sequences relatively slowly. Consequently, there might be a significant difference between the theoretical complexity and the practical usage of Linformer.

Recently, Transformer-VQ has adopted a unique perspective, performing Vector Quantization (VQ) on the key matrix $K$ in attention calculation. This is achieved by making each vector in $K$ closest to the vector in $C$, where $C$ is the training parameter and the VQ codebook. VQ can be mathematically represented as:
$\widehat{K} =  VQ(K, C), K\in R^{n \times d_k},\ C\in R^{c\times d_k}$
Since each vector in $\widehat{K}$ is from $C$, we can first calculate $QC^T$, which is linear due to the fixed size of $C$. Transformer-VQ cleverly constructs a $\triangle \in \{0, 1\}^{n \times c}$ such that:
\begin{equation}
    exp(QK^T)V = exp(QC^T\triangle^T)V = exp(QC^T)(\triangle^TV)
\end{equation}
The computational complexity of this calculation is $O(ncd_k + ncd_v + ncd_v) = O(n)$, achieving linear attention.
Moreover, this method can naturally be applied to autoregressive tasks, making it a promising approach.

\begin{figure}[!t]
    \centering
    \includegraphics[width=3.5in]{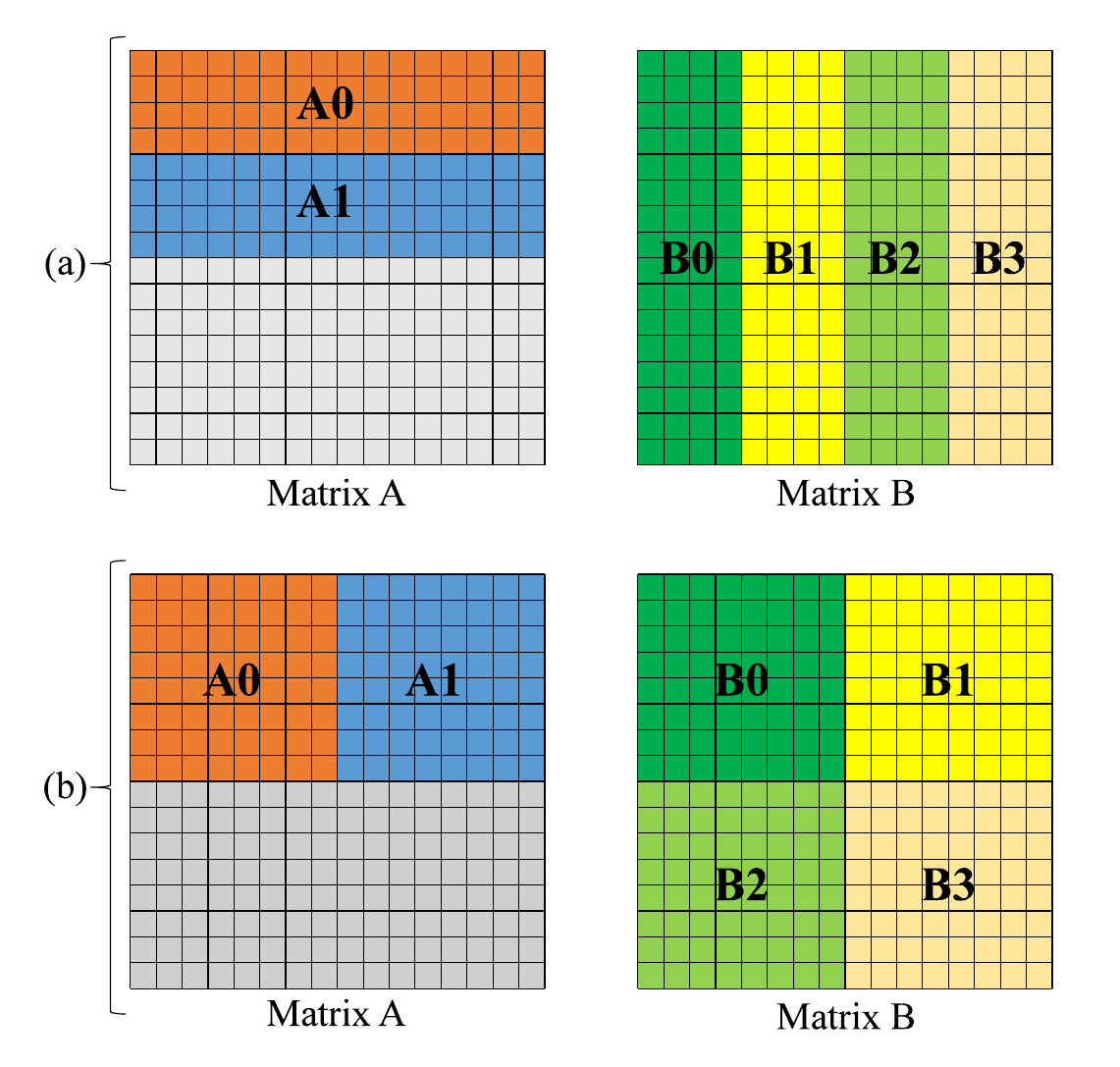}
    \caption{(a):Standard matrix multiplication, (b):Tiling matrix multiplication.}
    \label{Tiling}
\end{figure}

Unlike the previous approximate attention methods, FlashAttention\cite{Flashattention} focuses its improvements on reducing the memory access overhead with great success. It achieves acceleration of training and inference while reducing the memory footprint of attention. More importantly, it is not an approximate attention method, meaning its computational results are precisely equal to the standard attention results.

At its core is tiling, which is the chunked transfer of matrices involved in computation to shared memory to improve overall read and write speed.
For example, as shown in Fig.~\ref{Tiling} , suppose we now need to compute the upper half of the result of multiplying a matrix $A$ with a matrix $B$, $C_0$. For standard matrix multiplication there is $C_0 = (A_0B_0 + ... + A_0B_3) concat (A_1B_0 + ... + A_1B_3)$. And for Tiling matrix multiplication, $C_0 = (A_0B_0 + A_0B_1) concat (A_1B_2 + A_1B_3)$ reduces the memory access to half of the standard matrix multiplication.

In fact, in the Naive version of matrix multiplication, only one row and one column of the two matrices are read from memory each time for computation. The memory access efficiency is very low, and the two examples here read the same number of elements from the matrices A and B, respectively, just for comparability.
For matrix multiplication, we can use Tiling directly by chunking. Still, there are softmax operations in attention, and the denominator of softmax contains the summation term associated with all the elements, so the real difficulty in applying tilting to attention lies in the chunking of softmax.

For standard attention, we usually use the numerically safe softmax:
\begin{equation}
    Softmax(x_i)=\frac{e^{x_i-m}}{\sum_{j=1}^N e^{x_j-m}}
\end{equation}
where $m$ denotes the maximum of all $x_i$. To get the final result, we need three rounds of iterations:
\begin{itemize}
    \item Iterate over all $x_i$ to find the maximum value $m$
    \item Iterate through all $x_i$ to find the $sum = \sum_{j=1}^N e^{x_j-m}$
    \item Calculating each $Softmax(x_i)$
\end{itemize}
Since each round of iteration depends on the results of the previous iterations, the computation cannot be done independently in chunks.
One existing method is to define a single sequence $l^{'}:l_i^{'}=\sum_{\mathrm{j=1}}^\mathrm{i}\mathrm{e^{x_j-m_i}}$, thus having:
\begin{equation}
    \begin{aligned}l_i^{\prime}=&\sum_{j=1}^ie^{x_j-m_i}=\left(\sum_{j=1}^{i-1}e^{x_j-m_i}\right)e^{m_{i-1}-m_i}+e^{x_i-m_i}\\=&\ l_{i-1}^{\prime}e^{m_{i-1}-m_i}+e^{x_i-m_i}\end{aligned}
\end{equation}
It's a matter of cobbling together a $\sum_{j=1}^{i-1}e^{x_j-m_i}$ out and replacing it with an incremental computation of $l_{i-1}^{'}$, and it's clear that, after we get to this point, our sequences can be computed in the same round of iterations as $l^{'}$ and $m$, and in the end, $l_n^{'}$ will be equivalent to $l$, and in that way we'll be able to reduce the three rounds of iterations to two rounds of iterations.
However, the two-step iteration is still coupled and cannot be chunked for separate computations. Inspired by the previous derivations, FlashAttention derives methods to obtain the final O-matrix after one round of iterations. A row in matrix $O$ is a weighted summation of V and Softmax results, which can then be expressed as: 
\begin{equation}
    o_i\leftarrow\sum_{j=1}^N\left(\frac{e^{x_j-m_N}}{l_N}V[j,:]\right)
\end{equation}
Using and the same trick, introduce a sequence of $o^{'}$ alone and let it participate in the computation using the local $m_i$ and $l_{i}^{'}$: 
\begin{equation}
    o_i^{\prime}\leftarrow\sum_{j=1}^i\left(\frac{e^{x_j-m_i}}{l_i^{\prime}}V[j,:]\right)
\end{equation}
It is easy to see that for $N$, $o_i$ is equal to $o^{'}_N$, and the problem translates into figuring out how to cobble together a $\sum_{j=1}^{i-1}\left(\frac{e^{x_j-m_{i-1}}}{l_{i-1}^{\prime}}V[j,:]\right)$ out of the formula replacing it with $o^{'}_ {i-1}$:
\begin{equation}
    \begin{aligned}o'_i&=\sum_{j=1}^i\left(\frac{e^{x_j-m_i}}{l'_i}V[j,:]\right)\\&=\left(\sum_{j=1}^{i-1}\frac{e^{x_j-m_{i-1}}}{l'_{i-1}}V[j,:]\right)\frac{e^{m_{i-1}}}{e^{m_i}}\frac{l'_{i-1}}{l'_i}+\frac{e^{x_i-m_i}}{l'_i}V[i,:]\\&=o'_{i-1}\frac{l'_{i-1}e^{m_{i-1}-m_i}}{l'_i}+\frac{e^{x_i-m_i}}{l'_i}V[i,:]\end{aligned}
    \label{Flashattention-eq-o}
\end{equation}
Calculating Attention requires only one round of iterations with the above formula, so we can chunk the calculation to find the final result. 
FlashAttention-2\cite{Flashattention-2} improves on this by improving the formula \ref{Flashattention-eq-o} as follows:
\begin{equation}
    o_i^{\prime}=o_{i-1}^{\prime}l_{i-1}^{\prime}e^{m_{i-1}-m_i}+e^{x_i-m_i}V[i,:]
\end{equation}
Compared to the original $o_i^{\prime}$, we only need to divide $l_{N}^{\prime}$ by one more $l_{N}^{\prime}$ in the final computation $o_N^{\prime}$ to get the correct result, thus avoiding the intermediate multistep scaling division operation. 
It also reduces the memory write overhead. Specifically, in FlashAttention, it is fixed $K_j, V_j$ enumeration of $Q_i, O_i,l_i^{\prime},m_i$ for computation; in this way, for each computed $O_i$ we need to write it back to memory, which requires $O(N^2d^2M^{-1})$ write complexity, where $M$ denotes the size of the shared memory.
In FlashAttention-2, we fixed $Q_i, O_i,l_i^{\prime},m_i$ to enumerate $K_j, V_j$, so that the final result of $O_i$ can be computed at once and then written back to memory, and the complexity of writing is reduced to $O(Nd)$. In addition, it also parallelizes the dimension of sequence length; when the batch size and the number of heads are small, it increases the parallelism on the sequence length to improve the GPU occupancy, significantly improving the computation speed.

In general, efficient attention optimization methods mainly include sparse attention, linearized attention, and FlashAttention. However, there is often a gap between the practical and theoretical effects of many efficient attention methods, for example, many sparse attention methods are difficult to achieve the theoretical effects in practice due to the discontinuous memory accesses, which is mostly because we do not take into account the characteristics of the existing hardware when improving the methods.

\subsection{Neural Architecture Search.}
Although there have been significant advances in compression and acceleration methods for LLMs, many current hyperparameters that determine the final shape of the model still need to be determined by hand design. This hand design approach often requires a great deal of specialized knowledge and experience on the part of the designer, and it also has the problems of requiring long training time and high cost. In this dilemma, one promising solution is Neural Architecture Search (NAS)\cite{Hat, NAS_survey, ENAS_survey, NAS, Fbnetv2, Fbnetv, EvolvedTransformer, ME_F_TNAS, Darts}. For simplicity's sake, next, we will present a representative work from one of them.

The high computational cost of the Transformer model makes it difficult to deploy on some hardware devices and to realize the low-latency of inference on hardware devices with limited resources, HAT\cite{Hat} has emerged. The idea of HAT is to search for the best-performing model structure parameter that satisfies the requirement (given the hardware conditions and resources) for a given latency requirement. However, searching out the model structure and training and evaluating it from scratch is costly and slow. It avoids expensive retraining by constructing a Super Transformer such that it approximately contains all Sub Transformer models in the search space by sharing weights. Meanwhile, HAT trains a delay predictor to predict the delay through an offline method, which further speeds up the search. In addition, it observes several important properties: 
\begin{itemize}
    \item First, focusing on multiple encoding layers is beneficial for decoding layers. 
    \item Second, different hardware has different preferences for the model, with GPUs preferring shallow and wide Transformers and ARM CPUs preferring narrow and deep Transformers.
\end{itemize}
Overall, HAT provides an efficient NAS scheme for transformer models under different hardware conditions and latency requirements. At the same time, it can be well combined with other compression acceleration methods because it finds suitable model structure parameters for a given condition without changing the model's architecture.

%% file: dynamic_inference.tex
Scaling up the size of language models has been proven to be an effective approach for enhancing their performance on NLP tasks~\cite{MoE-power_law1, MoE-power_law2}.
However, the substantial computation costs and memory demands associated with scaling present a major challenge in the advancement of LLMs.
To address these issues while still harnessing the benefits of scaling,~\emph{dynamic neural networks} (DyNNs) engage only a subset of the network for processing each input, making the entire model more flexible and efficient in meeting computational demands under resource-constrained environments.
In the field of NLP and the domain of LLMs, current research on DyNNs primarily encompassess the following three methodologies:~\emph{early exit, cascade inference} and~\emph{mixture of experts (MoE)}.

\textbf{Early exit} is designed to dynamically terminate the inference process at the early layers of deep neural networks (DNNs), thereby reducing computational costs and improving response time~\cite{MoE-DNN_NLP_survey}.
The intuition is that the predictions for less complex words can often be accurately accomplished in earlier layers of the network~\cite{MoE-DoLa}.
These methods typically integrate a series of internal classifiers within the network, which provide signals for early exiting during inference.
Various exit criterions have been proposed~\cite{MoE-DeeBert, MoE-FastBert, MoE-RomeBert, MoE-SkipBert, MoE-LeeBert, MoE-PLM_EE, MoE-TR-BERT, MoE-learning_to_skip}.
This line of work mainly focuses on and is applied to small or medium-size language models, such as Bert.
And the accuracy may not be sufficient enough to support the application of general LLMs in more complex and realistic scenarios.

\textbf{Casacade inference} utilizes a series of language models of varying sizes to process requests with different levels of complexities. 
Tabi~\cite{MoE-Tabi} proposes an inference system with multi-level inference models and a probability-based dispatcher to determine the handling strategy for input queries and balance both accuracy and efficiency.
FrugalGPT~\cite{MoE-FrugalGPT} learns to adaptively triage quries from diverse datasets and tasks and direct them to an appropriate combination of LLM APIs.
Both EcoAssistant~\cite{MoE-EcoAssistant} and~\cite{MoE-optimal_caching} employ a query cache to reference historical data for faster responses and a hierarchy of LLMs to handle those mismatched new queries.
Mixture-of-Thoughts~\cite{MoE-Mixture_of_thoughts} considers the consistency of answers from weaker LLMs as an indicator of the question difficulty to decide whether to leverage stronger LLMs.
Generally, this line of works has emerged recently and demonstrates a promising direction for the development of more efficient LLM systems.

Compared to the two types of methods above, the study of~\emph{MoE} has an extensive history spanning multiple machine learning fields including NLP.
MoE horizontally extends a feed-forward network (FFN) with multiple sub-networks, of which only one or few will be activated during a single forward pass.
It is widely incorporated into the architectures of today's LLMs~\cite{MoE-GPT4, MoE-Mixtral} to provide both efficient and powerful services.
So in the remainder of this section, we will delve into the realm of MoE.
Section~\ref{sec:MoE-subsec1} begins with an introduction to the basic concepts of MoE, followed by an extensive survey of contemporary research on incorporating MoE into LLMs, which includes algorithmic and architectural design, training strategies and pratical applications.
Section~\ref{sec:MoE-subsec2} offers a concise review of some representative studies on integration of MoE with previously dicussed model compression and acceleration techniques, highlighting its potential in the development of more comprehensive and cost-efficient LLM systems.

\subsection{Mixture of Experts}
\label{sec:MoE-subsec1}

\begin{figure}[!t]
    \centering
    \includegraphics[width=\linewidth]{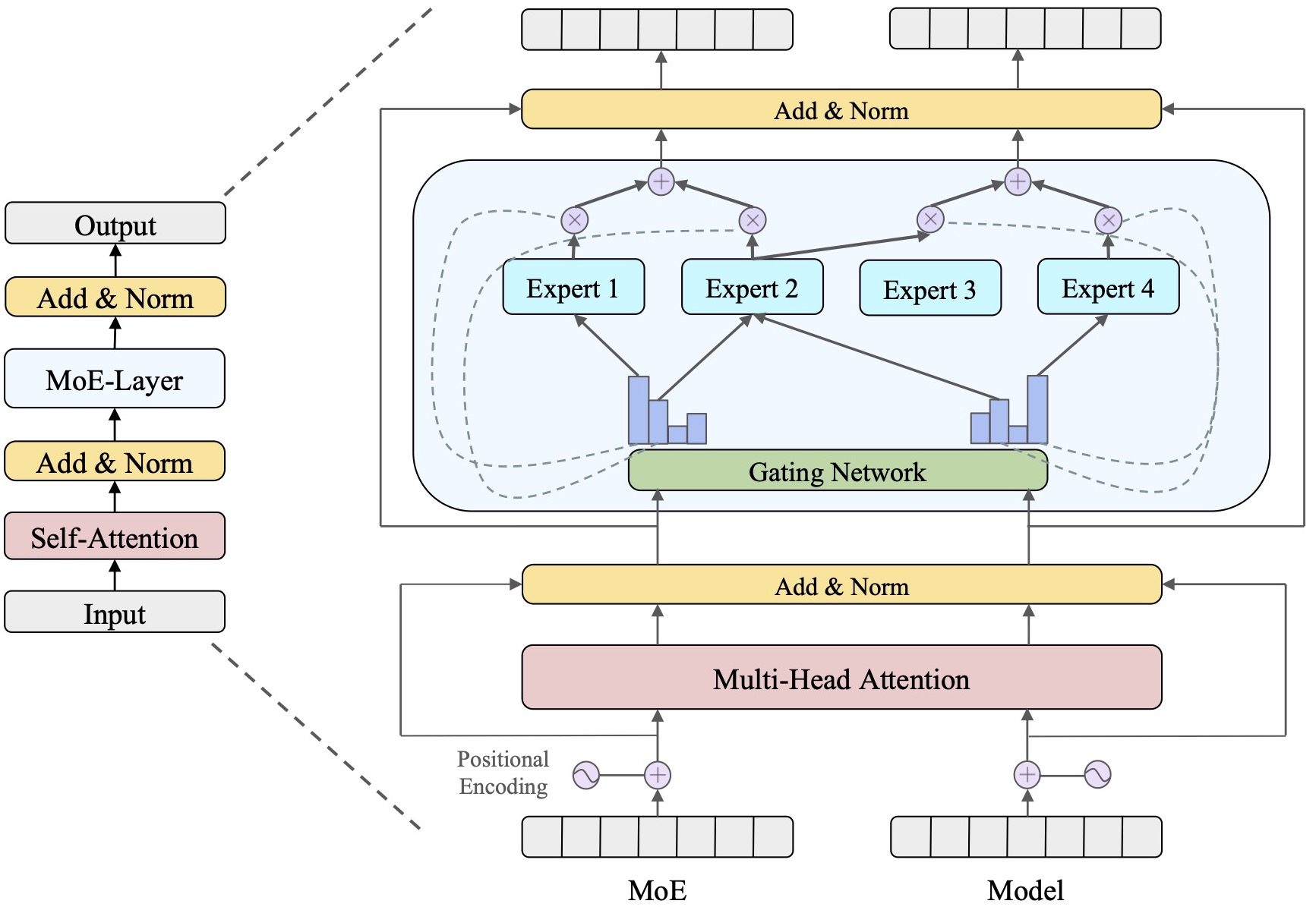}
    \caption{Illustration of a transformer block with an integrated MoE layer.}
    \label{MoE}
\end{figure}

\begin{table*}[!t]
\renewcommand{\arraystretch}{1.3}
\caption{A summary of various Mixture-of-Experts (MoE) methods. 
For models of the largest size, we present the total number of paremeters along with the number of experts per MoE layer. 
For methods utilizing shared experts~\cite{MoE-DeepSpeedMoE, MoE-DeepSeekMoE}, we include both (the number of experts used for sharing + the number of experts used for routing). }
\label{MoE_table}
\centering
{\footnotesize
    \begin{tabularx}{\textwidth}{| >{\setlength{\hsize}{.35\hsize}\centering}X
        | >{\setlength{\hsize}{.25\hsize}\centering}X
        | >{\setlength{\hsize}{.25\hsize}\centering}X
        | >{\setlength{\hsize}{.4\hsize}\centering}X
        | >{\setlength{\hsize}{.75\hsize}}X
        |}
    \hline
    \textbf{Methods} & \textbf{Base Model} & \textbf{Sparsity} & \textbf{Largest Model Size (Params / Num. experts)} & \textbf{Load Balance}   \\
    \hline
    Sparsely-Gated \cite{MoE-sparse_gating} & LSTM  & top-$k$  & 137B / 131072 & Noisy top-k gating and auxiliary loss term.  \\
    \hline
    GShard \cite{MoE-GShard} & NMT & top-2 &  600B / 2048 & Local group dispatching and auxiliary loss term. \\
    \hline
    Switch \cite{MoE-Switch} & T5 & top-1 &  1571B / 2048 & Auxiliary loss term. \\
    \hline
    Expert Choice \cite{MoE-expert_routing} & Transformer & Expert Choice & 143B / 64 & Expert choice routing. \\
    \hline
    DeepSpeed-MoE \cite{MoE-DeepSpeedMoE} & GPT-3 & Residual-MoE & 52B / (1+127) & Multi-expert and multi-data parallelism. \\
    \hline
    M6-T \cite{MoE-M6-t} & M6 & $k$ top-1 & 10003B / 960 & -  \\
    \hline
    Brainformer \cite{MoE-Brainformer} & Non-uniform & Expert Choice & 158B / 64 & Expert choice routing. \\
    \hline
    Mixtral 8x7B~\cite{MoE-Mixtral} & Mistral 7B & top-2 & 47B / 8 & - \\
    \hline
    DeepSeekMoE~\cite{MoE-DeepSeekMoE} & DeepSeek & Shared Experts & 145B / (4+128) & Auxiliary loss terms. \\
    \hline
    \end{tabularx}
}
\end{table*}

The earliest concept of MoE dates back to three decades ago \cite{MoE-earliest_work1, MoE-earliest_work2} but firstly demonstrates its effectiveness in massively improving model capacity without paying a proportional computation overhead using sparse gating \cite{MoE-sparse_gating}. In sparse MoE models, a subset of model parameters are partitioned into a set of $N$ expert networks $\{E_i(\cdot)\}_{i=1}^N$, each operates independently on the input with unshared weight. During training and inference, each input example $x$ (i.e., a token representation in language models) would be routed to specific expert(s) via gating function $G(\cdot)$ whose input is also $x$ and the output is a sparse $n$-dimensional vector. The final output $y$ of MoE module is a weighted combination which can be written as:
\begin{equation}
    y = \sum_{i=1}^N G(x)_i E_i(x)
\end{equation}
Based on the sparsity of $G(x)$, we can skip the computation of $E_i(x)$ wherever $G(x)_i=0$ . Since only part of model parameters are activated for each input and all experts have the potential to be utilized by different samples, MoE model theoretically enjoys a competitive learning ability with a much faster inference speed comparing to its dense counterpart. As Transformer\cite{MoE-Transformer} has become the standard choice for language model, the most common way to introduce MoE into Transformer is to replace the feed-forward layer of certain block(s) with MoE layer, of which each expert is itself a regular Transformer feed-forward network. An example of MoE layer is given in Fig.~\ref{MoE}. By increasing the number of experts, the parameter size could grow from hundreds of millions to hundreds of billions or even trillions to match the size of LLMs. There are mainly three key elements to characterize one MoE method: 
(1) \textbf{Routing method} decides where and how each input will be routed in the MoE layer, which is the most critical element of MoE algorithms.
(2) \textbf{MoE model architecture} discusses common or specific design choices towards building scaled models that are more performance-effective and parameter-efficient.
(3) Special \textbf{training strategies} sometimes are needed to accommodate the uncertainty raised from learning-based routing methods. We summarized some representative MoE methods in TABLE~\ref{MoE_table}.

\subsubsection{Routing Method}

For an MoE system, the most crucial factor affecting its performance and the primary design concern is the ensurance of \textbf{load balancing} among the experts.

In a standard distributed training settings, experts from the same layer are scattered across multiple devices. 
Each of them is a simple feed-forward network (FFN) in most cases and computes in parallel.
And the maximum number of tokens each expert can process during a single forward pass, also known as the \textit{expert capacity}, is limited by the memory of device it resides.
Generally, without specific algorithm or architecture design, tokens from the same batch are assigned unevenly among experts by the gating function due to its sparsity. 
Therefore, if too many tokens are routed to the same expert, surpassing its capacity, this will lead to an \textit{overflow} issue.
The computation for the overflown part will be skipped and those tokens will be passed directly to the next layer via a residual connection.
Thus unbalance loads across experts lead to under-processing of tokens and a waste of computation and memory for experts with empty slots.


Besides performance decline, imbalanced loads across experts could also lead to a \textit{self-reinforcing} phenomenon that inherently limits the capabilities of MoE system through training.
This phenomenon manifests as the gating network converging to a state where it always produces large weights for the same few experts, which rapidly trained with large amount of data and are favored by the gating network even more.
Consequently, the remaining experts remain undertrained and underutilized, 
which results in the original MoE network collapses to a smaller network comprising only the few active experts.

Alongside the problem of load imbalance, there are also methods dedicated to mitigate other adverse effects stemming from the sparse nature of MoE, such as unstable routing or the trade-off between sparsity and accuracy.

Based on the primary problem each method address, we categorize the existing routing methods into the following two groups and begin our review.

\textbf{Towards load balancing.} 
Most routing methods apply an additional learnable gating network in each MoE layer. A simple choice yet adopted by most works is to use one linear layer with trainable matrix $W \in \mathbb{R}^{d\times N}$, where $d$ is model dimension and $N$ is the number of experts, then followed by a non-linear function like softmax or sigmoid. The $N$ columns of $W$ $\{w_1, \cdots, w_N\}$ can also be seen as embeddings of $N$ experts respectively and readers may find this expression in some works. For each token $x \in \mathbb{R}^d$, the routing score between $x$ and $i$-th expert is given by the dot-product similarity metric $s_i=x\cdot w_i$. To add sparsity to the gating network, i.e., to use only $k$ experts ($k \ll N$) each MoE layer, only experts with the highest top-$k$ values of $\{s_i\}_{i=1}^N$ (set of indices $\mathcal{T}$) will be selected for token $x$. In general, if expert $i$ is activated, its gating value is given by
\begin{equation}
\label{gating-value}
G(x)_i = \begin{cases}
\exp(s_i) / \sum_{j \in \mathcal{T}} \exp(s_j), & \text{softmax gating, $k>1$} \\
\exp(s_i) / \sum_{j=1}^N \exp (s_j), & \text{softmax gating, $k=1$} \\
\sigma (s_i), & \text{sigmoid gating}
\end{cases}
\end{equation}
where $\sigma(\cdot)$ is the sigmoid function. Note that for softmax gating, top-$1$ methods formulate sightly differently to make $G(x)_i$ non-trivial. 
The above idea is first proposed in~\cite{MoE-sparse_gating} and applied to LSTM models~\cite{MoE-LSTM}.
To mitigate the issue of self-reinforcing, they add a tunable Guassian noise to the routing scores and employ two additional loss terms to encourage more balanced routing scores and selection rates across the experts.
GShard~\cite{MoE-GShard} integrates MoE into transformers by replacing every other feed-forward layer with an MoE layer using top-2 gating.
They propose several strategies to ensure load balancing, including (1) partioning tokens into groups and limiting the number of tokens that each experts can receive from a single group; (2) an auxiliary loss term which has been widely adopted by later works~\cite{MoE-Switch, MoE-representation_collapse, MoE-MoEC} and (3) a random routing strategy.
To further align the computational cost with that of a vanilla transformer, Switch Transformer~\cite{MoE-Switch} routes each token to only one expert with top-1 gating.

The aforementioned studies primarily suggest intermediary strategies, such as auxiliary loss terms, to promote load balancing during training.
Other researches aim at improving the gating function to directly regularize or even guarantee perfectly balanced loads across experts.
BASE Layer \cite{MoE-BASE} formulates token-to-expert allocation as a linear assignment problem, which maximizes the scores between expert and their assigned tokens while subject to a constraint that each expert must receive an equal amount of tokens.
Expert choice \cite{MoE-expert_routing} is another routing strategy guaranteeing perfect load balancing with no additional regularization required. Instead of letting tokens select the top-$k$ experts, each expert chooses top-$k$ tokens which also based on routing scores. 
Consequently, each expert can have exactly same workloads and each token can be processed by a variable number of experts.
Clark et al. \cite{MoE-unified} propose a routing algorithm using reinforcement learning, in which each router is seen as policy with actions and rewards defined by the selection of experts and the predicted probability of the correct output token respectively.
In addition to employing learnable gating networks, some studies implement non-learnable strategies to regulate the load distribution across experts.
Hash layer~\cite{MoE-Hash} employs a parameter-free hash function to replace dynamic gating with a pre-defined fixed mapping from tokens to specific experts, which consequently eliminates load imbalance.
MoWE~\cite{MoE-MoWE} routes each word to one specific expert based on auxiliary vocabulary mapping and ensures that the words assigned to each expert are of approximately the same frequency in the pretraining data.
Inspired by Hash layer, PanGu-$\Sigma$~\cite{MoE-PanGu-sigma, MoE-LocMoE} deploys a two-level routing that first maps each token to a group of candidate experts by domain and then uses random hash to choose a particular expert from that group for processing.
THOR \cite{MoE-THOR} also simplifies the gating process with a parameter-free approach by randomly selecting a pair of experts from each layer during a training iteration and dispatching the whole batch of tokens to those experts. 

The methods we have described so far can only be utilized to the fullest extent with sufficient device resources (\ie, no overflow issues are encountered).
In more realistic senarios, one must take additional strategy to handle overflow.
The simplest solution is to stop assigning tokens to experts with full workloads. 
However, in this way only either the prefix of the input sentences or the sentences with small indices in the batch dimension will be processed, based on whether the batch is flattened along the sequence length or the batch dimension, which leads to a biased selection and underutilization of training or inference data. 
To address this issue, Z-code \cite{MoE-Zcode} and BASE \cite{MoE-BASE} propose to shuffle the input tokens in the flattened batch to disentangle the probability of a token being selected from its position. In the domain of vision transformer, V-MoE \cite{MoE-VMoE} introduces Batch Prioritized Routing algorithm (BPR), which additionally compute a priority score for each token (e.g., the maximum of routing scores between the token and all experts) and sort tokens accordingly before allocation. Therefore only insignificant tokens will be discarded. ST-MoE \cite{MoE-ST_MoE} finds that BPR benefits language models as well, especially in low-resource regimes where the expert capacity is even less than the average number of tokens each expert receives. 
Note that BPR can only be applied to encoder side of encoder-decoder model since the inputs of encoder are not autoregressive thus are allowed to see each other, otherwise model could cheat by using future token information.

\textbf{Effective routing with sparsity.}
Although sparsity bounds the computational cost in a large-scale MoE system, it generally limits the network's capability and impedes convergence due to unstable routing dynamics.
DeepSpeed-MoE \cite{MoE-DeepSpeedMoE} observes that increasing the number of experts each token goes through helps accuracy. 
In order to leverage this property while keeping the computation costs as top-1 gating, they propose Residual-MoE by fixing one expert and varying the second expert for each token to achieve the benefit of using 2 expert.
DeepSeek-MoE~\cite{MoE-DeepSeekMoE} employs the same methodology, utilizing a fixed set of shared experts to capture and consolidate common knowledge, augmented by a distinct set of routed experts dedicated to knowledge specialization.
M6-T \cite{MoE-M6-t} also notices the advantage of top-$k$ gating over the top-$1$. They propose to split expert into $k$ groups and perform $k$ top-$1$ routing in parallel to match the efficiency of top-$1$ gating.
To ensure each expert can receive rich and diverse tokens under the sparse settings, MoEC \cite{MoE-MoEC} encourages experts to form clustered structure by closing the routing probability among neighbor experts with designed loss term and randomly drops some experts in each cluster before the global top-1 gating.

Besides sparsity, 
DSelect-k \cite{MoE-DSelect_k} notices the discontinuity in top-k gating methods and suggest this could lead to convergence and performance issues when training with gradient-based methods. They propose a continuously differentiable and sparse gating function, which densely selects experts at beginning of training but fast converges to sparse expert selection by adding an regularization term.
X-MoE~\cite{MoE-representation_collapse} points out that current routing mechanisms tend to push token representations towards expert embeddings which potentially harms the representation capacity. 
They propose to calculate the routing scores between tokens and experts in a low-dimensional space with additional normalization and learnable gating temperature. 
ST-MoE~\cite{MoE-ST_MoE} conducted an extensive study on the training stability and fine-tuning quality of sparse MoE models.
They framed the training process of sparse models as a quality-stability trade-offs: various stability techniques, such as dropout and gradient clipping, could enhance the training stability but often come at the expense of model quality.
Therefore they propose router z-loss to address both the instability issue and the problem of quality degradation.

\subsubsection{MoE model architecture}

In this part we discuss how to arrange MoE layers into a Transformer-based model, such as the frequency of expert layers and the number of experts, which could significantly affect the scale of models and the overall performances.

In transformer, sparse model usually starts with a dense model and scales up by substituting or inserting MoE layers at a fixed interval or heuristically. 
A common design deployed in most large sparse expert models \cite{MoE-GLaM, MoE-GShard, MoE-expert_routing, MoE-Switch, MoE-efficient_LSLM_MoE} is to replace the feed-forward component of every other Transformer block with a MoE layer (i.e., at a frequency of 0.5). 
Other frequencies are also adopted, such as 0.25 (i.e., substituting every fourth FFN layer) in \cite{MoE-ST_MoE} and 1.0 (i.e., placing in every layer).
In general, experiments~\cite{MoE-unified} suggest a frequency at 0.5-1.0 and lower frequency undermines the performance. 
However, there are also works \cite{MoE-BASE, MoE-Hash, MoE-representation_collapse, MoE-MoEC, MoE-StableMoE} introducing a fixed number of expert layers to baseline models by spreading MoE layers unevenly across the network.
For instance, BASE\cite{MoE-BASE} inserts a large MoE layer consisting of stacked FFNs only after middle layer.  

As for the number of experts per-layer, although using more experts continuously brings quality gains in most cases,  diminishing returns in the improvements are also reported in earlier works\cite{MoE-GShard, MoE-Switch}. 
Further analyses \cite{MoE-unified, MoE-ST_MoE} also points out the drastically-diminishing incremental benefits from routing as the base model size increases. 
A fixed number of experts per-layer in $\{64, 128\}$ is recommended by \cite{MoE-unified} and also have been practiced in a lot of sparse large language models \cite{MoE-GLaM, MoE-ST_MoE, MoE-no_language_left_behind}. 
Moreover, DeepSpeed-MoE\cite{MoE-DeepSpeedMoE} questions the standard MoE architecture putting the same number of experts in all MoE layers. 
Their experiments suggest that a large number of experts in deeper layers boost the performance more effectively. 
Therefore they introduce a pyramid structure by utilizing more experts only in the last two layers and achieve comparable results as standard MoE models but with fewer parameters.

Above works are all built upon uniform transformer blocks and by interleaving dense and sparse layers, 
Brainformer \cite{MoE-Brainformer} explores a non-uniform architecture with sparse layer inspired by the success of EfficientNet \cite{MoE-efficientnet} and sandwich transformer \cite{MoE-sandwich_transformer}. 
An evolutionary search algorithm is applied to explore the best Brainformer block architecture in the search space consisting of different layer types (namely self attention, MoE and dense FFN sub-layers), interleaving orders and hyperparameter settings such as model dimension and number of attention heads. 
The whole network is constructed by stacking a variable number of blocks according to different scales. 
Experiment results demonstrate a clear advantage in terms of both efficiency and capacity over its GLaM\cite{MoE-GLaM} counterpart and Primer\cite{MoE-Primer} dense model produced by NAS\cite{MoE-NAS}.

\subsubsection{Training Strategies}

Existing learning-based routing methods usually train both the gating and expert networks jointly from scratch. As the parameters of MoE layers are randomly initialized, the routing behavior at the beginning stage of training can be seen as random routing and the correspondences between tokens and experts are highly \emph{unstable}. As a result, MoE models take a longer time to converge with a potential risk of reinforcing improper routing behavior which eventually limits the model quality. 

To handle this problem, a two-stage training strategy is introduced in \cite{MoE-StableMoE, MoE-EvoMoE} to separate the training of the gating network and expert networks. 
In the first stage, StableMoE~\cite{MoE-StableMoE} learns a balanced and cohesive routing strategy following a standard MoE training process with additional balance loss.
Throughout the second stage, the routing strategy is freezed to provide a stable token-to-expert assignment for the training of the rest of model. Experiments confirm that the consistency of routing strategy boost both the convergence speed and final performance.
Conversely, EvoMoE \cite{MoE-EvoMoE} starts from training then diversifying from one common expert at stage one before learning the gating network and sparsifying the network in the second stage. 
In this way experts are able to get sufficient training in the early stage and more suitable routing strategy can be built on the exploitation of specialized experts.

Another line of works set out to alleviate the \emph{overfitting} problem raised from the imbalance between vast number of parameters and limited training examples via special dropout\cite{MoE-dropout} mechanism at MoE layer. 
For instance, Switch Transformer\cite{MoE-Switch} increases the dropout rate solely inside the experts, named as \textit{expert dropout}, to aid the performance on downstream tasks with very few training data. 
Gating Dropout \cite{MoE-GatingDropout} further pushes traditional dropout to another level to reduce the communication cost for dispatching tokens across devices and also improve the performance during training. Specifically, they permit tokens to ignore the assignment from gating network with certain probability and instead route them to the experts on the same device. This also encourages experts to function more robustly and learn a generalization ability.
The results demonstrate that Gating Dropout indeed accelerate the convergence of sparse MoE models in terms of wallclock time and enhance the model quality.

\subsubsection{MoE Applications}
\label{applications and analysis}

The success of MoE models promotes a series of works deploying sparse MoE algorithms in actual LLM applications or combining with other model compression and acceleration techniques in pursuit of greater efficiency. 
CPM-2\cite{MoE-CPM2} integrates BASE Layer \cite{MoE-BASE} into their largest Chinese-English bilingual models with 198 billion parameters. 
Following GShard~\cite{MoE-GShard}, GLaM \cite{MoE-GLaM} trains a family of decoder-only sparse MoE models, the largest of which has 1.2T parameters and yields better zero, one and few-shot learning abilities in comparison with its dense GPT-3\cite{MoE-GPT3} counterparts.
Seeking to close the performance gap between high and low-resource languages and break the 200 language barrier, a Sparsely-Gated MoE model with 54.5B parameters is developed by \cite{MoE-no_language_left_behind}, following the optimaztion process in~\cite{MoE-GShard}, and casts light on a promising approach towards a universal translation system.
A revealing article on the technical details of GPT-4 \cite{MoE-GPT4} confirms the deployment of a MoE model consisting of 16 experts inside GPT-4. Each expert is tuned to specialize in a specific domain or task, thereby endowing GPT-4 with the multi-task ability.
Mixtral~\cite{MoE-Mixtral} builds upon Mistral 7B and replaces all FFN sub-blocks by MoE layers, each consisting of 8 experts and a top-2 gating network.
The resulted Mixtral 8x7B only uses 13B active parameters during inference but surpasses Llama 2 70B~\cite{MoE-Llama2} and GPT-3.5 on several benchmarks.
DeepSeekMoE~\cite{MoE-DeepSeekMoE} proposes a series of MoE models with sizes of 2B, 16B, and 145B as well as aligned versions, to demonstrate the adaptability and versatility of their MoE architectures.
OpenMoE~\cite{MoE-OpenMoE} also releases a suite of open-sourced MoE models, building upon the architectures of ST-MoE~\cite{MoE-ST_MoE} and Residual-MoE~\cite{MoE-DeepSpeedMoE}. In addition, they offer a comprehensive study and a few insights on MoE's routing behavior throughout training.

\subsection{Combineing MoE with other efficient techniques}
\label{sec:MoE-subsec2}

The Mixture-of-Experts approach inspires the field as an alternative pathway for buidling more powerful and efficient LLMs.
Given that MoE is akin to an art of architecture design and is orthogonal to most model compression and acceleration techniques, there are also works exploring ways to merge its inherent sparsity with other optimization strategies, such as pruning, distillation, and PEFT.
In this section, we will examine the most representative studies in this area and highlight the potential it holds for future research. 

\subsubsection{Model Compression}

In the realm of sparse MoE models, most existing works can be viewed as trading memory consumption for model quality. 
To reduce the memory footprint while retaining most of their capabilities, researchers have explored several ways to introduce traditional model compression techniques into MoE models.

Switch Transformer\cite{MoE-Switch} made the first attempt to distill large sparse models into small dense models. Their findings indicate that it is possible to preserve approximately 30\% of the quality gains achieved through scaling when distilling to a FLOP-matched dense variant for both pre-training and fine-tuning tasks. 
DeepSpeed-MoE \cite{MoE-DeepSpeedMoE} studies the potential of distilling a large teacher MoE model into a smaller student MoE model with shallower expert networks. Additionally, they suggest a stage-KD training strategy (\ie, halting distillation at specific steps) to mitigate the under-fitting issue stemming from the student model's limited expert capacity.

As another prominent tool for parameter reduction, the pruning of MoE models aims to remove redundant or less influential components, usually a subset of less important expert networks, with minimal impact on the performance. The hypothesis behind pruning, as suggested by Z-code\cite{MoE-Zcode}, is that different experts can specialize in distinct aspects of the task during training, making a subset of experts competent enough for a given task to a certain extent. Z-code tried two methods for expert selection: random selection and selection based on utilization rates in the validation set. 
Chen \etc \cite{MoE-task-specific} observe the long-tailed distribution of expert contributions in downstream tasks. Different from Z-code's approach, they propose to progressively prune away most experts throughout the fine-tuning process, leaving only the most professional one for the target downstream task. The experiment results highlight the effectiveness of their pruning strategy, preserving 99.3\% of the benefits from MoE while enjoying the same resource consumption as vanilla dense models during inference.
As an alternative approach, MPOE \cite{MoE-MPOE} introduce a parameter-efficient MoE architecture by decomposing the parameter matrix of each expert into \textit{central} and \textit{auxiliary} tensors. The \textit{central} tensor is believed to encode the majority of the information present in the original matrix, which is likely to be similar across experts and thus suitable for sharing among them. On the other hand, the \textit{auxiliary} tensors capture the individual characteristics and serve as a complement to the central tensor. This parameter-sharing method has been shown to be effective, achieving a 27.2x reduction in total parameters while yielding better performance comparing to the Switch Transformer.

Witnessing the great success of MoE models, there are also efforts to introduce sparsity into a standard transformer model with the purpose of reducing the number of parameters involved in computation while retaining the representation power.
MoEBERT \cite{MoE-MoEBert} adapts the feed-forward networks in a pre-trained BERT\cite{MoE-Bert} into multiple experts and activates only one expert during inference to increase the speed. To preserve the original representation power, they share the most important neurons in the FFNs among the experts based on the importance score~\cite{MoE-importance_score} when initializing MoEBERT. The training of MoEBERT incorporates layer-wise distillation, leading to a resulting model that outperforms other task-specific distilling methods.
MoEfication \cite{MoE-Moefication} aims to generalize the conversion from FFNs to MoE layers for various Transformer models. The idea is driven by the insight that only a tiny fraction of neurons of FFNs will respond to most inputs. To split the feed-forward layer into experts, neurons that often activates simultaneously are grouped into the same expert network. And the routing strategy is learned by approximating the calculation of the original model. 
To further reduce the computational and memory demards of standard transformers, $\sigma$-MoE~\cite{MoE-sigma-MoE} and SwitchHead~\cite{MoE-SwitchHead} introduce additional sparsity to the FFN and attention components, drawing on the principles of the MoE methodology.

\subsubsection{Efficient Finetuning}
In search of more efficient and powerful model architectures, researchers are also exploring the combination of MoE methods and other cost-effective techniques such as Mixer\cite{MoE-Mixer} and PEFT methods. These collaborative approaches primarily leverage the expressiveness provided by MoE while aggressively reducing the training and computation cost. 
Sparse Mixers \cite{MoE-sparse_mixer} replaces most of the self-attention sublayers with mixing and FFNs with MoE sublayers. SMLP \cite{MoE-sparse_all_mlp} goes one step further by substituting the self-attention modules with linear transformations, which also employs a MoE mechanism with routing in the feature dimension to ensure tokens from the same sentences are delivered to the same expert. AdaMix \cite{MoE-AdaMix} proposes a mixture of adapters \cite{MoE-Adapter} or a mixture of low-rank decomposition matrices \cite{MoE-LoRA} with stochastic routing mechanism \cite{MoE-THOR} as a novel fune-tuning technique to enhance the downstream performance. The result illustrates that AdaMix surpasses SOTA parameter-efficient fine-tuning and even full model fine-tuning algorithms on both NLU and NLG tasks.
Based on a similar idea, MixDA~\cite{MoE-MixDA} also utilizes a set of domain adapters to inject domain-specific knowledge in parallel and then train a mixture-of-adapters gate to dynamically fuse the knowledge from different domain adapters. This plug-in approach showcases its scalibility and efficiency on several domain tasks.
The same methodology is also adopted by~\cite{MoE-MoLE, MoE-SiRA, MoE-LoRAMoE, MoE-SPT} to achieve efficient finetuning on domain-specific or instruction datasets and to mitigate the catastrophic forgetting arising from continual learning.

%% file: AccelerationFramework.tex
\begin{table*}[!t]
\renewcommand{\arraystretch}{1.3}
\caption{
A summary of various acceleration frameworks.
}
\label{Framework_table}
\centering
{\footnotesize
    \begin{tabularx}{\textwidth}{
    | 
    >{\setlength{\hsize}{.35\hsize}\centering\arraybackslash}X
        | >{\setlength{\hsize}{.2\hsize}\centering\arraybackslash}X
        | >{\setlength{\hsize}{.45\hsize}\centering\arraybackslash}X
        |}
    \hline
    \textbf{Framework/Passage} & \textbf{Generalization} & \textbf{Method}  \\
    \hline
    DNNFusion\cite{DNNFusion} & General & Operator fusion\\
    \hline
    DeepSpeed Inference\cite{DeepSpeed-inference} & General & Operator fusion, tensor parallelism, inference pipeline\\
    \hline
    TurboTransformer\cite{Turbotransformer} & Specialized & Operator fusion, scheduling optimization\\
    \hline
    ByteTransformer\cite{ByteTransformer} & Specialized & Operator fusion, scheduling optimization\\
    \hline
    FlexGen \cite{Flexgen} & Specialized & Offloading system\\
    \hline
    Power-Infer\cite{PowerInfer} & Specialized & Offloading system\\
    \hline
    \end{tabularx}
}
\end{table*}

With the rapid development of Transformer-based models, various models have emerged. Because of different application scenarios, they have additional requirements in terms of latency, throughput, memory, \etc, making it difficult for us to deploy the models. In this section, we introduce some recently developed inference acceleration frameworks for LLM, which effectively improve the efficiency of the models in different scenarios, as shown in TABLE~\ref{Framework_table}. We classify the general framework and specialized framework based on the generalization.
Here are some more acceleration frameworks\cite{Alpa, Megatron-lm, Colossal-ai, Fairscale, Pax, Deepspeed, composer} specific to training, and since this paper focuses on inference, we will not discuss them specifically. If you want to deploy trained models to get efficient inference quickly, you can refer to these frameworks\cite{OpenLLM, RayLLM, MLC-LLM, Saxml, MOSEC, LLMFoundry}.

\subsection{General Framework}
In this section, we will introduce some relatively generalized frameworks\cite{DNNFusion, DeepSpeed-inference} proposed recently. No matter what kind of scenarios the models are deployed in, we can consider using them or combining their ideas to accelerate the inference and thus obtain higher efficiency. Since most big models are still deployed and run on GPUs, our generalization here refers to generalization under GPU hardware.

Operator fusion is a common method to accelerate model inference by eliminating unnecessary intermediate results, lowering memory requirements, and reducing unnecessary memory IO and kernel startup overhead, thus improving the utilization of computational resources such as GPUs, CPUs, and registers. Meanwhile, operator fusion is an essential optimization for many state-of-the-art DNN compilation frameworks, such as TensorFlow XLA\cite{TensorFlowXLA}, TVM\cite{TVM}, MNN\cite{MNN}, PyTorch JIT\cite{PytorchJIT}, and so on. However, these frameworks have stringent requirements for operator fusion, e.g., TVM uses relatively fixed schedule templates, resulting in missing many potential fusion opportunities. DNNFusion\cite{DNNFusion} has better coverage and fusion identification capabilities through algebraic simplification and rational classification of operators. In addition, it further improves the efficiency of operator fusion by eliminating unnecessary operators through heuristic methods.

Recently, Microsoft proposed DeepSpeed Inference\cite{DeepSpeed-inference}, an efficient integrated inference system for the increasingly diverse Transformer model, which reduces latency by a factor of 7.3 in state-of-the-art latency-oriented scenarios and increases throughput by more than 1.5 in throughput-oriented scenarios.
It includes the following two components:
\begin{itemize}
    \item \textbf{Multi-GPU Inference Solution:} It minimizes latency while maximizing throughput in dense and sparse Transformer models with GPU memory aggregation.
    \item \textbf{Heterogeneous Inference Solution:} Besides GPU memory and computation, it utilizes CPU and NVMe memory to achieve high inference throughput for large models that do not lend themselves to aggregated GPU memory.
\end{itemize}
Many strategies have been used to maximize the training throughput, such as tensor parallelism, pipeline parallelism, ZeRo, expert parallelism, and so on. However, inference with small batch size suffers from the following problems due to insufficient parallelism:
\begin{itemize}
    \item A smaller amount of data is processed at a time, which results in the need to read model weights from the HBM and call the kernel frequently, incurring a significant overhead.
    \item Each kernel call writes data to global memory, which the GPU has to re-read on the next kernel call, adding additional communication overhead.
    \item The current cuBLAS and CUTLASS GeMM libraries are not optimized for small batch sizes and have low memory bandwidth utilization.
\end{itemize}

\begin{figure}[!t]
    \centering
    \includegraphics[width=3.5in]{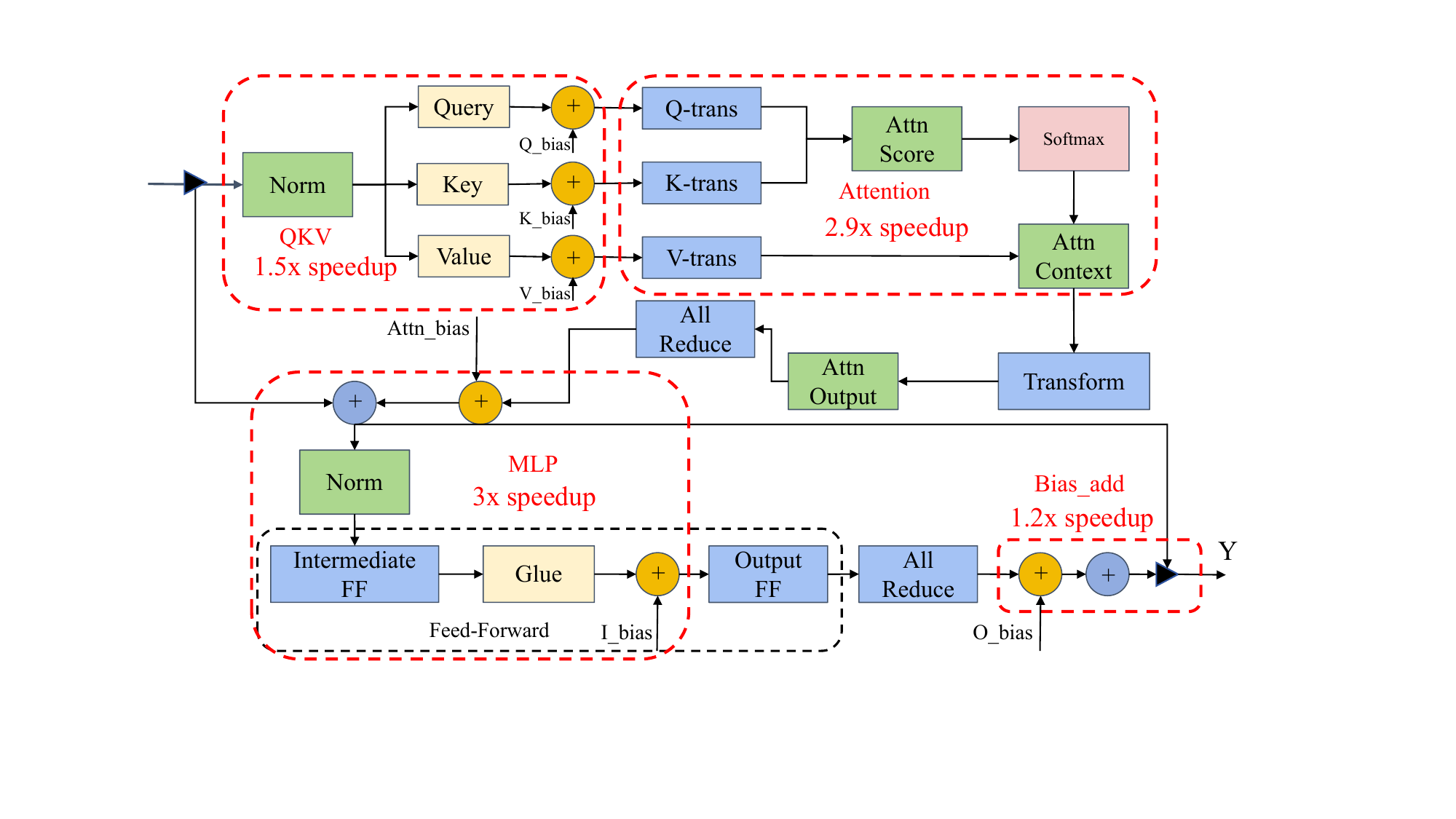}
    \caption{Deep-Fusion strategy for the small-batch inference.}
    \label{DeepFusion}
\end{figure}
On the other hand, regular operator fusion can only be done for element-wise operators. In contrast, operators in the Transformer structure introduce data dependencies across thread blocks, making it challenging to do operator fusion. This is because if another consumes data generated by one thread block on GPUs, a global memory synchronization is needed to invoke a new kernel. To avoid the need for global synchronization, Deep-Fusion tiles the computational space along the dimensions of the iteration space so that no cross-block data dependencies arise. In addition, it custom-designed GeMM for small batch sizes and implemented it to be fusible with Deep-Fusion for maximum memory bandwidth utilization.
Deep-Fusion does four operator fusions in a Transformer layer for small batch sizes to obtain four customized kernels, as shown in the red dashed box in the Fig.~\ref{DeepFusion} below. For large batch sizes, the operator fusion strategy is the same, with the difference that the GeMM in cuBLAS is used directly instead of the custom GeMM.

\subsection{Specialized Framework}
In this section, we will introduce some specialized frameworks that have been proposed recently. They are customized for specific scenarios and needs and can be tailored to meet different requirements. If you deploy models individually in high demand in certain aspects, consider using them.

Compared to other scenarios, efficiently deploying the Transformer model to servers needs to meet the service's low latency and high throughput requirements, which presents a significant challenge. In addition, due to the unpredictability of requests and the fact that NLP tasks employ variable-length sentences, the variability of input dimensions poses a severe problem for effective memory management and service optimization. TurboTransformer\cite{Turbotransformer} proposes a sequence-length-aware memory allocation algorithm and a batch scheduling algorithm that aims to maximize the response throughput by treating it as a dynamic programming problem. Efficient memory reuse of variable dimensional intermediate tensor and optimal batch scheduling scheme are realized. TurboTransformer also proposes a parallel approximation algorithm for high-frequency operators such as Softmax and LayerNorm, significantly improving efficiency. However, TurboTransformer's active grouping approach still introduces non-eliminable padding overheads. Based on this, ByteTransformer\cite{ByteTransformer} proposes a padding-free algorithm to free the whole Transformer from the redundant computation of zero-padded tokens. In addition, ByteTransformer optimizes multi-head attention for the zero-filling algorithm so that the attention is no longer faced with redundant computation of useless tokens, further improving performance.

Unlike the previous work, FlexGen\cite{Flexgen} sacrifices the latency of the inference computing service almost completely to polarize the design of an LLM computing system that focuses only on throughput. Thus, it is only suitable for offline computing. Every aspect of the LLM accelerator can be reconsidered when pursuing only throughput, including storage management, latency-hiding design of memory accesses in cross-domain memory hierarchies, and parallelization strategies.
FlexGen proposes a new offloading-based inference system based on a zig-zag parallelization strategy, which achieves more than 40 times higher throughput than DeepSpeed Zero-Inference. We believe the most inspiring aspect of this work is that it highlights the importance of divergent thinking and emphasizes the need to dig deeper into the details of the problem and explore alternative solutions.

Recently, the open-source inference framework PowerInfer\cite{PowerInfer} recently made LLM Inference 11 times faster. Without quantization and with FP16 precision, it allows 40B models to run smoothly on an RTX4090 PC; if quantization is added, a 2080 Ti can also run 70B models smoothly. It is based on highly localized sparse activation based on LLM, i.e., a small fraction of neurons \textbf{\textit{hot neurons}} are activated all the time on input, while the majority of neurons \textbf{\textit{cold neurons}} respond according to a specific input.
PowerInfer exploits this feature and the fact that CPUs are good at conditional computation and GPUs are good at simple parallel computation to develop an innovative GPU-CPU hybrid inference engine. This means \textbf{\textit{hot neurons}} are preloaded into the GPU for fast access. In contrast, \textbf{\textit{cold neurons}} are computed on the CPU, dramatically reducing the memory requirements of the GPU and the amount of data transferred between the CPU and the GPU. In addition, PowerInfer incorporates an adaptive predictor and neuron-specific sparse optimizations to improve the sparsity efficiency of neuron activation and computation.
Overall, PowerInfer enables PC users to run advanced LLM locally without needing expensive specialized hardware. This facilitates the popularization of AI applications and provides unprecedented opportunities for hobbyists, researchers, and small businesses. 

There are already more accelerators built for large model multi-GPU distributed inference, but relatively few accelerators are built for edge device deployment. As the demand for deploying AI large models on edge devices grows, this will become a pressing problem.

%% file: PEFT.tex
\renewcommand{\thesubsection}{\Alph{subsection}}  
\label{subsec:peft}
When training a large language model for specific tasks, it is often faster and more efficient to fine-tune a pre-trained model. There are two types of fine-tuning methods: full fine-tuning and parameter-efficient fine-tuning (PEFT). However, full fine-tuning can be expensive and may cause catastrophic forgetfulness problems. To address these issues, PEFT was developed. PEFT methods can be categorized into three categories: additive, selective, and reparameterization.

\begin{figure}[!t]
    \centering
    \includegraphics[width=3.5in]{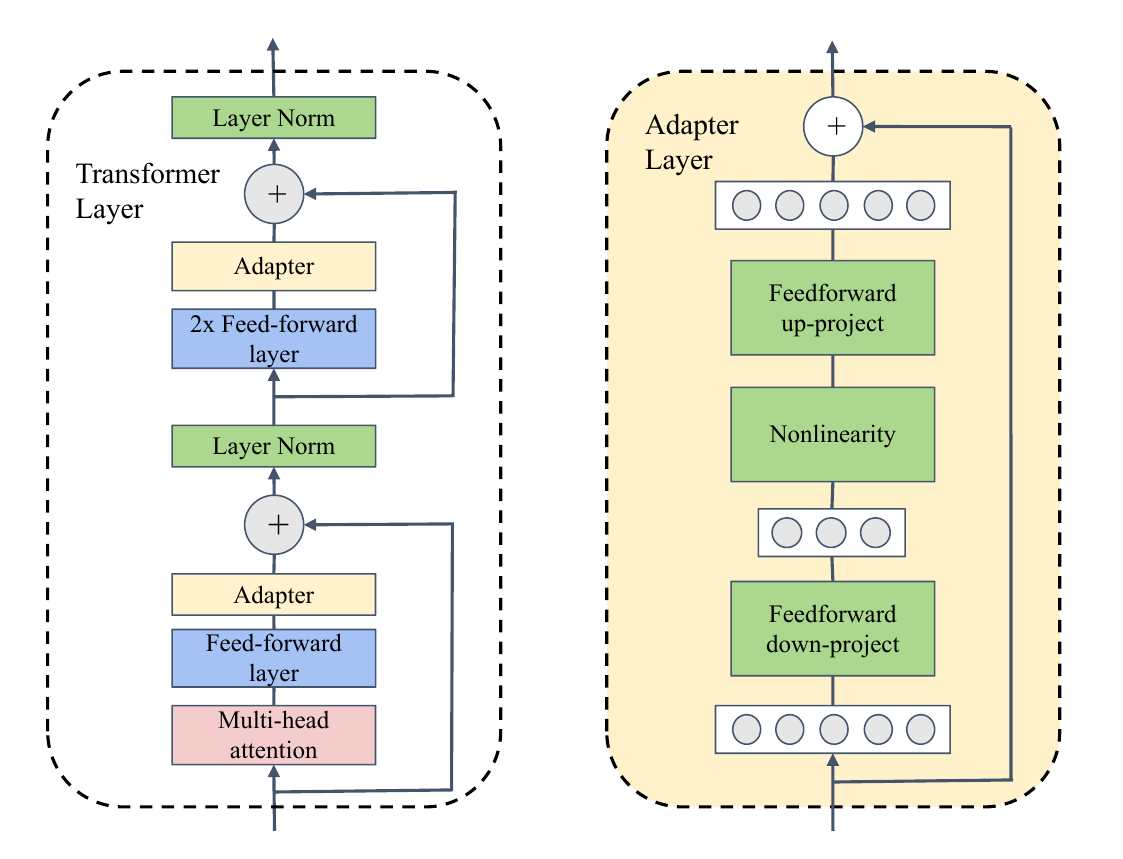}
    \caption{Architecture of the adapter module and its integration with the Transformer.}
    \label{adapter}
\end{figure}

\subsection{Additive methods}
The main idea behind additive methods is to add additional parameters to the model while keeping the original parameters fixed and train the model by fine-tuning the additional parameters\cite{Adapter, AdapterDrop, AdapterFusion, Sparseadapter, Prefix-tuning}.

The Adapter\cite{Adapter} is the pioneering work of the additive methods. As shown in Fig.~\ref{adapter}, the Adapter fine-tunes the model by inserting an adapter module between the feed-forward layer and skip-connection within the Transformer layer. The Adapter module is a small fully-connected network consisting of an upper projection layer, a nonlinear layer, a lower projection layer, and an additional skip-connection. Whenever a new downstream task arises, we can mitigate the problem of full fine-tuning and catastrophic forgetting by adding an adapter module to the model to produce an easily scalable downstream model. The adapter idea has been widely used and many adapter variants have been proposed.

There are two main strategies for integrating multi-task knowledge: sequential fine-tuning and multi-task learning. However, both strategies have specific problems. Sequential fine-tuning requires prior knowledge to determine the order of tasks which can result in the model forgetting knowledge learned from previous tasks. On the other hand, multi-task learning makes balancing data from various tasks challenging as different tasks can interfere with each other. As a result, both methods face challenges in effectively transferring knowledge.
\cite{AdapterFusion}proposes a variant called AdapterFusion, which effectively mitigates the above problems in multi-task training using a two-stage learning algorithm, achieves effective knowledge sharing across multiple tasks, and outperforms fully fine-tuned models on a single target task.

Typically, fine-tuning a Transformer with the adapter is 60$\%$ faster than full fine-tuning in training but 4-6$\%$ slower in inference. The AdapterDrop method proposed by \cite{AdapterDrop} can efficiently and dynamically remove adapters with minimal impact on task performance. This dramatically improves the model's efficiency during backpropagation (training) and forward propagation (inference).

Existing methods increase the bottleneck dimension to match the performance of full fine-tuning as much as possible because the number of trainable parameters determines the adapter's capacity. However, these methods increase the overall parameter count and FLOPs, violating the adapter's original intention. \cite{Sparseadapter} combines the adapter with pruning to propose SparseAdapter, using a frustratedly easy setting called Large-Sparse, which effectively improves the capacity of the adapter under a given parameter budget and can also be effectively applied to other types of adapter methods, such as LoRA.

Another similar class of additive methods is Soft Prompts, of which the most representative work is Prefix-Tuning\cite{Prefix-tuning}, which effectively avoids data cross-pollination and dramatically reduces the number of parameters by adding a continuous, independent, learnable, task-specific prefix to the input. At the same time, it can easily switch tasks and realize processing examples from multiple users/tasks in one batch.

Overall, while additive methods add additional parameters to the model and inevitably bring about additional inference delays, they significantly improve training speed and memory efficiency by reducing the gradient size and optimizer state.

\subsection{Selective methods}
Selective methods select a portion of the parameters on the original model for fine-tuning and keep the rest frozen, such as BitFit\cite{Bitfit}, which selectively fine-tunes only the model's bias terms. It is more suitable for cases with less training data because of the limited expression of the parameters it tunes. 

These methods are relatively more straightforward to implement. Still, the problem is that we can only determine which part to tune empirically and experimentally, which can be inefficient and inaccurate.In addition, since these methods modify the original model, we also need to consider the problem of model forgetting.

\subsection{Reparameterization methods}
The reparameterization methods take advantage of low rank in the weight matrix of the pre-trained model. Instead of fine-tuning the entire weight matrix, these methods construct smaller fine-tuned modules by reparameterization\cite{LoRA, QLoRA, QA-LoRA, Dylora, AdaLoRA, Compacter}.

The most representative work in this area is LoRA\cite{LoRA}, which approximates the entire weight matrix by fine-tuning the two-rank decomposition matrices instead of fine-tuning them. In this case, the rank $r$ of the decomposition matrices is a hyperparameter; the larger $r$ is, the more parameters need to be fine-tuned and the larger the capacity of the decomposition matrices, and vice versa. Due to the low-rank nature of the pre-trained model's weight matrix, it is generally sufficient for us to set r relatively small, better mitigating the forgetting problem. In addition, when reasoning, we can merge the decomposition matrix with the weight matrix to avoid introducing inference delays. When switching tasks, we must subtract the original parts and add new ones.

There are two problems with LoRA. The first is that its size is fixed, which means that once we have set the rank and trained on it, it cannot be modified anymore. The second is that it is difficult to find the optimal rank unless we perform a search, which is expensive. \cite{Dylora} has developed a new algorithm called DyLoRA based on LoRA. DyLora has developed a unique algorithm called DyLoRA based on LoRA. DyLora has developed a new algorithm called DyLoRA based on LoRA. DyLora has developed a new algorithm called DyLoRA based on LoRA. It dynamically searches for the optimal rank through LoRA selects it, and performs dynamic inference without incurring additional costs.

Another related work is AdaLoRA\cite{AdaLoRA}, which adaptively assigns parameter budgets based on importance scores. In AdaLoRA, the incremental update of the weight matrix is parameterized in the form of a singular value decomposition. The parameter budget is dynamically allocated among the total matrices by manipulating the distinct values according to the new importance metrics. This method effectively improves model performance and parameter efficiency.

LoRA has another advantage: it can be naturally orthogonalized to other model compression acceleration methods. As mentioned earlier, QLoRA\cite{QLoRA}, which combines LoRA with quantization, dramatically reduces the memory footprint during training, realizing the advancement of fine-tuning 65B parametric models on a single 48GB memory GPU, making it possible for more researchers to participate in the study of large models. However, QLoRA only considers resources at training, not inference. Although QLoRA quantizes the model during the training process, since the LoRA parameters for training are of FP16 type, when reasoning, the quantized model is fused with the LoRA parameters, the quantization will be destroyed and go back to the unquantized form, and then if you want to reason efficiently, you have to perform another step of PTQ (Post-training quantization). PTQ will bring extra errors and affect the performance of the model. Recently, QALoRA\cite{QA-LoRA} solved the problem by introducing group-wise operators, which realize one-click low-precision deployment of models.

\subsection{Hybrid methods}
Combining the previous methods is a natural idea to get better results. Compacter\cite{Compacter} combines the ideas of Adapter and reparameterization to balance the number of trainable parameters, task performance, and memory footprint. The above is a brief overview of representative work on PEFT. For more detailed information, see \cite{PEFT-survey}.

PEFT methods offer an effective way to train models, but it's still difficult to deploy trained models on edge devices like cell phones. We can tweak the models before compression, which may affect their performance. Therefore, it's important to explore new PEFT methods and find ways to combine them with other compression acceleration techniques to compress models while minimizing performance loss in downstream tasks. This remains a valuable area of research. 